\documentclass[10pt]{article} 
\usepackage[preprint]{tmlr}

\usepackage{amsmath,amsfonts,bm}

\def\eqref#1{equation~\ref{#1}}

\def\1{\bm{1}}

\DeclareMathAlphabet{\mathsfit}{\encodingdefault}{\sfdefault}{m}{sl}
\SetMathAlphabet{\mathsfit}{bold}{\encodingdefault}{\sfdefault}{bx}{n}

\usepackage{blindtext}
\usepackage{etoolbox}

\newcommand{\printthis}[2][true]{%
\ifbool{#1}{%
#2%
}{%
}%
}

\usepackage{xcolor}

\usepackage{tocloft}
\usepackage{hyperref}
\usepackage{url}

\usepackage{makecell}
\usepackage{graphicx}
\usepackage{algorithm} 
\usepackage{booktabs}
\usepackage{algpseudocode}
\usepackage{multirow}
\usepackage{adjustbox}
\usepackage{wrapfig}
\usepackage{caption}
\usepackage{subcaption}

\usepackage{amssymb}
\usepackage{calc}
\definecolor{priv}{RGB}{207,227,249}
\definecolor{privRe}{RGB}{197,241,207}
\definecolor{inv}{RGB}{252,215,241}
\definecolor{acc}{RGB}{238,178,28}
\definecolor{attacc}{RGB}{118,173,65}
\newcounter{oldtocdepth}
\newcommand{\hidefromtoc}{%
  \setcounter{oldtocdepth}{\value{tocdepth}}%
  \addtocontents{toc}{\protect\setcounter{tocdepth}{-10}}%
}
\newcommand{\unhidefromtoc}{%
  \addtocontents{toc}{\protect\setcounter{tocdepth}{\value{oldtocdepth}}}%
}
\def\D{{\mathcal D}}

\title{Random Erasing vs. Model Inversion: A Promising Defense or a False Hope?}

\author{\name Viet-Hung Tran$^{*}$ \email h.tran@qub.ac.uk \\
      \addr Temasek Laboratories, Singapore University of Technology and Design; The Queen's University Belfast 
      \vspace{-0.6em}
      \AND
      \name Ngoc-Bao Nguyen$^{*}$ \email thibaongoc\_nguyen@sutd.edu.sg \\
      \addr Temasek Laboratories, Singapore University of Technology and Design
      \vspace{-0.6em}
      \AND
      \name Son T. Mai \email thaison.mai@qub.ac.uk\\
      \addr The Queen's University Belfast 
      \vspace{-0.6em}
      \AND
      \name Hans Vandierendonck \email h.vandierendonck@qub.ac.uk\\
      \addr The Queen's University Belfast      
      \vspace{-0.6em}
      \AND
      \name Ira Assent \email ira@cs.au.dk\\
      \addr Aarhus University      
      \vspace{-0.6em}
      \AND
      \name Alex Kot \email eackot@ntu.edu.sg\\
      \addr Nanyang Technological University (NTU); VinUniversity, Hanoi, Vietnam       
      \vspace{-0.6em}
      \AND
      \name Ngai-Man Cheung$^{\dag}$  \email ngaiman\_cheung@sutd.edu.sg\\
      \addr Temasek Laboratories, Singapore University of Technology and Design
      }

\def\month{08}  
\def\year{2025} 
\def\openreview{\url{https://openreview.net/forum?id=S9CwKnPHaO}} 

\newcolumntype{L}{>{\centering\arraybackslash}m{1cm}}

\newcommand\blfootnote[1]{%
  \begingroup
  \renewcommand\thefootnote{}\footnote{#1}%
  \addtocounter{footnote}{-1}%
  \endgroup
}
\begin{document}

\maketitle
\hidefromtoc

\blfootnote{$^*$ The first two authors contributed equally. $\dag$ Corresponding author.}

\begin{abstract}\label{sec:abstract}
Model Inversion (MI) attacks pose a significant privacy threat by reconstructing private training data from machine learning models. 
While existing defenses primarily concentrate on model-centric approaches, the impact of data on MI robustness remains largely unexplored.
In this work, we explore
{\em Random Erasing (RE)}—a technique traditionally used for improving model generalization under occlusion—and uncover its surprising effectiveness as a defense against MI attacks.

Specifically, our novel feature space analysis shows that model trained with RE-images introduces a significant discrepancy between the features of MI-reconstructed images and those of the private data. At the same time, features of private images remain distinct from other classes and well-separated from different classification regions. These effects collectively degrade MI reconstruction quality and attack accuracy while maintaining reasonable natural accuracy. Furthermore, we explore two critical properties of RE including Partial Erasure and Random Location. 
First, {\em Partial Erasure} prevents the model from observing entire objects during training, and we find that this has significant impact on MI, which aims to reconstruct the entire objects. Second, the {\em Random Location} of erasure plays a crucial role in achieving a strong privacy-utility trade-off. 
Our findings highlight RE as a simple yet effective defense mechanism that can be easily integrated with existing privacy-preserving techniques. Extensive experiments of 37 setups demonstrate that our method achieves SOTA performance in privacy-utility tradeoff. The results consistently demonstrate the superiority of our defense over existing defenses across different MI attacks, network architectures, and attack configurations.
For the first time, we achieve significant degrade in attack accuracy {\em without} decrease in utility for some configurations.
Our code and additional results are available at: \url{https://ngoc-nguyen-0.github.io/MIDRE/}

\end{abstract}
    
\section{Introduction}
\label{sec:intro}

\begin{figure}[ht!]
  \centering
  \includegraphics[width=1.0\columnwidth]{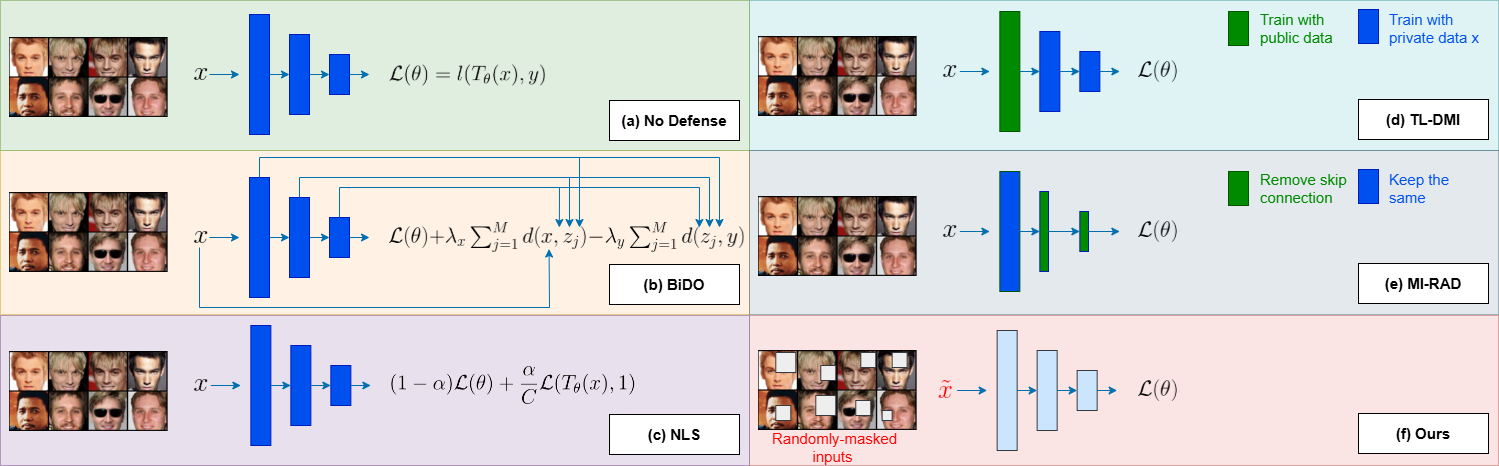}
  \caption{
{\bf Our Proposed Model Inversion (MI) Defense via Random Erasing (MIDRE).}  
(a)  ``No Defense'': Training a model without MI defense. $\mathcal{L}(\theta)$ is the standard training loss, e.g., cross-entropy.
Training a model with state-of-the-art MI defense (SOTA) (b) BiDO \citep{peng2022bilateral}, (c) NLS \citep{struppekcareful}, and (d) TL-DMI \citep{ho2024model}, (e) MI-RAD \citep{koh2024vulnerability} , (f) Our method.
Studies in \citep{peng2022bilateral,struppekcareful} focus on \textbf{adding new loss} to the training objective in other to find the balance between model utility and privacy. 
Both TL-DMI \citep{ho2024model}  and MI-RAD \citep{koh2024vulnerability} focus on \textbf{the model's parameters} to defend against MI.
For our proposed method (f), the training procedure and objective are the same as that in  (a) ``No Defense''. However, the training samples presented to the model are partially masked, thus, reducing  private training sample's information encoded in the model and \textbf{preventing the model from observing the entire images}. 
Therefore, \textbf{MIDRE 
is different from other approaches and focuses on input data only to defend.}
We find that this can significantly degrade MI attacks, which  
require substantial amount of private training data information encoded inside the model in order to reconstruct high-dimensional private images.
}
  \label{fig:Architecture_Overview}
\end{figure} 
Machine learning and deep neural networks (DNNs) \citep{lecun2015deep} have demonstrated their utility across numerous domains, including computer vision \citep{voulodimos2018deep, o2020deep}, natural language processing \citep{otter2020survey}, and speech recognition \citep{deng2013new, nassif2019speech}. DNNs are now applied in critical areas such as medical diagnosis \citep{azad2021medical}, medical imaging \citep{shen2017deep, lundervold2019overview}, facial recognition \citep{ wang2021deep, guo2019survey, masi2018deep}, and surveillance \citep{zhou2021deep, harikrishnan2019vision,  hashmi2021application}. 
However, the potential risks associated with the widespread deployment of DNNs raise significant concerns.
In many practical applications, privacy violations involving DNNs can result in the leakage of sensitive and private data, eroding public trust in these applications. Defending against privacy violations of DNNs is of paramount importance.

One specific type of privacy violation is Model Inversion (MI) attacks on machine learning and DNN models. MI attacks aim to reconstruct private training data by exploiting access to machine learning models.
Recent advancements in MI attacks including GMI \citep{zhang2020secret}, KedMI \citep{chen2021knowledge},
PPA \citep{struppek2022plug}, MIRROR \citep{an2022mirror}, IF-GMI \citep{qiu2024closer}, PPDG-MI
\citep{NEURIPS2024_3a797b10}, PLG-MI \citep{yuan2023pseudo}, and LOMMA \citep{nguyen2023re} have achieved remarkable progress in attacking important face recognition models.  This raises  privacy  concerns for models that are trained on sensitive data, such as face recognition, surveillance and medical diagnosis.

\textbf{Related works.}
Existing MI defenses primarily focus on model-centric strategies like model gradients \citep{dwork2006differential,dwork2008differential}, loss functions \citep{wang2021improving,peng2022bilateral,struppekcareful}, model features \citep{ho2024model}, and architecture designs \citep{koh2024vulnerability} (see Fig. \ref{fig:Architecture_Overview}).
Earlier works \citep{dwork2006differential,dwork2008differential} demonstrated the ineffectiveness of traditional Differential Privacy (DP) mechanisms against Model Inversion (MI) attacks. Recent research \citep{wang2021improving,peng2022bilateral,struppekcareful} has explored the impact of loss functions on MI resilience. \cite{wang2021improving} restricted the dependency between model inputs and outputs, while BiDO \citep{peng2022bilateral} focused on limiting the dependency between model inputs and latent representations. To partially restore model utility, BiDO maximized the dependency between latent representations and outputs. \cite{struppekcareful} proposed using negative label smoothing factors as a defense. However, these loss function-based approaches often introduce conflicting objectives, leading to significant degradation in model utility.
Recently, TL-DMI \citep{ho2024model} restricted the number of layers to be encoded by the private training data, while MI-RAD \citep{koh2024vulnerability} found that removing skip connections in final layers enhances robustness. However, both approaches 
experience difficulty in achieving competitive balance between utility and privacy.

Data augmentation, a technique that creates new, synthetic samples from existing data points, offers a promising avenue for enhancing model robustness. In this paper, we pioneer the investigation of Random Erasing (RE) \citep{zhong2020random} for MI defense. RE, traditionally used to improve model generalization for detecting occluded objects by removing randomly a region in training samples, demonstrates its effectiveness as a powerful defense against MI attacks.
In MI attacks, adversaries optimize reconstructed images to align with the target model's feature space representation of training samples. 
As will be shown in our novel analysis, 
thanks to RE, the target model's feature representations are inherently biased towards the RE-private images, the training data, rather than the private data. Consequently, \textbf{RE creates a discrepancy between the features of MI-reconstructed images and that of private images}, resulting in degraded MI attacks. Meanwhile, features of private images remain distinct from other classes, maintaining reasonable natural accuracy.
Furthermore, we highlight \textbf{two crucial properties of RE that serve as an effective MI defense: \textit{Partial Erasure} and \textit{Random Location}}. On the one hand, \textit{Partial Erasure significantly reduces the amount of private information embedded in the training data, preventing the model from observing the entire image}, and consequently degrades the MI attacks.
On the other hand, \textit{Random Location improves the diversity of training data}, thereby,
enhances the model utility. 
Our proposed method leads to substantial degradation in
MI reconstruction quality and attack accuracy (See  Sec.~\ref{Sec:MI_Defense_Analysis} for our comprehensive analysis and validation).
Meanwhile, 
natural accuracy of the model is only moderately affected.
Overall, we can achieve state-of-the-art  performance in privacy-utility trade-offs as demonstrated in our extensive  experiments of 37 setups -- 9 SOTA MI attacks including both white-box, black-box, and label-only MI attacks, 11 model architectures (including vision transformer), 6 datasets and different resolution including $64\times64$, $160\times160$, and $224\times224$ resolution -- and user study (in Supp.).  
Our contributions are:

\begin{itemize}
\item Our novel defense method, Model Inversion Defense via Random Erasing (MIDRE), is the first work to consider the well-known RE technique as a privacy protection mechanism, leveraging its powerful ability to reduce MI attack accuracy while maintaining model utility.  All results support the SOTA effectiveness of a simple technique in addressing a critical security concern.
\item We provide a deep understanding on feature space analysis of Random Erasing's defense effectiveness which leads to reduce of MI attacks in MIDRE model.
\item 
Our analysis investigates two crucial properties of RE that serve as an effective MI defense: Partial Erasure and Random Location. With these two properties, our defense method degrades the attack accuracy while the impact on model utility is small.

\item
We conduct extensive experiments
(Sec.
\ref{sec:experiment}, Supp.) and user study (Supp.)
to demonstrate that our MIDRE can achieve SOTA privacy-utility trade-offs.
Notably, in the high-resolution setting, our MIDRE is the first to achieve competitive MI robustness without sacrificing natural accuracy.
Note that our method is very simple to implement and is complementary to existing MI defense methods.
\end{itemize}

\section{Our Approach: Model Inversion Defense via Random Erasing (MIDRE) }

\subsection{ Model inversion}
A model inversion (MI) attack aims to reconstruct private training data from a trained machine learning model. The model under attack is called a {\em target model}, $T_\theta$.
The target model $T_\theta$ is trained on a private dataset $\D_{priv}=\{(x_i, y_i)\}^N_{i=1}$, where $x_i$ represents the private, sensitive data and $y_i$ represents the corresponding ground truth label. For example, $T_\theta$ could be a face recognition model, and $x_i$ is a face image of an identity.
The model is trained with standard loss function 
$\ell$ that penalizes the difference between model output $T_\theta(x)$ and $y$.

\textbf{MI attacks.} The underlying idea of MI attacks is to 
seek a reconstruction $x$ that achieves maximum likelihood for a label $y$ under $T_\theta$:
\begin{equation}
\max_x \mathcal{P}(y;  x, T_\theta)
\label{eq:L_MI}
\end{equation}
In addition, some 
prior to improve reconstructed image quality can be included \citep{zhang2020secret,chen2021knowledge}.
SOTA MI attacks \citep{zhang2020secret,chen2021knowledge,nguyen2023re,struppek2022plug} also apply GAN trained on a public dataset $\D_{pub}$ to limit the search space for $x$. 
$\D_{pub}$ has no identity intersection with 
$\D_{priv}$, assuming attackers can not access to any private samples.

\subsection{Random Erasing (RE) as a defense} \label{random_erasing}

Random Erasing (RE) \citep{zhong2020random} involves employing a random selection process to identify an region inside an image. Subsequently, this region is altered through the application of designated pixel values, such as zero or the mean value obtained from the dataset, resulting in {\em  partial masking} of the image. Traditionally, RE is applied as a  data augmentation technique to improve  robustness of machine learning models in the presence of object occlusion \citep{zhong2020random}.

We propose a simple configuration of RE as a MI defense, requiring only one hyper-parameter.
Given a training sample $x$ with dimensions $W\times H$, we propose a square region erasing strategy to restrict private information leakage from $x$.
We initiate by randomly selecting a starting point, denoted as  $(x_e, y_e)$, within the bounds of $x$.
Next, we randomly select the erased area portion $a_e$ within the specified range of $[0.1, a_h]$, guaranteeing at least 10\% of $x$ is erased during training, while 
$a_h$ is the only hyper-parameter of our defense.
The size of the erased region is $\sqrt{s_{RE}} \times \sqrt{s_{RE}}$ where  $s_{RE} = W\times H\times a_e$ is the erased region. 
With the designated region, we determine the coordinates of the erased region $(x_e, y_e, x_e+\sqrt{s_{RE}}, y_e+\sqrt{s_{RE}})$. However, we need to ensure this selected region stays entirely within the boundaries of $x$, 
i.e. $x_e+\sqrt{s_{RE}} \leq W$, $y_e+\sqrt{s_{RE}} \leq H$. 
If the erased region extends beyond the image width or height, we simply repeat the selection process until we find a suitable square erased region that fits perfectly within $x$. 
We fill the erased regions with ImageNet mean pixel value (See Supp
for a detailed discussion on the impact of the erased value) to obtain the RE-image. 
Note that RE is applied to all private training samples and the size and position vary each training iteration. 
We depict our method in Algorithm 1 (Supp.)

\section{Analysis of privacy effect of MIDRE}\label{Sec:MI_Defense_Analysis}

In this section, we analyze the privacy impact of RE within our proposed MIDRE framework. We conduct a thorough analysis and demonstrate that RE can achieve a surprisingly state-of-the-art balance between utility and privacy. Specifically, when employed as a defense against MI attacks, RE is the first method to significantly reduce attack accuracy without compromising utility in certain configurations, whereas all prior MI defenses exhibit noticeable degradation in utility to achieve similar reductions in attack success. Experimental results in Sec. 
\ref{sec:experiment}
further validate this finding.

We delve deeper into the mechanisms that underpin the effectiveness of RE.
Importantly, 
we conduct a feature space analysis to explain RE's defense effectiveness, showing that model trained with MIDRE leads to a discrepancy between the features of MI-reconstructed images and that
of private images, resulting in degrading of attack accuracy. 
At the same time, private images remain distinct from other classes and well-separated from different classification regions, maintaining reasonable natural accuracy.
Furthermore, our analysis reveals that partial erasure, as implemented in RE, is a highly effective method for mitigating MI attacks. 
Particularly, to present the model with less private pixels during training, our approach of applying partial erasure while maintaining the original number of training epochs proves to be more effective than the alternative approach of reducing the number of epochs without using partial erasure.
We attribute this to the fact that MI attacks rely on the target model to reconstruct the {\bf entire image}, and RE's partial erasure prevents the target model from ever fully observing the entire image throughout the training process. 
Additionally, we show that applying partial erasure at random locations, as is done in RE, is more effective than erasure at fixed locations.

\subsection{RE degrades MI significantly, achieving SOTA privacy-utility trade-off}

\begin{figure*}[t!]
\centering
\includegraphics[width=0.27\textwidth]{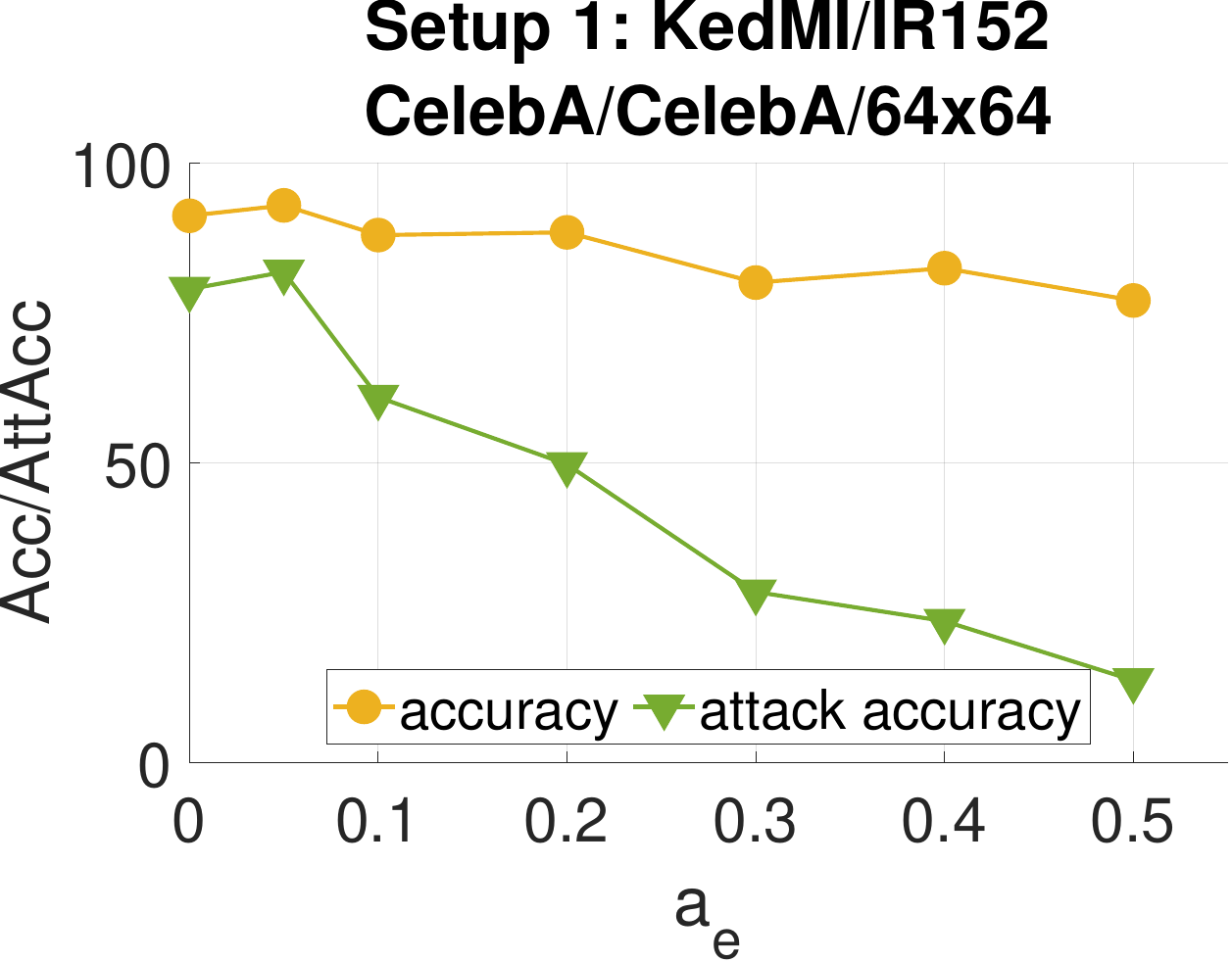}
\hspace{0.3cm}
\includegraphics[width=0.27\textwidth]
{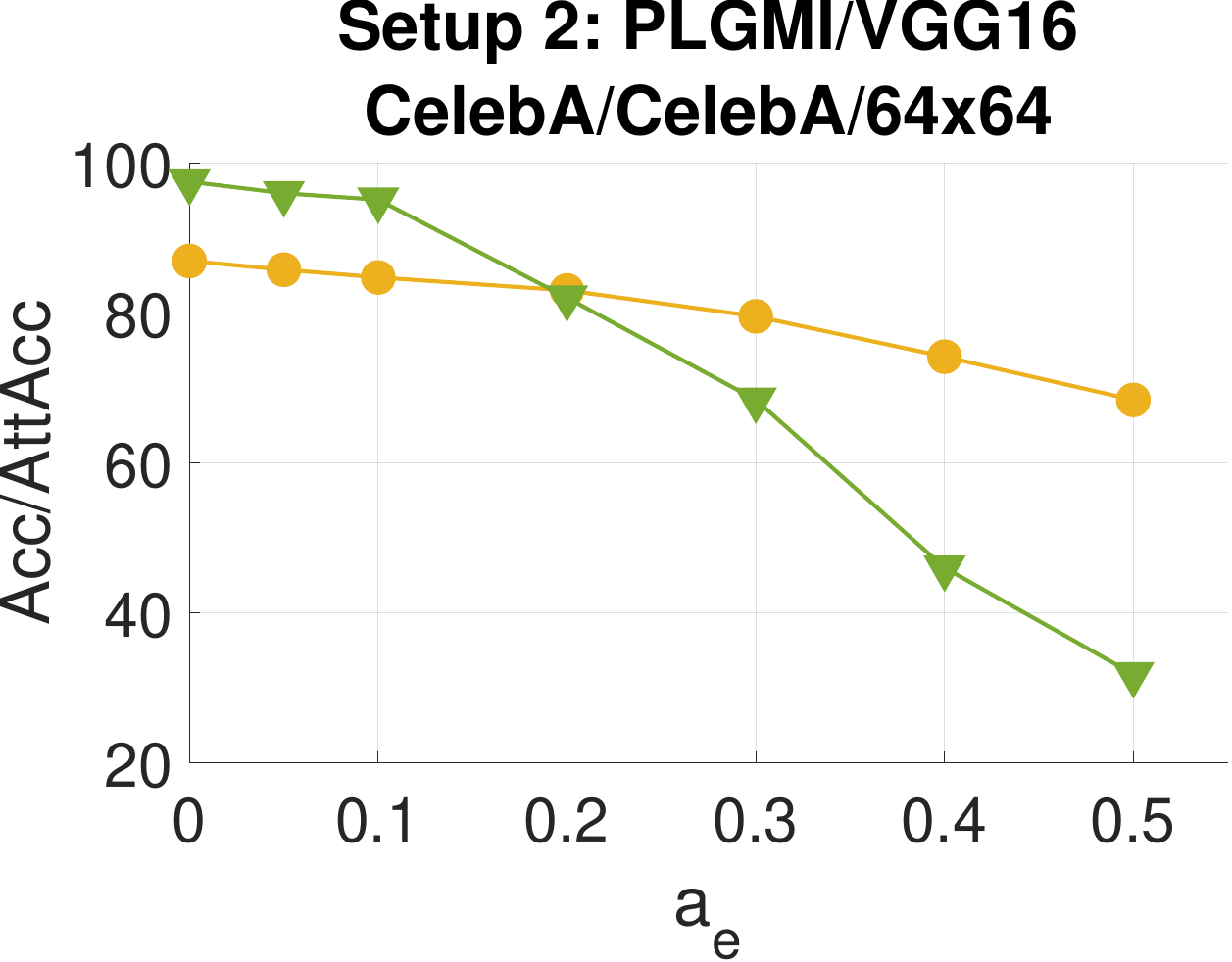}
\hspace{0.3cm}
\includegraphics[width=0.27\textwidth]{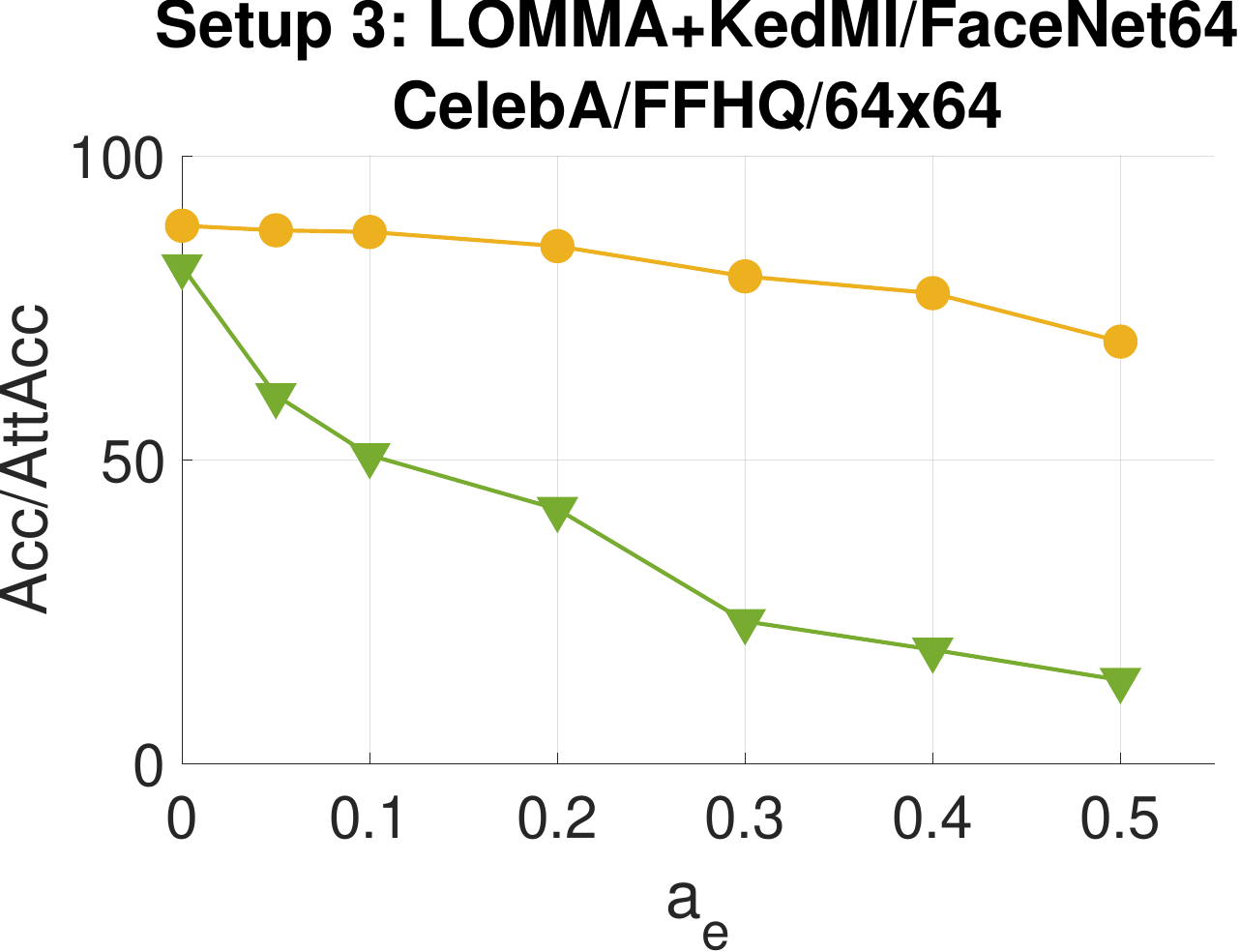}

\includegraphics[width=0.27\textwidth]{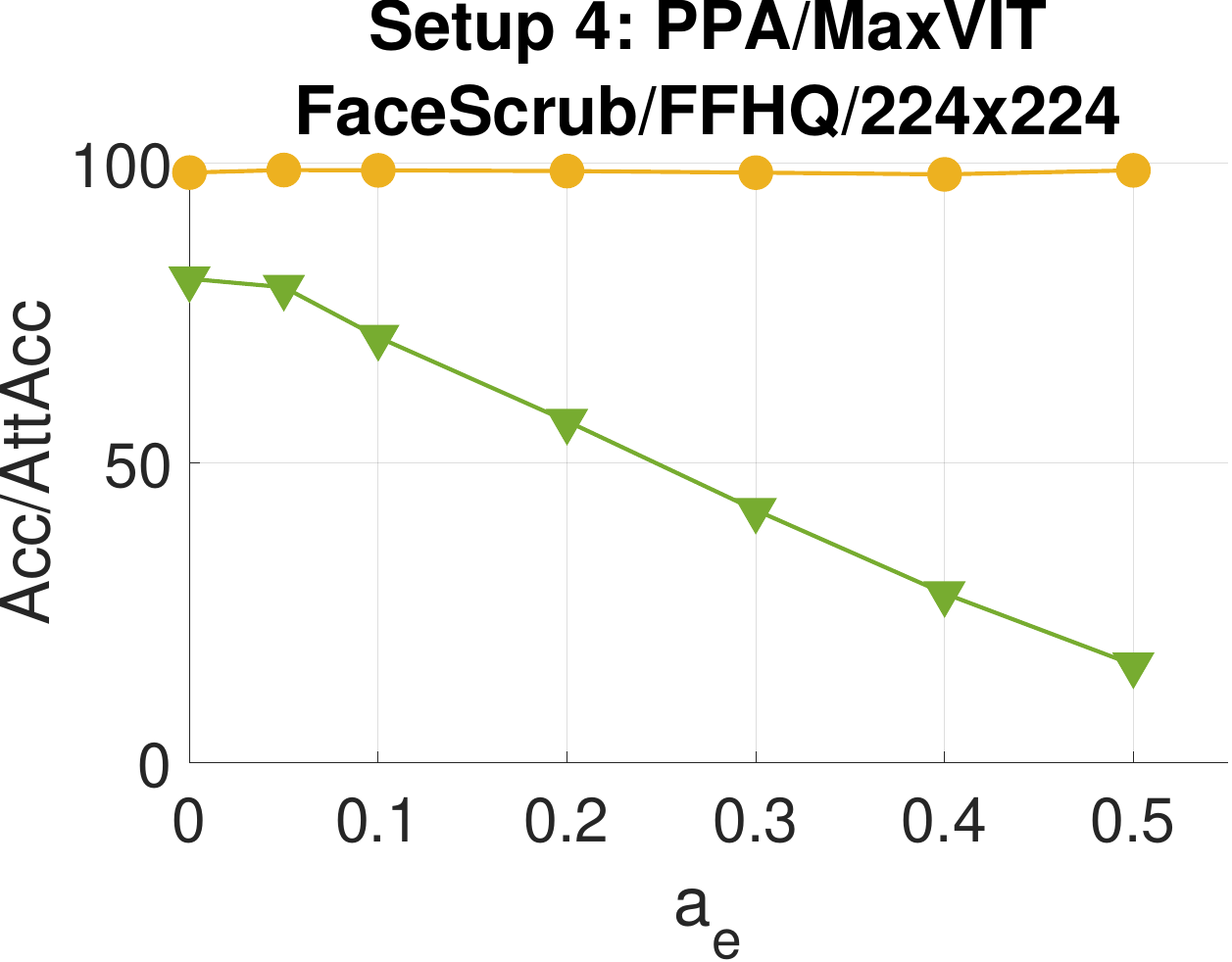}
\hspace{0.3cm}
\includegraphics[width=0.27\textwidth]{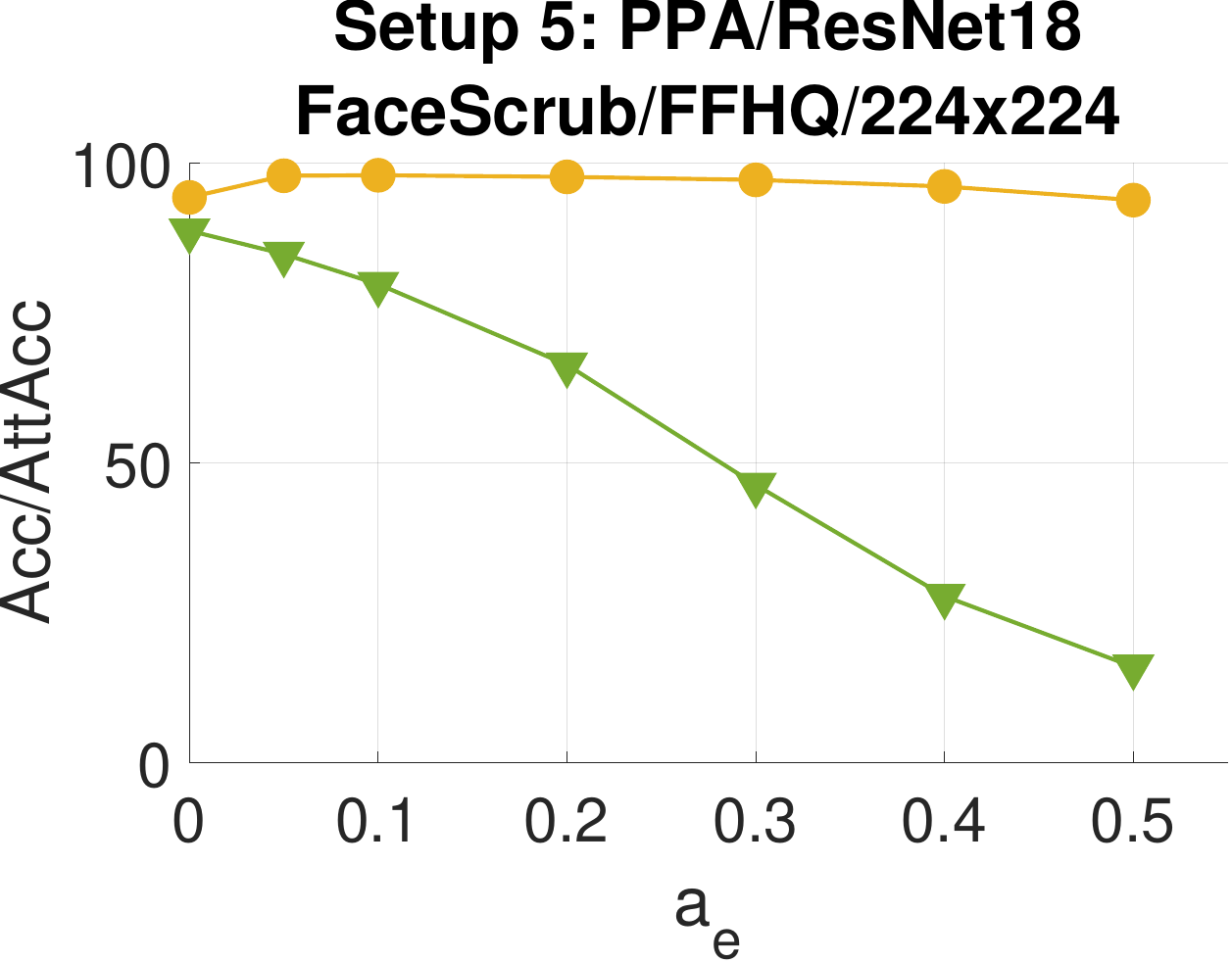}
\hspace{0.3cm}
\includegraphics[width=0.27\textwidth]{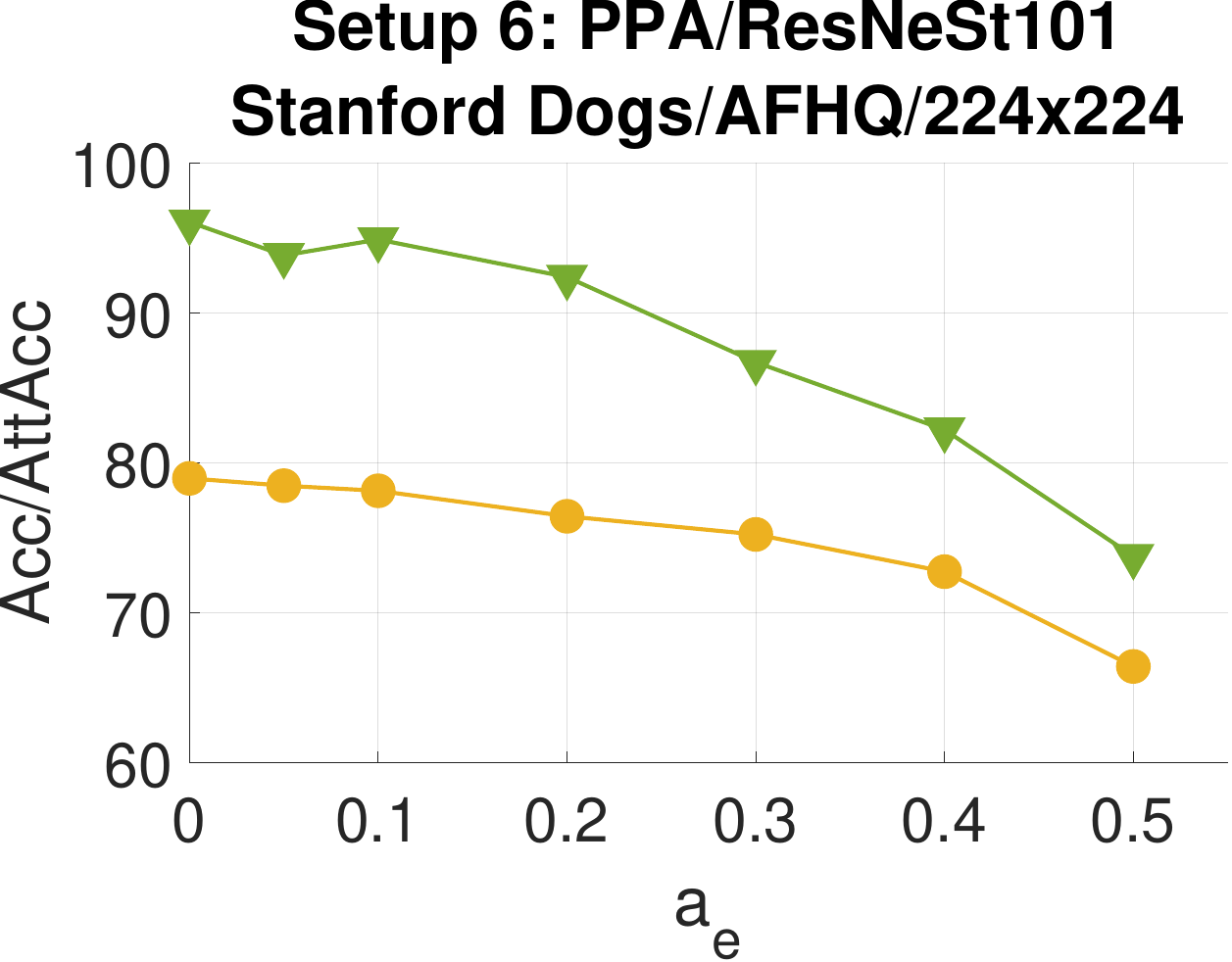}
\caption{{\bf Our analysis shows that Random Erasing (RE) can lead to substantial degradation in MI reconstruction quality and attack accuracy, while natural accuracy of the model is only moderately affected.}
In this analysis, we conduct 6 experimental setups with different \textit{MI attacks/target models architecture/private/public datasets/image resolution}. 
We analyze the attack ({\color{attacc}green line}) and natural accuracy ({\color{acc}orange line}) of the target models under different extents of random erasing applied in the training stage, using random erasing ratio 
$a_e = s_{RE}/(W \times H)$
as discussed in 
Sec. \ref{random_erasing}.
To properly reconstruct private high-dimensional facial images of individuals, MI attacks require  
 significant amount of private training data information encoded inside the model. We find that the model using RE by small percentages can significantly degrade MI attacks, with MI attack accuracy decreasing, for example, from 15.47\% to 39.93\%. However, the natural accuracy of the model only decreases slightly, less than 4\%, as sufficient information remained in the RE-images for the model to learn to discriminate between individuals (Setup 1-3).
We also observe a high degradation in MI attack accuracy while the model accuracy increases. For instance, model accuracy increases by 0.37\%, while attack accuracy decreases by 69.39\% (Setup 4).
Overall, our defense demonstrates state-of-the-art privacy-utility trade-offs and can improve model utility in certain setups  
} 
\label{fig:Redundant_Analysis}
\vspace{-0.2cm}
\end{figure*}

In the analysis, we study attack accuracy and natural accuracy of a target model $T_\theta$  under different erased region portions $a_e$.  Recall   $a_e = s_{RE} / (W\times H)$, and $\sqrt{s_{RE}}\times\sqrt{s_{RE}}$ is the dimension of the erased region. 
For the target model, which is a face recognition model, in each setup, we employ the same architecture and hyper-parameters, while modifying the erased region portions 
$a_e$. 
Specifically, we fix the values of $a_e$ for this analysis, to study
the effect of $a_e$ to model utility (accuracy) and model privacy (attack accuracy).
We vary $a_e$ from 0.0 
 (indicating no random erasing and the same as No Defense) to 0.5 (erasing 50\% of each input image). After the training of $T_\theta$, we proceed to evaluate its top 1 attack accuracy using SOTA MI attacks. This evaluation is conducted for all target models trained with different $a_e$. 
 In order to ensure diversity in our study, we employ six distinct setups for model inversion attacks, target model architecture, private dataset, and public dataset, and both low- and high-resolution datasets.

{\bf RE has small impact on model utility while degrading MI attacks significantly.}
Fig. \ref{fig:Redundant_Analysis} summarizes the impact of erased portions on model performance and model inversion attacks. In all setups, we demonstrably improve robustness against MI attacks with small sacrifice to natural accuracy.
For instance, 
introducing erased portions $a_e$ at a ratio of 0.2 in Setup 1 caused a small 2.76\% decrease in natural accuracy while the attack accuracy plummeted by 29.2\%.
This trend continues in Setup 2 – a 0.2 ratio of $a_e$ led to a modest 3.92\% decrease in natural accuracy, but a substantial 15.47\% drop in attack accuracy. We note that in Setup 3, LOMMA+KedMI attack accuracy degrades by 39.93\%.
For high resolution images (Setup 4, 5), we observe an increase in model accuracy when using RE.
In Setup 4, there is a significant 69.39\% drop in attack accuracy while natural accuracy slightly increase (0.37\%) when $a_e$ = 0.5. Similar trend for Setup 5, attack accuracy drops from 88.67\% to 27.75\% when $a_e$ = 0.4 while natural accuracy increases 1.83\%.
In conclusion, \textit{using RE-images during training significantly
degrades MI attack while impact on natural accuracy is small.}
\subsection{Feature space analysis of Random Erasing's defense effectiveness}
\label{sec:feature_space}

In this section, we present a novel observation that explains RE's defense effectiveness. We observe  {\color{blue}\textbf{Property P1:}} \textbf{Model trained with RE-private images following our  MIDRE leads to a discrepancy between the features of MI-reconstructed images and that of  private images}, resulting in degrading of attack accuracy.

The following analysis explains why MIDRE has {\color{blue}\textbf{Property P1}}.
We use the following notation: $f_{train}$, \colorbox{priv}{$f_{priv}$}, \colorbox{privRe}{$f_{RE}$}, and \colorbox{inv}{$f_{recon}$} represent the features of training images, \colorbox{priv}{private images}, \colorbox{privRe}{RE-private images}, and \colorbox{inv}{MI-reconstructed images}, respectively. To extract these features, we first train the target model without any defense (NoDef) and another target model  with our MIDRE. Then, we pass images into these models to obtain the penultimate layer activations. Specifically, we input private images into the models to obtain $f_{priv}$. Next, we apply RE to private images, pass these RE-private images into the models to obtain $f_{RE}$. We also perform MI attacks to obtain reconstructed images from NoDef model (resp. MIDRE model), and then feed them into the NoDef model (resp. MIDRE model) to obtain $f_{recon}$.
We use the same experimental setting as in Sec. \ref{sec:RE_freq}. Then, we visualize penultimate layer activations  $f_{priv}, f_{RE}, f_{recon}$ by both NoDef and our MIDRE model. We use $a_e$ = 0.4 to train MIDRE and to generate RE-private images.
Additionally, we visualize the convex hull of these features.

\begin{figure}[t]
 \centering
 \begin{subfigure}[b]{0.95\textwidth}
     \centering
     \includegraphics[width=0.24\textwidth]{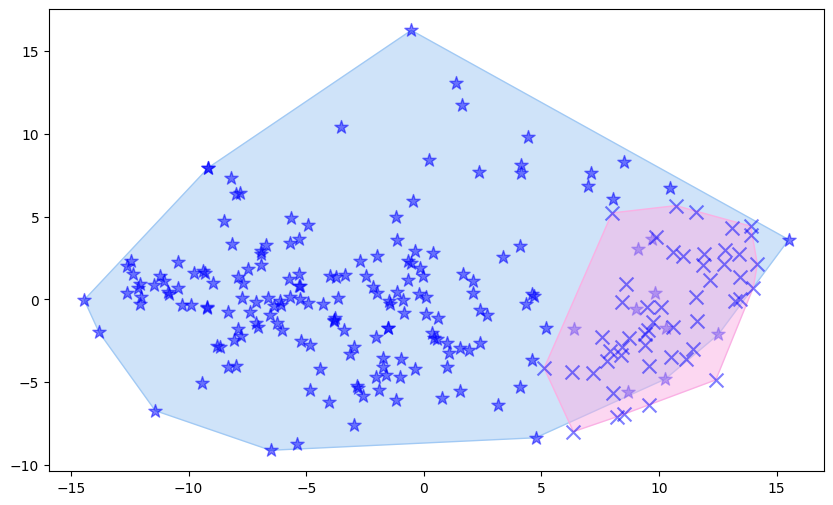}
     \includegraphics[width=0.24\textwidth]{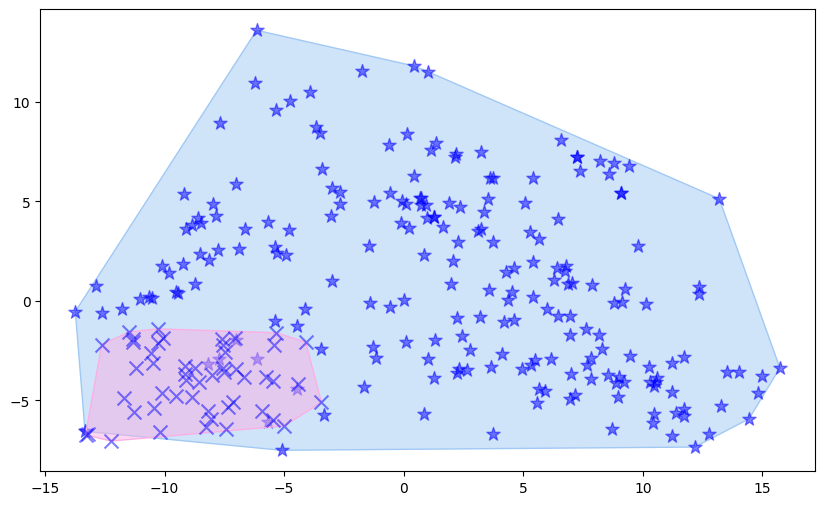}
     \includegraphics[width=0.24\textwidth]{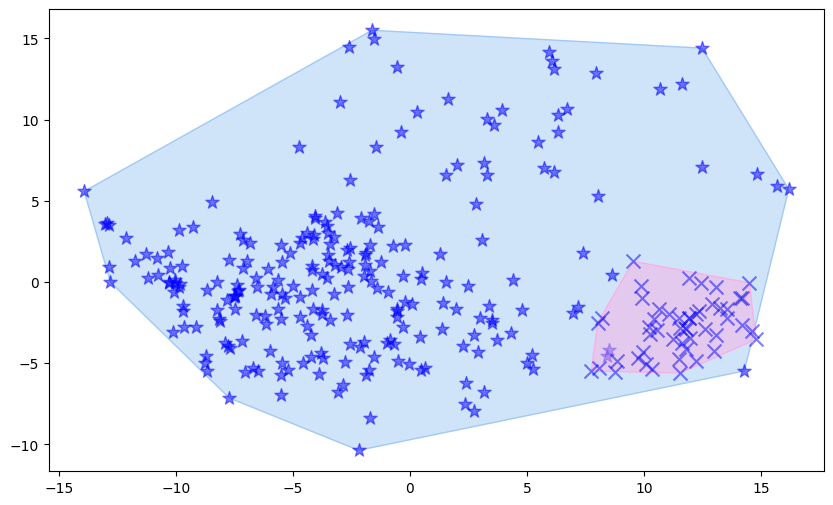}
     \includegraphics[width=0.24\textwidth]  {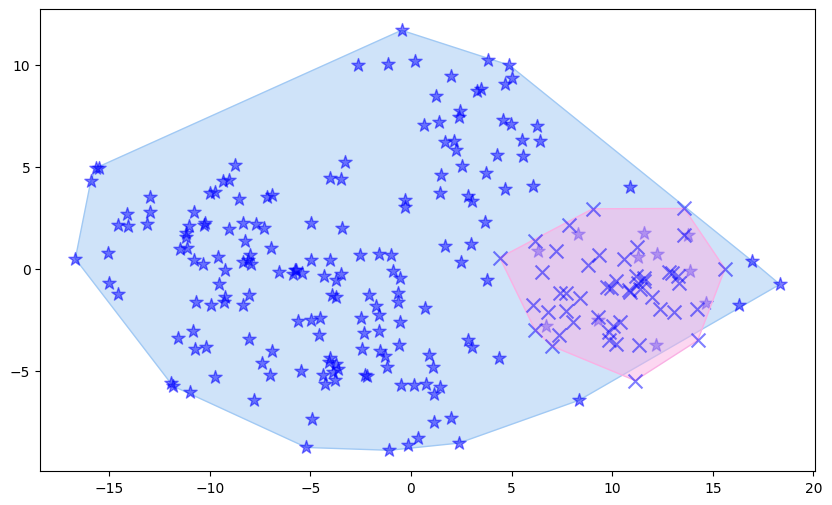}
     \vspace{-0.15cm}
     \caption{NoDef, AttAcc = 88.67\%, $f_{train}^{NoDef} = $ \colorbox{priv}{$f_{priv}^{NoDef}$}}
 \end{subfigure}
 
 \begin{subfigure}[b]{0.95\textwidth}
     \centering
     \includegraphics[width=0.24\textwidth]{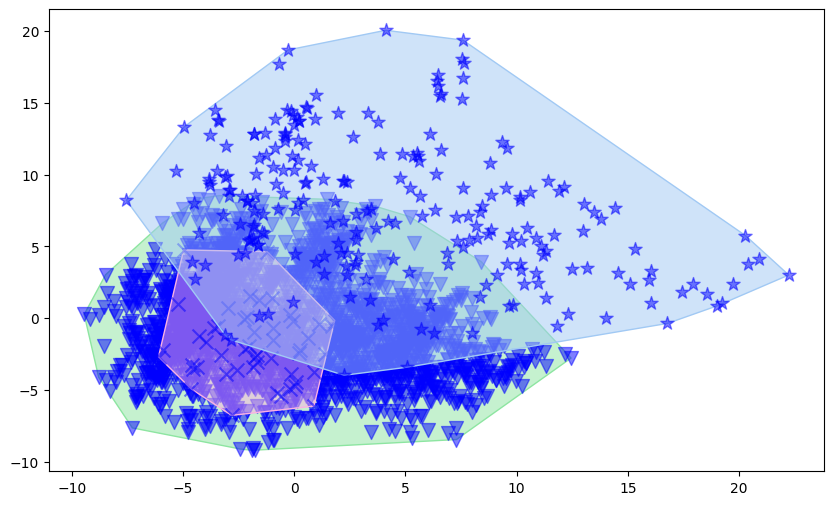}
     \includegraphics[width=0.24\textwidth]{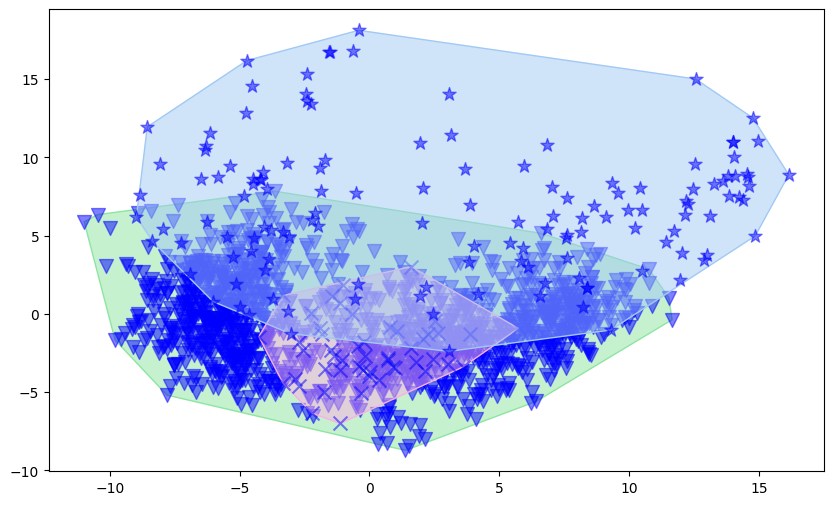}
     \includegraphics[width=0.24\textwidth]{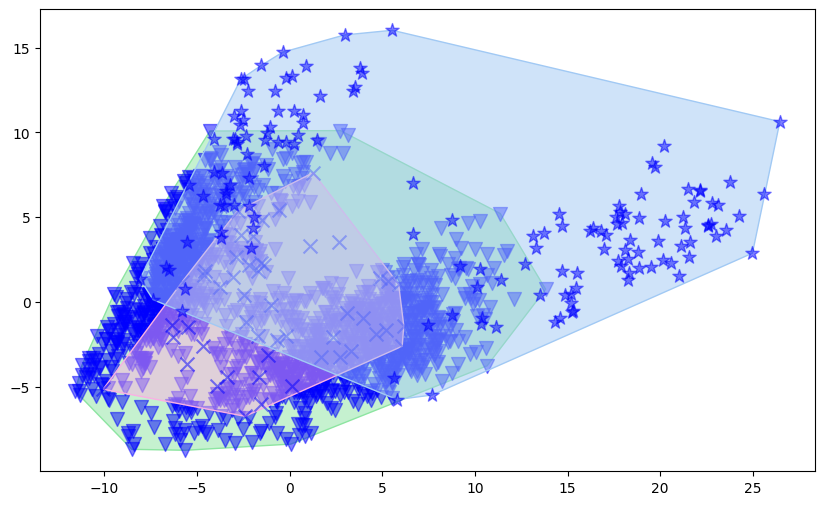}
     \includegraphics[width=0.24\textwidth]{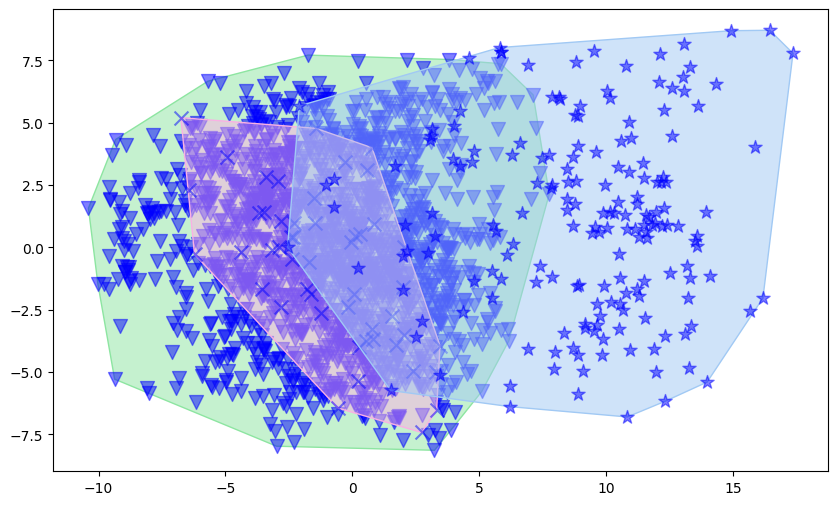}
     \vspace{-0.15cm}
     \caption{MIDRE, $a_e$ = 0.4, AttAcc = 27.75\%, $f_{train}^{MIDRE} =$ \colorbox{privRe}{$f_{RE}^{MIDRE}$}}
 \end{subfigure}
 \vspace{-0.2cm}
\caption{\textbf{Feature space analysis to show that, under MIDRE, $f_{recon}^{MIDRE}$ and $f_{priv}^{MIDRE}$ has a discrepancy, degrading MI attack.} We visualize penultimate layer activations of private images 
({\bf\color{blue}$\star$} $f_{priv}$), RE-private images 
({\color{blue}$\blacktriangledown$} $f_{RE}$), 
and MI-reconstructed images ({\bf\color{blue}$\times$} $f_{recon}$) generated by both (a) NoDef and (b) our MIDRE model. 
We also visualize the convex hull for  \colorbox{priv}{private images},  \colorbox{privRe}{RE-private images}, and \colorbox{inv}{MI-reconstructed images}. 
In (a), \colorbox{inv}{$f_{recon}^{NoDef}$} closely resemble \colorbox{priv}{$f_{priv}^{NoDef}$}, consistent with high attack accuracy. 
In (b), \colorbox{priv}{private images} and \colorbox{privRe}{RE-private images} share some similarity but they are not identical, with partial overlap between  \colorbox{priv}{$f_{priv}^{MIDRE}$} and \colorbox{privRe}{$f_{RE}^{MIDRE}$}.
Importantly, \colorbox{inv}{$f_{recon}^{MIDRE}$} closely resembles \colorbox{privRe}{$f_{RE}^{MIDRE}$} as RE-private is the training data for MIDRE. This results in \textbf{a reduced overlap between \colorbox{inv}{$f_{recon}^{MIDRE}$} and \colorbox{priv}{$f_{priv}^{MIDRE}$}, explaining that MI does not accurately capture the private image features under MIDRE.}  More visualization can be found in Supp.
}
\label{fig:f1}
\vspace{-0.2cm}
\end{figure}

\textbf{Features of MI-reconstructed images tend to match  features of training data.}
SOTA MI attacks aim to reconstruct images that maximize the likelihood under the target model
(Eq. \ref{eq:L_MI}) in order to extract training data (which possess a high likelihood under the target model). Under attacks of high accuracy,
\begin{wrapfigure}{r}{0.4\textwidth}
    \centering
    \includegraphics[width=1.0\linewidth]{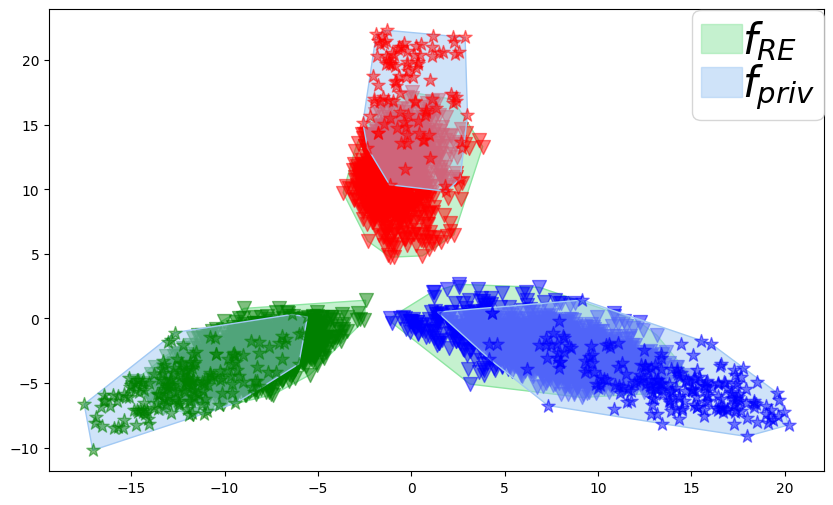}
    \caption{\textbf{MIDRE target model achieves high accuracy despite partial overlap of \colorbox{privRe}{$f_{RE}^{MIDRE}$} and \colorbox{priv}{$f_{priv}^{MIDRE}$}}. We visualize the penultimate layer activations of \colorbox{privRe}{RE-private images} and  \colorbox{priv}{private images}  for three identities. While $f_{RE}^{MIDRE}$ and $f_{priv}^{MIDRE}$ do not completely overlap, the model can still classify private images with high accuracy. 
This is because the private images remain distinct from other classes and distant from other classification regions, even when their representations are partially shared with  RE-private images (the training data).
We remark that RE randomly erases different regions from the images in different iterations, preventing the model to learn shortcut features and forcing the model to learn intrinsic features and become more generalizable beyond training data.  More visualization can be found in Supp.
}
\label{fig:midre_acc}
\vspace{-1cm}
\end{wrapfigure}
$f_{recon}$ tends to match the features of training data $f_{train}$ \citep{nguyen2023re}. 

\textbf{Evidence.} In Fig. \ref{fig:f1} (a), as the training data of NoDef is private images $f_{train}^{NoDef} = f_{priv}^{NoDef}$, we observe that in NoDef model, $f_{recon}^{NoDef}$ overlaps $f_{priv}^{NoDef}$, i.e. there is significant overlap between the pink and blue polygons.
In Fig. \ref{fig:f1} (b), the MIDRE model is trained with RE-private images $f_{train}^{MIDRE} = f_{RE}^{MIDRE}$, and as a result, pink polygon $f_{recon}^{MIDRE}$ and green polygon $f_{RE}^{MIDRE}$ overlap. 
This confirms \textbf{features of reconstructed images tend to match the features of training data}, i.e. $f_{priv}^{NoDef}$ in NoDef and $f_{RE}^{MIDRE}$ in MIDRE.

\textbf{Mismatch in feature space of MIDRE.} MIDRE is trained using RE-private images and is generalizable to images without RE as shown in
\citep{zhong2020random}. 
Under MIDRE target model, 
$f_{RE}^{MIDRE}$ and $f_{priv}^{MIDRE}$ have partial overlaps, but they  are not identical. Meanwhile, $f_{recon}^{MIDRE}$ tend to match with $f_{RE}^{MIDRE}$ (RE-private images are training data for MIDRE, and follows the  discussion above).  
Therefore, $f_{recon}^{MIDRE}$ do not replicate $f_{priv}^{MIDRE}$, significantly degrading the MI attack.

\textbf{Evidence.} 
In Fig. \ref{fig:f1} (b), $f_{RE}^{MIDRE}$ and $f_{priv}^{MIDRE}$ are partial overlap.
Importantly, $f_{recon}^{MIDRE}$, which overlaps with
$f_{RE}^{MIDRE}$ as explained above,
only partially overlaps with $f_{priv}^{MIDRE}$, suggesting MI attacks fail to guide the reconstructed features to replicate private features. Consequently, \textbf{MIDRE introduces a discrepancy between MI-reconstructed and private images in feature space of the target model, degrading the attack accuracy.}

\textbf{Remark.} 
Note that the mismatch between $f_{RE}^{MIDRE}$ and $f_{priv}^{MIDRE}$ does not cause the reduction of model utility (see Fig. \ref{fig:midre_acc}). This is because the private images remain distinct from other classes and distant from other classification regions, even when their representations are partially overlapped with  RE-private images (the training data).

\subsection{Importance of partial erasure and random location for privacy-utility trade-off}
\label{sec:RE_freq}


In  addition to two properties discussed in Sec. \ref{sec:feature_space} which contribute to outstanding effectiveness of applying  RE to degrade MI, we analyse in this section two properties of Random Erasing that are: {\color{blue}\textbf{Property P2:}} \textbf{Partial Erasure}, and {\color{blue}\textbf{Property P3:}} \textbf{Random Location}. To investigate the effect of each property, we conduct the experiment using the following setup: 
We use $T$ = ResNet-18 \citep{simonyan2014very}, $D_{priv}$ = Facecrub \citep{ng2014data}, $D_{pub}$ = FFHQ \citep{karras2019style}, attack method = PPA \citep{struppek2022plug}. 
To evaluate the effectiveness of \textbf{Partial Erasure} and \textbf{Random Location}, we conduct experiments on three schemes: \textbf{Entire Erasing (EE), Fixed Erasing (FE)}, and \textbf{Random Erasing (RE)}. These schemes are compared against a No Defense baseline, which is trained for 100 epochs without any defense.
In the Entire Erasing (EE) scheme, we randomly reduce the number of erased samples per ID in each training epoch to simulate varying pixel concealment levels. Specifically, we train the model for 100 epochs, and for each epoch, we erase randomly 50\%, 40\%, 30\%, 20\%, 10\%, and 0\% of images per ID.
For  Fixed Erasing (FE), a fixed location within each image is erased throughout the entire training process. However, the erased location varied across different images. For Random Erasing (RE), the location of erased areas is randomly selected for each image and training iteration. We train the RE model for 100 epochs with different values of the erasure ratio, $a_e$ = 0.5, 0.4, 0.3, 0.2, 0.1 corresponding to 50\%, 40\%, 30 \%, 20\%, and 10\% pixel concealment, respectively.


\begin{table}[ht!]
\centering
\caption{We compare three different techniques for pixel concealment, to reduce the amount of private information presented to the model during training.
The results show that simply reducing epochs as in ``Entire Erasure'' is insufficient for degrading attack performance. Meanwhile, RE improves model utility while degrading attack accuracy effectively.}
\begin{adjustbox}{width=0.85\columnwidth,center}
\begin{tabular}{cccccccc}
\hline
\multirow{3}{*}{\textbf{Concealment}} & \multicolumn{5}{c}{\textbf{Partial Erasure}} &  \multicolumn{2}{c}{\multirow{2}{*}{\textbf{Entire Erasing}}} \\  \cmidrule{2-6} 

 & & \multicolumn{2}{c}{\textbf{Random Erasing}} & \multicolumn{2}{c}{\textbf{Fixed Erasing}} &  \\  \cmidrule{3-4} \cmidrule(l{2pt}r{2pt}){5-6}  \cmidrule(l{2pt}r{2pt}){7-8}

\multicolumn{1}{c}{} & \multicolumn{1}{c}{\textbf{$a_e$}} & \textbf{Acc ($\uparrow$)} & \textbf{AttAcc ($\downarrow$)} & \textbf{Acc ($\uparrow$)} & \textbf{AttAcc ($\downarrow$)} & \textbf{Acc ($\uparrow$)} & \textbf{AttAcc ($\downarrow$)} \\ \hline

0\%  & 0 & 97.69 & 87.12 & 97.69 & 87.12 & 97.69 & 87.12 \\
10\% & 0.1 & \textbf{97.91} &  \textbf{79.76}  & 96.13 & 85.26 &  97.33 & 87.83 \\
20\% & 0.2 &  \textbf{97.64} &  \textbf{66.32}  & 96.79 &  69.81 & 97.52 & 87.03 \\
30\%  & 0.3 & 97.14 & \textbf{46.30} & 96.13 & 50.71 & \textbf{97.53} & 89.15 \\
40\%  & 0.4 & 96.05 & \textbf{27.75} & 93.10 & 28.49  & \textbf{97.30} & 89.03 \\
50\%  & 0.5 & 93.77 & 15.98 &  86.69 & \textbf{14.86} & \textbf{97.21} & 87.59 \\ \hline

\end{tabular}
\end{adjustbox}
\label{tab:reduc_epoch}
\end{table}

\textbf{{\color{blue}\textbf{Property P2}} brings the privacy effect to defend against MI attacks}. By erasing portions of training images, it reduces the amount of private information exposed to the model during training. By erasing more information, we can effectively degrade the accuracy of privacy attacks. Importantly, partial erasures prevent the model from seeing \textbf{entire images}. 
Consequently, RE-images provide less  information about the entirety of the face, such as inter-pupillary distance, relative distances between the eyes, nose, and mouth, the position of the cheekbones, etc.
 Note that, features need to be presented to the model many times during training in order for the model to learn the features, as suggested by the common practice of using multiple epochs to train a robust model. \textbf{Property P2} reduces the frequency of presenting the features 
to the model during training. 
Such reduced frequency using partial erasure makes it more difficult for the model to memorize 
 the identity features.

\textbf{Evidence.} In Tab. \ref{tab:reduc_epoch}, 
in terms of degrading the attack, 
partial erase (fixed or random) is more effective than entire erase (reduce numbers of sample per ID) although
the percentages of pixel concealment are the same.
Specifically, EE  (reduce 50\% images per ID) is significantly more vulnerable to attacks than RE and FE (50\% image areas are erased,  trained in 100 epochs), suffering approximately 71\% higher in attack accuracy.

\textbf{{\color{blue}\textbf{Property P3}} recovers the model utility}. 
While information reduction can improve privacy, it may also negatively impact model utility if too much information is erased. Fixing the erasing location for an image means some identity feature of this image will never be presented to the model, model may  not have adequate information to learn effectively. RE avoids this issue. As the location of erased area is changed in each training iteration, RE improves the diversity of the training data and ensures that the model still observes a significant portion of the image.

\textbf{Evidence.} In Tab. \ref{tab:reduc_epoch}, RE improves the model accuracy while maintains the same attack accuracy as FE in different erased portion ratio $a_e$. For instance, RE
has higher model accuracy than FE by 7.08\% with $a_e$ = 0.5. 
With $a_e$ = 0.3 and 0.4, RE has higher accuracy and lower attack accuracy than EE model, showing that privacy effect of RE.

\section{Experiments}
\label{sec:experiment}
\subsection{Experimental Setting}\label{Experimental_Setting}

To demonstrate the generalisation of our proposed MI defense, we carry out multiple experiments using different SOTA MI attacks on various architectures. In addition, we also use different setups for public and private data. The summary of all experiment setups is shown in Tab. \ref{tab:setup}. In total, we conducted 37 experiment setups to demonstrate the effectiveness of our proposed defense MIDRE.

\textbf{Dataset}: We follow the same setups as SOTA attacks \citep{zhang2020secret,nguyen2023re,struppek2022plug} and defense \citep{peng2022bilateral, struppekcareful, ho2024model} to conduct the experiments on four datasets including: CelebA \citep{liu2015deep}, FaceScrub \citep{ng2014data}, VGGFace2 \citep{cao2018vggface2}, and Stanford Dogs \citep{dataset2011novel}. We use FFHQ \citep{karras2019style} and AFHQ Dogs \citep{choi2020stargan} for the public dataset. 
We strictly follow \citep{zhang2020secret,nguyen2023re,struppek2022plug,an2022mirror,peng2022bilateral,struppekcareful,ho2024model,koh2024vulnerability} to divide the datasets into public and private set. See Supp for the details of datasets.

\begin{table}[t!]
\centering
\caption{
In total, we conduct 37 experiment setups to demonstrate the effectiveness of MIDRE.}
\begin{adjustbox}{width=1.0\columnwidth,center}
\begin{tabular}{lcccl}
\hline
Attack & $T$ & $\D_{priv}$ & $\D_{pub}$ & Resolution \\  \hline
GMI \citep{zhang2020secret} & \multirow{6}{*}{\begin{tabular}[c]{@{}l@{}}VGG16 \citep{simonyan2014very}\\ IR152 \citep{he2016deep}\\ FaceNet64 \citep{cheng2017know}\end{tabular}} & \multirow{6}{*}{CelebA} & \multirow{6}{*}{CelebA/FFHQ} & \multirow{6}{*}{64$\times$64} \\
KedMI \citep{chen2021knowledge} &  &  &  &  \\
LOMMA \citep{nguyen2023re} &  &  &  &  \\
PLGMI \citep{yuan2023pseudo} &  &  &  &  \\
RLBMI \citep{han2023reinforcement} &  &  &  &  \\
BREPMI \citep{kahla2022label} &  &  &  &  \\ \hline
\multirow{7}{*}{PPA \citep{struppek2022plug}} & ResNet18 \citep{he2016deep} & \multirow{6}{*}{Facescrub} & \multirow{6}{*}{FFHQ} & \multirow{8}{*}{224$\times$224} \\
 & ResNet101 \citep{he2016deep} &  &  &  \\
 & ResNet152 \citep{he2016deep} &  &  &  \\
 & DenseNet121 \citep{huang2017densely} &  &  &  \\
 & DenseNet169 \citep{huang2017densely} &  &  &  \\
 & MaxVIT \citep{tu2022maxvit} &  &  &  \\ \cline{2-4}
 & ResneSt101 & Stanford Dogs & AFHQ Dogs &  \\ \hline
 \multirow{2}{*}{MIRROR \citep{an2022mirror}}
 & Inception-V1 \citep{Inceptionv1} & \multirow{2}{*}{VGGFace2} & \multirow{2}{*}{FFHQ} &  160$\times$160\\
  & ResNet50 \citep{he2016deep}&   &   &  224$\times$224 \\ \hline 
 \multirow{2}{*}{IF-GMI \citep{qiu2024closer}}
 & ResNet18  & \multirow{2}{*}{Facescrub} & \multirow{2}{*}{FFHQ} &  \multirow{2}{*}{224$\times$224}\\
  & ResNet152  &   &   &   \\ \hline 
 \end{tabular} 
\end{adjustbox}
\label{tab:setup}
\vspace{-0.2cm}
\end{table}

\textbf{Model Inversion Attacks.}  To evaluate the effectiveness of our proposed defense MIDRE, we employ a comprehensive suite of state-of-the-art MI attacks. This includes various attack categories: white-box and label-only, one type of black-box attack. To assess robustness at high resolutions, we employ PPA \citep{struppek2022plug} and IF-GMI \citep{qiu2024closer} against attacks targeting 224$\times$224 pixels and MIRROR \citep{an2022mirror} against attacks targeting 224$\times$224 and 160$\times$160 pixels.
For low resolution 64$\times$64 pixels, we leverage four SOTA white-box attacks: GMI \citep{zhang2020secret}, KedMI \citep{chen2021knowledge}, PLG-MI \citep{yuan2023pseudo}, and LOMMA \citep{nguyen2023re} (including LOMMA+GMI and LOMMA+KedMI). 
Additionally, we incorporate RLBMI \citep{han2023reinforcement} for black-box attack and 
BREPMI \citep{kahla2022label} for label-only attack. 
We strictly replicate the experimental setups in \citep{zhang2020secret,chen2021knowledge,yuan2023pseudo,nguyen2023re,struppek2022plug,peng2022bilateral,han2023reinforcement,an2022mirror} to ensure a fair comparison between NoDef (the baseline model without defense), existing state-of-the-art defenses, and our proposed method, MIDRE.

\textbf{Target Models.} 
We follow other MI research \citep{zhang2020secret,nguyen2023re,struppek2022plug,peng2022bilateral} to train defense models.
We use 11 architectures for the target model to assess its resistance to MI attacks using various experimental configurations. The details are summarized in Tab. \ref{tab:setup}.
We train target models with the same hyper-parameter ($a_h$) for all low-resolution data set-ups. In addition, for high-resolution data, we use two value for hyper-parameter $a_h=0.4$ and $a_h=0.8$ across all setups. This allows us to demonstrate MIDRE's effectiveness in achieving the optimal trade-off between utility and privacy with consistent hyper-parameter.

\textbf{Comparison Methods.}
We compare the performance of our model against no defending method (NoDef) and five defense methods, including NLS (Negative Label Smoothing)\citep{struppekcareful}, TL-DMI \citep{ho2024model}, MI-RAD (MI-resilient architecture designs) \citep{koh2024vulnerability}, BiDO \citep{peng2022bilateral}, and MID \citep{wang2021improving}. As for MI-RAD \citep{koh2024vulnerability}, we compare our results to Removal of Last Stage
Skip-Connection (RoLSS), Skip-Connection Scaling Factor (SSF),  Two-Stage Training Scheme (TTS).

We establish a baseline (NoDef) by training the target model from scratch without incorporating any MI defense strategy. 
According to NLS, TL-DMI, MI-RAD, we follow their setup and evaluation to compare with MIDRE.
We then carefully tune the hyper-parameters of each method to achieve optimal performance.

\textbf{Evaluation Metrics.}
MI defenses typically involve a trade-off between the model's utility and its resistance 

\begin{minipage}[ht!]{\textwidth}
\begin{minipage}[t]{1.0\textwidth}
\centering
\includegraphics[width=0.27\textwidth]{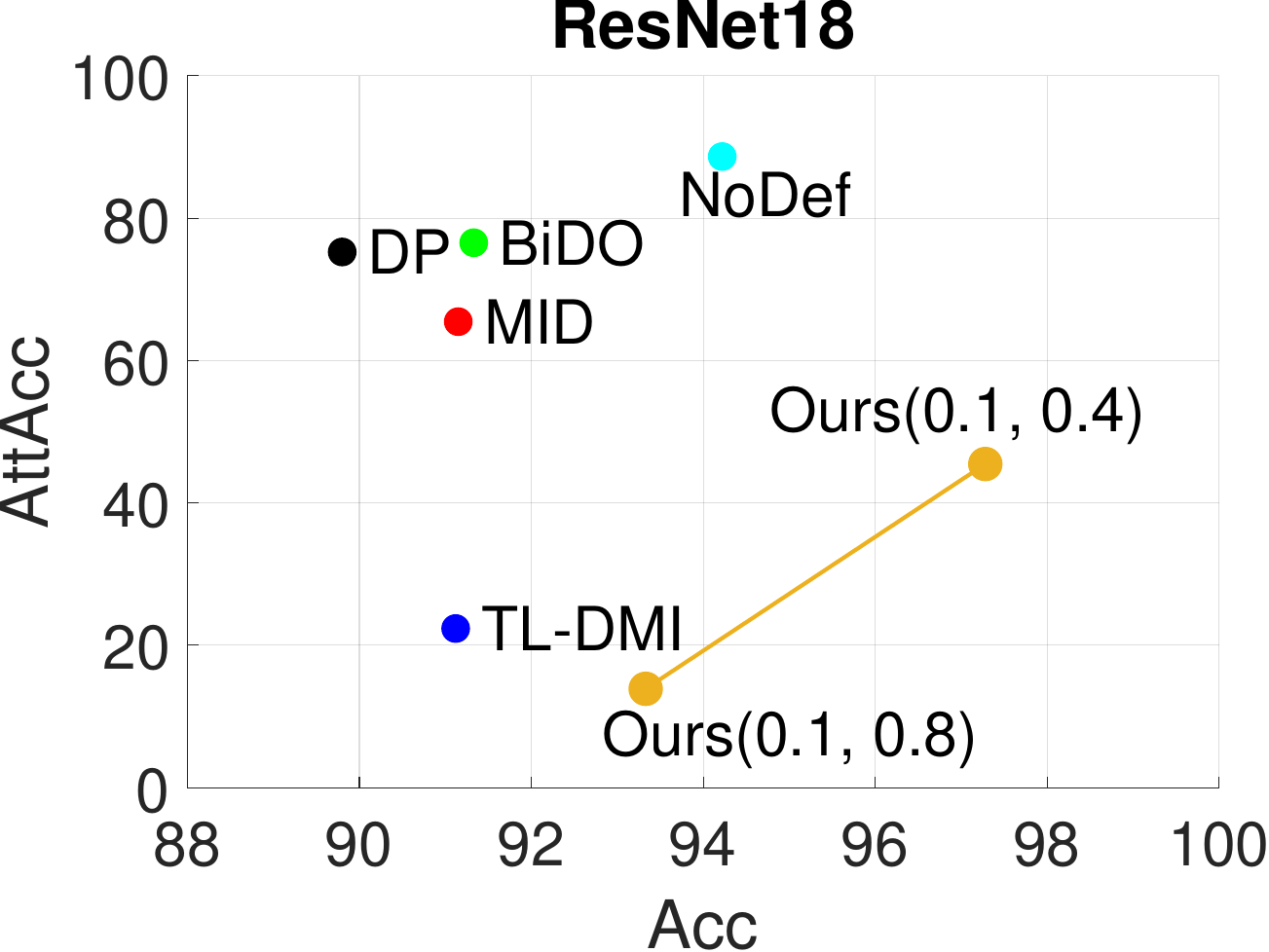} \hspace{0.3cm}
\includegraphics[width=0.27\textwidth]{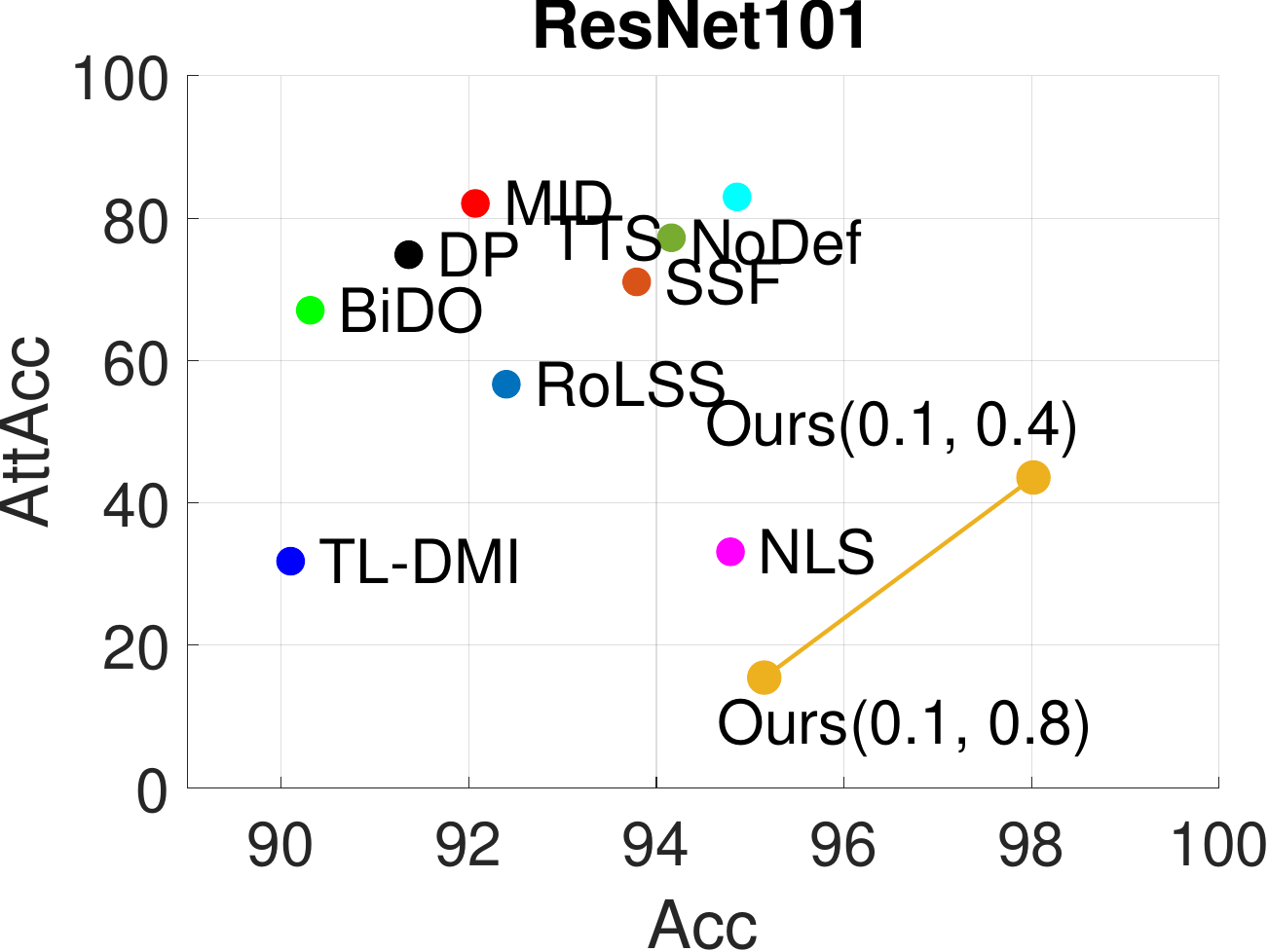} \hspace{0.3cm}
\includegraphics[width=0.27\textwidth]{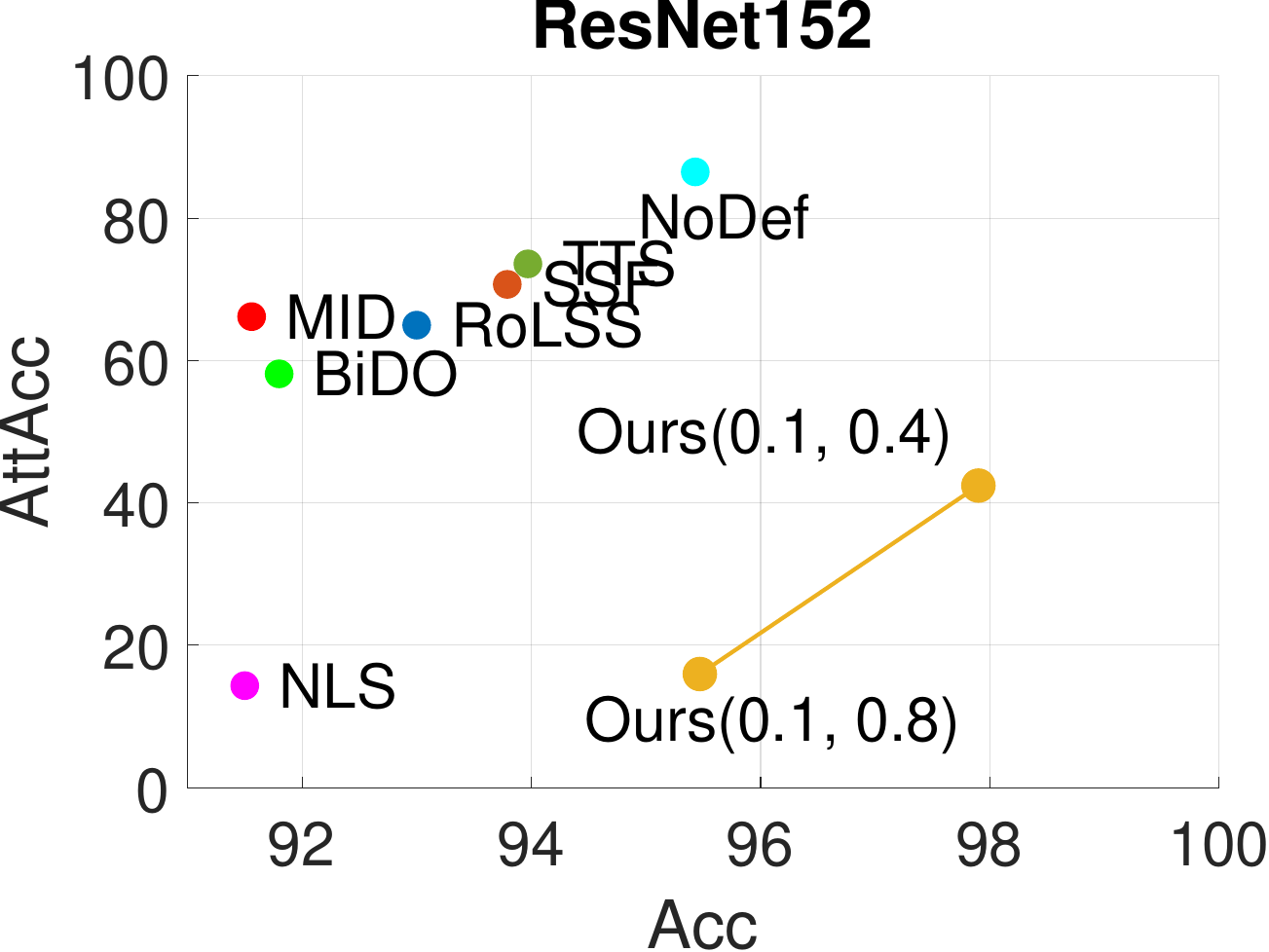}\hfill

\includegraphics[width=0.27\textwidth]{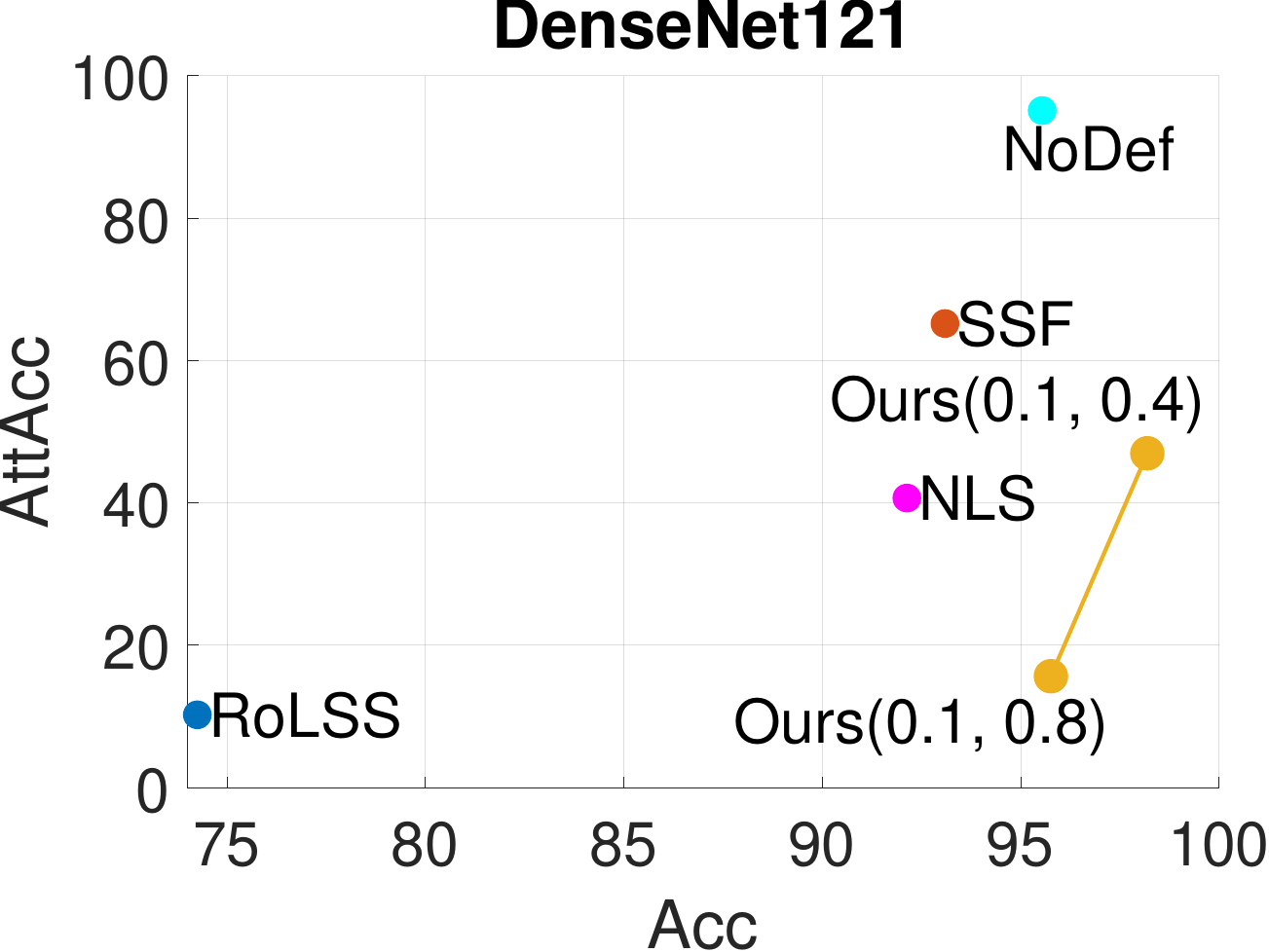}
\hspace{0.3cm}
\includegraphics[width=0.27\textwidth]{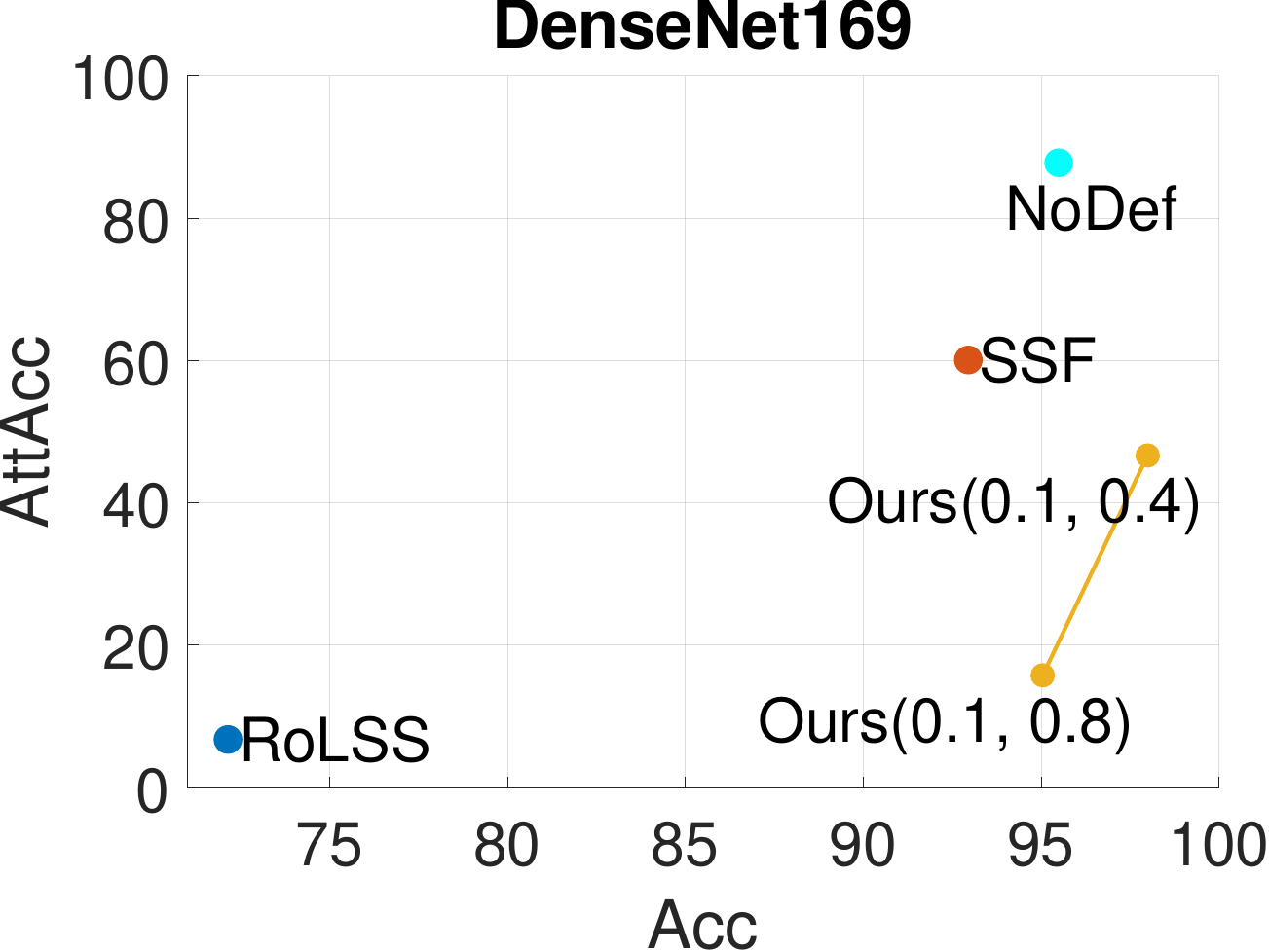}
\hspace{0.3cm}
\includegraphics[width=0.27\textwidth]{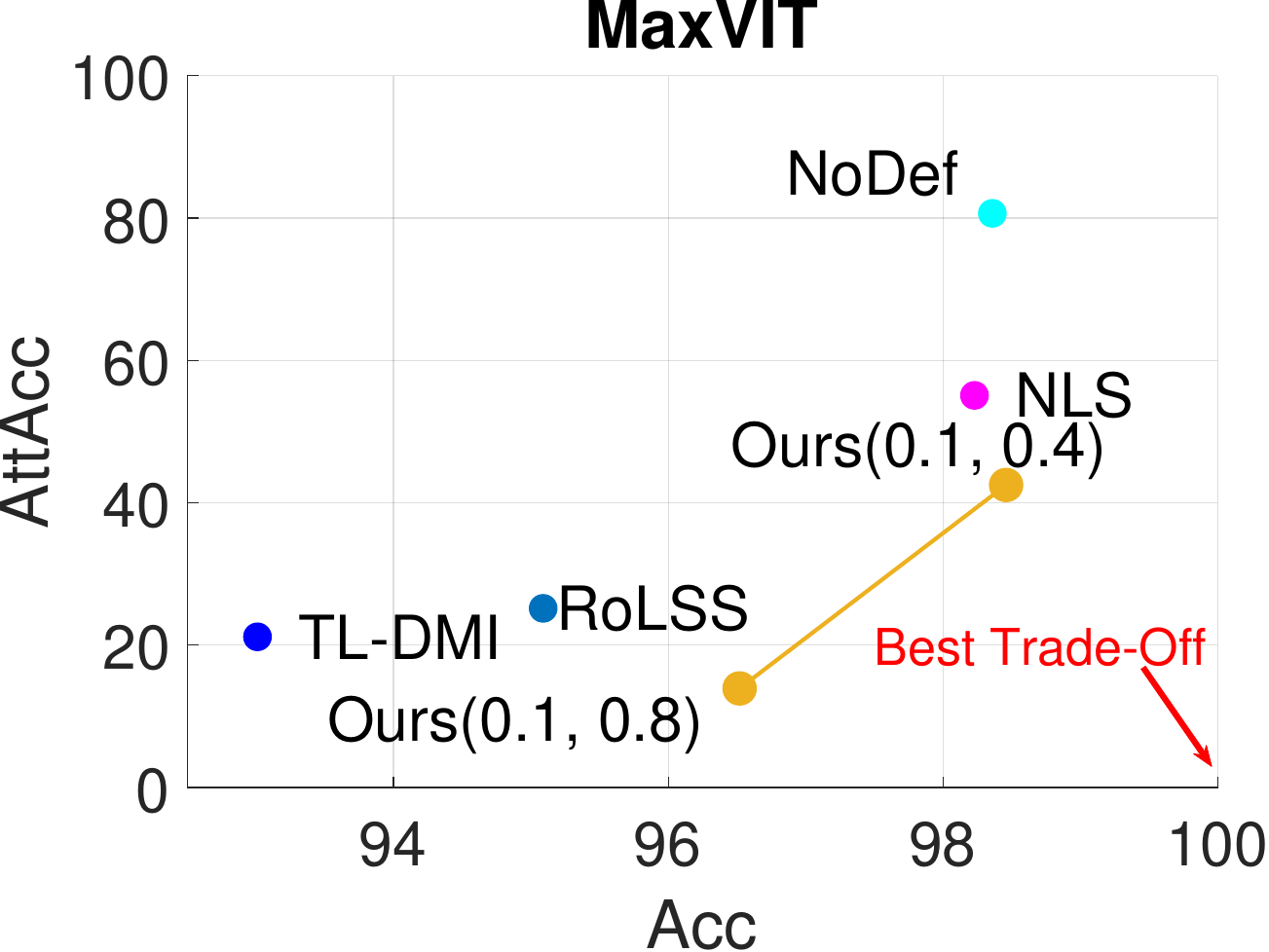}
\captionof{figure}{We evaluate PPA attack \citep{struppek2022plug} on our proposed method, NoDef, MID \citep{wang2021improving}, BiDO \citep{peng2022bilateral}, NLS \citep{struppekcareful}, and TL-DMI \citep{ho2024model}. Target models are trained on $\D_{priv}$ = Facescrub with 6 architectures.
The results show that our method archives the best trade-of between utility and privacy among state-of-the-art defenses.}
\label{fig:PPA_results}
\vspace{0.4cm}
\end{minipage}

\begin{minipage}[t]{0.49\textwidth}
    \centering
    \includegraphics[width=0.48\textwidth]{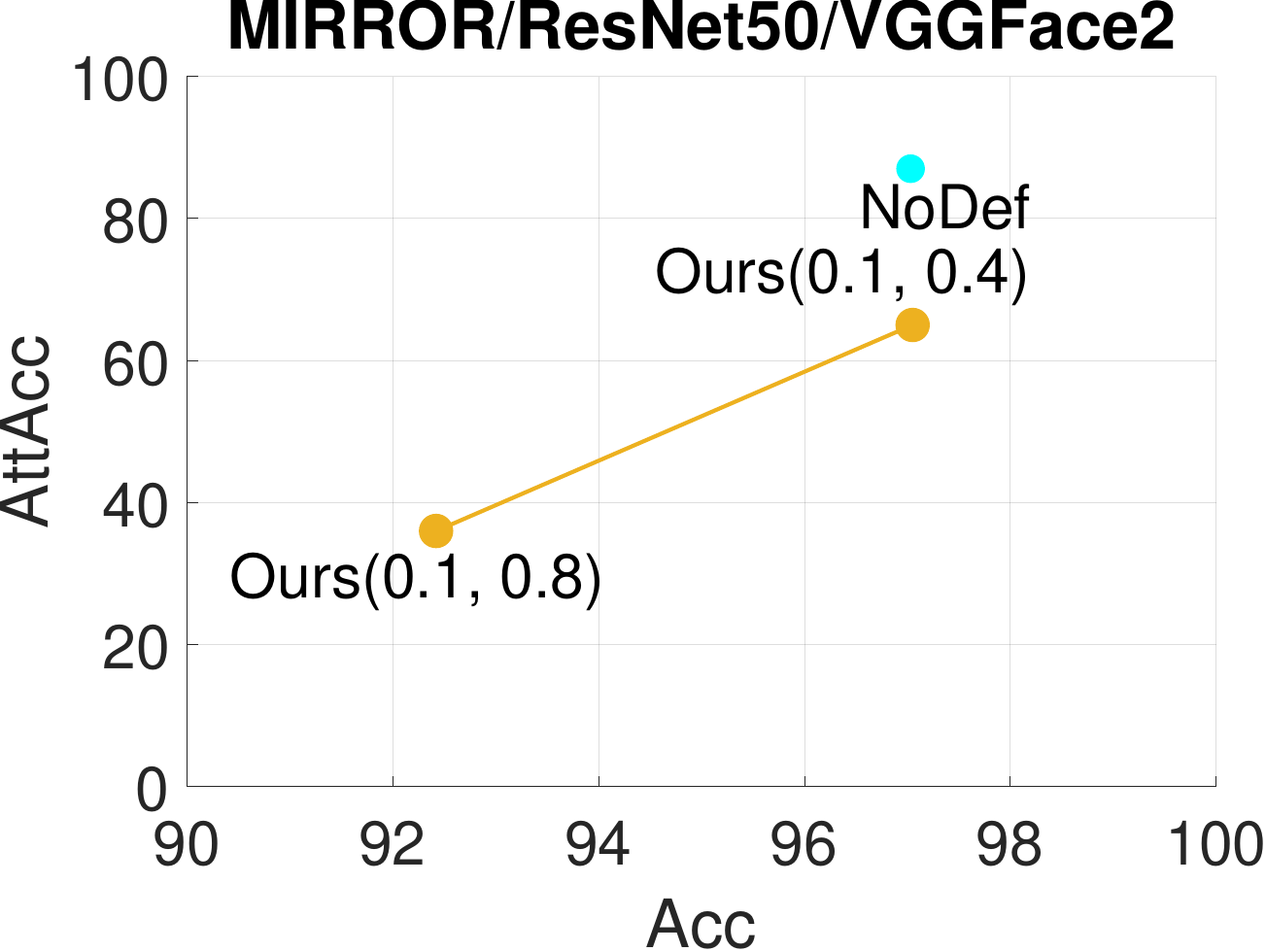}
    \includegraphics[width=0.48\textwidth]{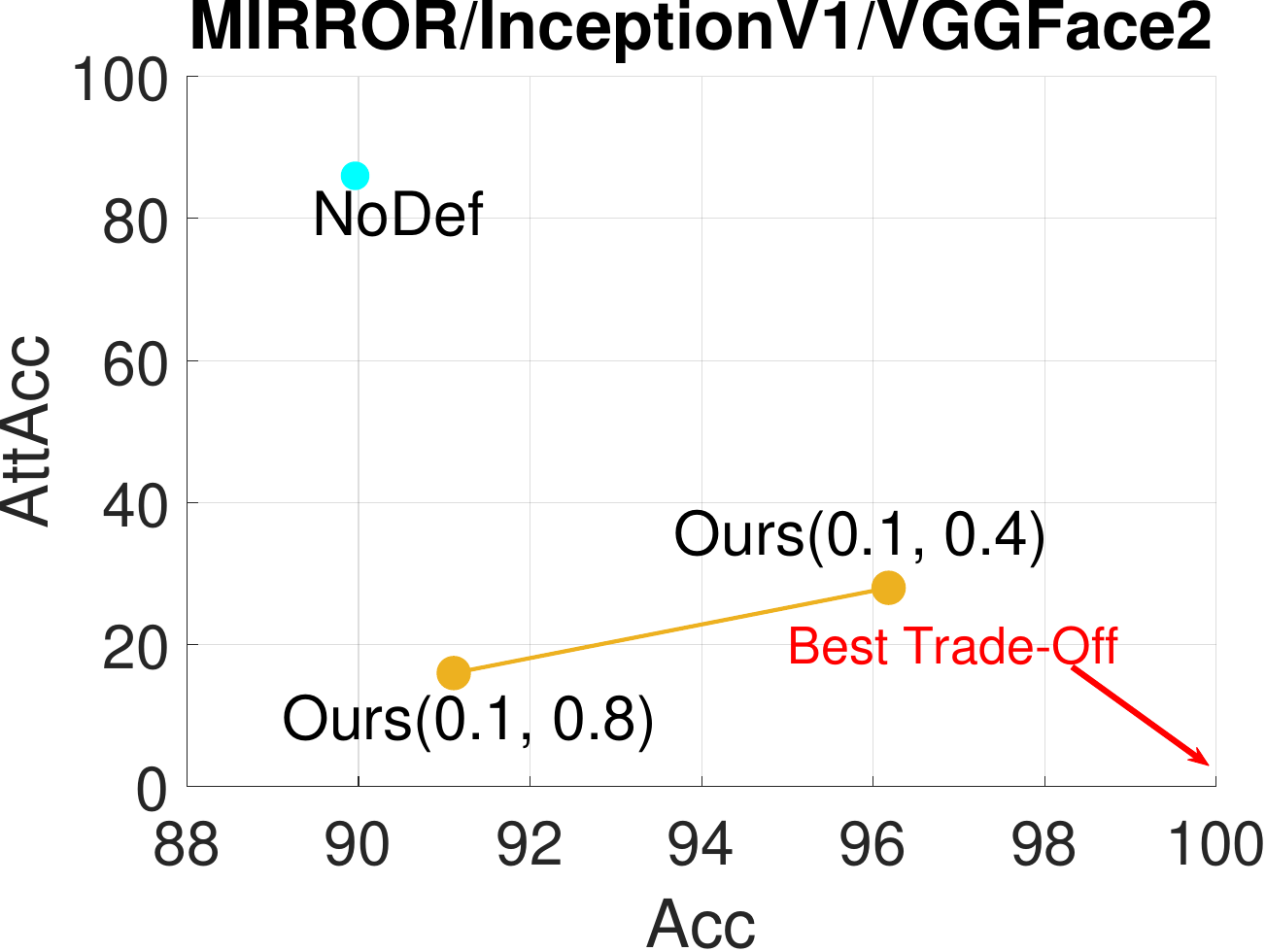}
    \captionof{figure}{We evaluate MIRROR attack \citep{an2022mirror} on VGGFace2 dataset.
    The results show that our method archives the best trade-of between utility and privacy than NoDef model.
    }
    \label{fig:PPA_MIRROR_results}
\end{minipage}
\hfill
\begin{minipage}[t]{0.49\textwidth}
    \centering
    \includegraphics[width=0.48\textwidth]{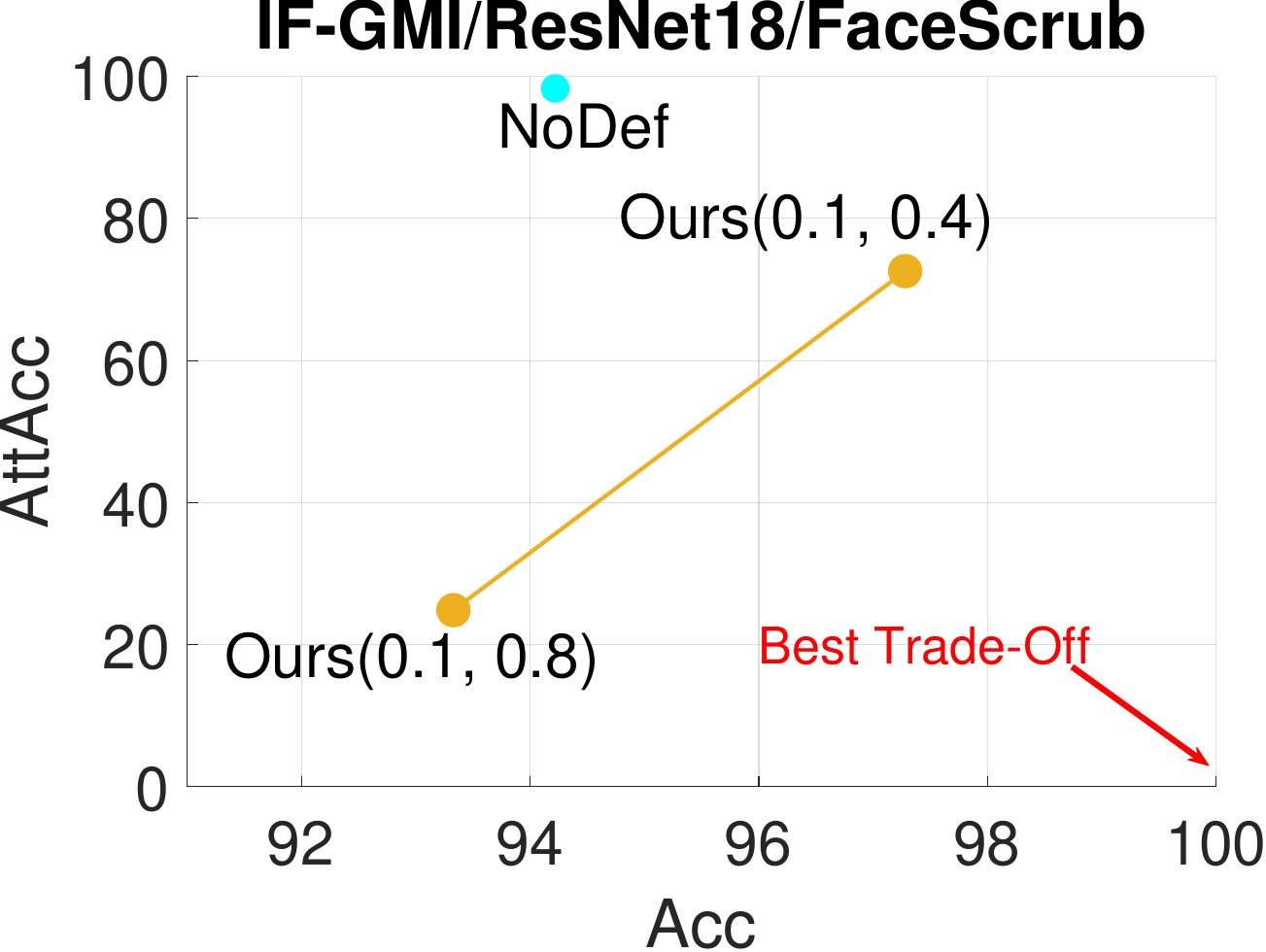}
    \includegraphics[width=0.48\textwidth]{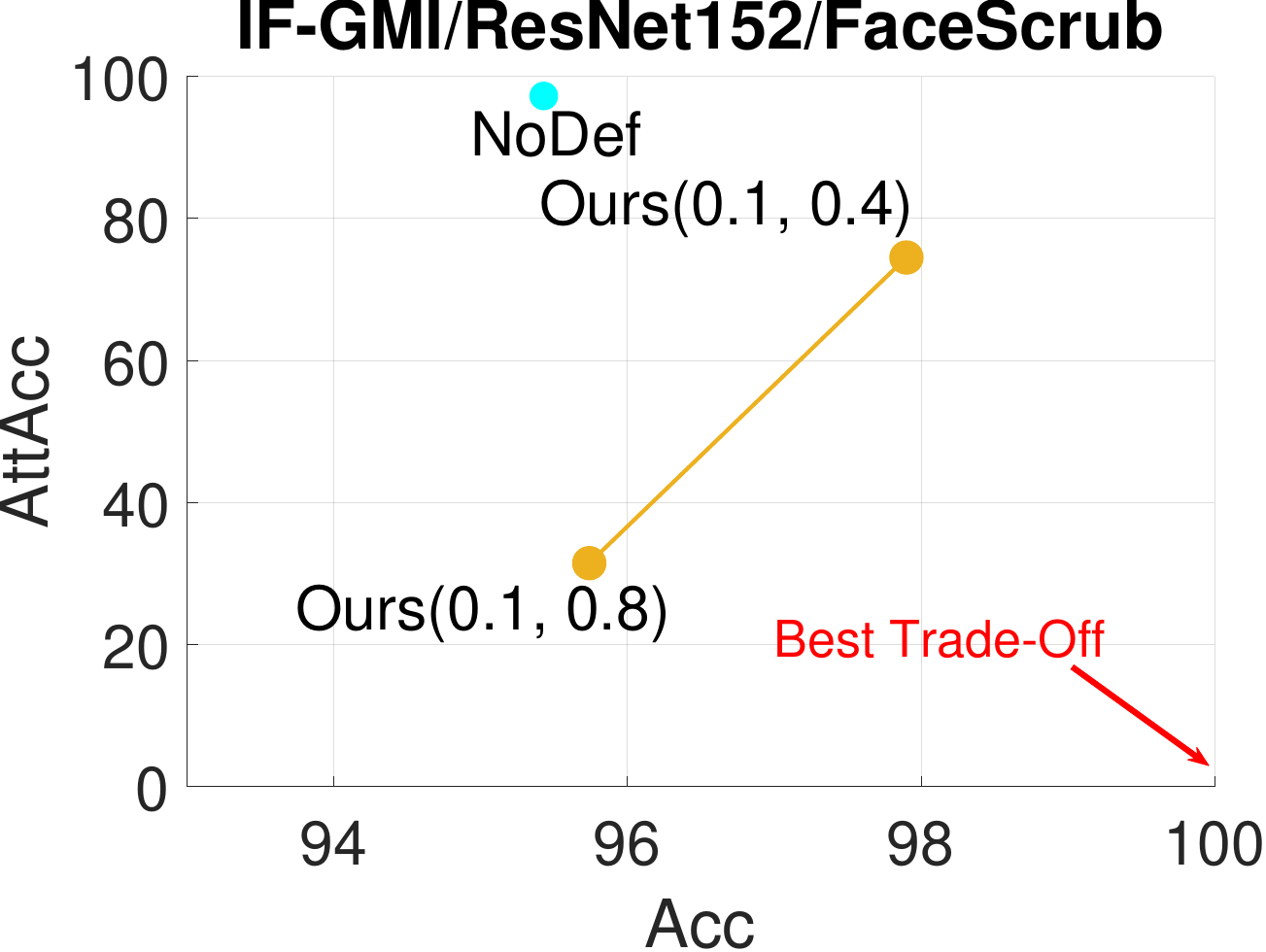}
    \captionof{figure}{Results of IF-GMI \citep{qiu2024closer} attack on Facescrub dataset. Here, we use $T$ = ResNet18/ ResNet152, $\D_{priv}$ = Facescrub, $\D_{pub}$ = FFHQ, image resolution = 224$\times$224 images.
    }
    \label{fig:IF_GMI_results}
\end{minipage}
\vspace{0.6cm}
\end{minipage}

to model inversion attacks. In this section, we evaluate these defenses using two key metrics: Natural Accuracy (Acc $\uparrow$) to evaluate the model utility and Attack accuracy (AttAcc $\downarrow$) and to evaluate the model privacy.
We further show other evaluation metrics, including K-Nearest Neighbor Distance, $\delta_{eval}$, $\delta_{face}$ \citep{struppek2022plug},
complement these results with qualitative results and a user study in Supp.

\subsection{Comparison against SOTA MI Defenses}

We compare the model accuracy and attack accuracy of defense models in 6 architectures using attack method PPA \citep{struppek2022plug} in  Fig.~\ref{fig:PPA_results}. All the target models are trained on Facescrub dataset.
Interestingly, we are the first to observe that our defense models achieve higher natural accuracy but lower attack acuracy than no defense model for larger image sizes (224$\times$224).
With small masking areas (Ours(0.1,0.4)), our proposed method consistently achieves the lowest attack accuracy among all defense models while its natural accuracy is higher than NoDef, BiDO, MID, and DP models. For example, using ResNet101, our model reduces attack accuracy by 39.42\% compared to NoDef while achieving the model accuracy is higher than NoDef model 3.16\%. 
MaxVIT, a recent advanced architecture, has very high attack accuracy (80.66\%). Our defense mechanism significantly enhances its robustness, lowering attack accuracy to 42.5\% without compromising model performance.
By increasing the masking areas (Ours(0.1,0.8)), they achieve a significant reduction in attack accuracy while maintaining high natural accuracy, outperforming other strong defense methods like NLS and TL-DMI. Specially, \textit{our attack accuracies are below 20\% for all architectures}.
This represents the best utility-privacy trade-off among all evaluated defense models, demonstrating our method's effectiveness in mitigating model inversion attacks.

We further show the effectiveness of our proposed method using two attacks MIRROR \citep{an2022mirror} and IF-GMI \citep{qiu2024closer}.
Regarding the MIRROR attack, we compare the results of MIDRE and the NoDef model using $\D_{priv}$ = VGGFace2 (see Fig. \ref{fig:PPA_MIRROR_results}). Our defense reduces the attack accuracy by 22\% and 70\% without harming model utility, where the target model $T$ = ResNet50/InceptionV1. 
Results of IF-GMI attack are shown in Fig. \ref{fig:IF_GMI_results}.
The results show that MIDRE reduces the attack accuracy by more than 22\%.


\begin{figure}
    \centering    
    \includegraphics[width=0.85\textwidth]{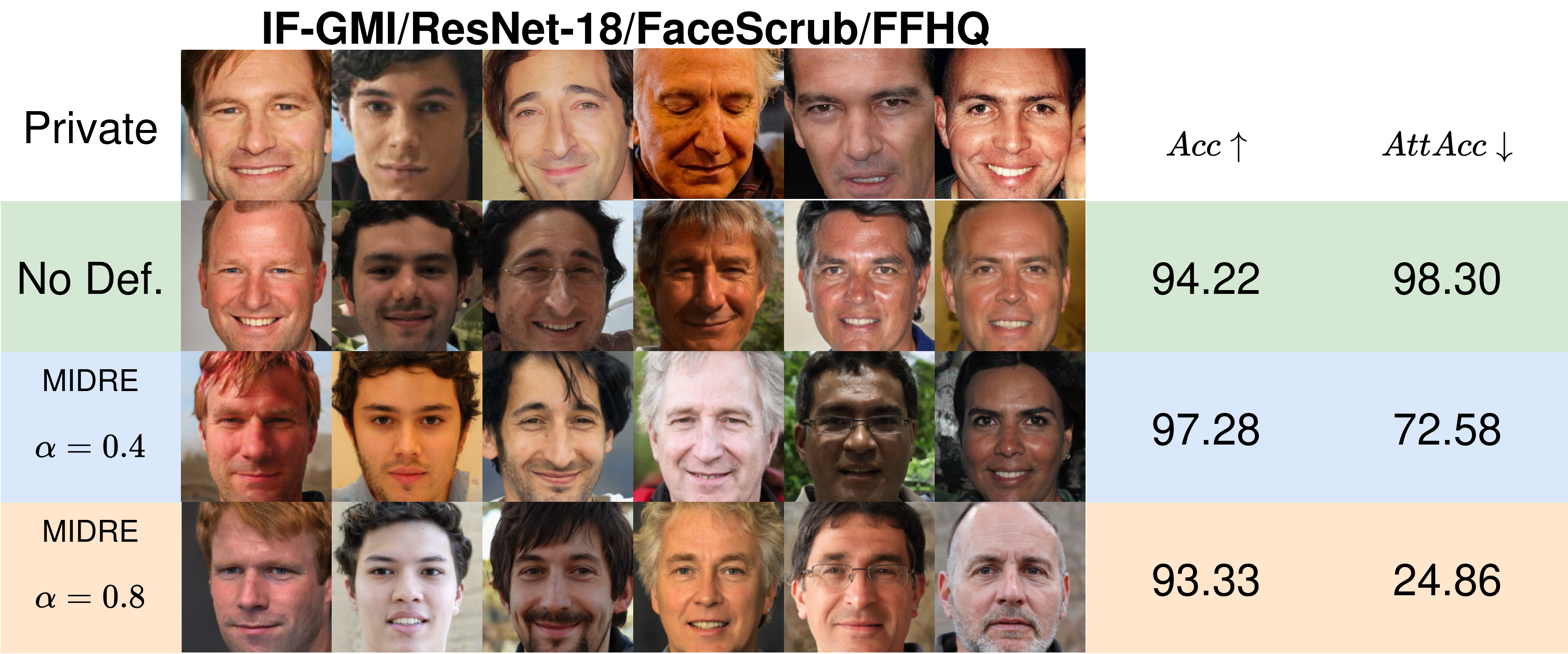} 
    \caption{Reconstructed image obtained from IF-GMI attack with $T$ = ResNet-18, $\D_{priv}$ = Facescrub, $\D_{pub}$ = FFHQ. 
    Overall, our reconstructed images have less similarity to private images compared to those from the no-defense model, suggesting the efficiency of our proposed defense MIDRE.}
\label{fig:Attacking_Samples_Visual_IFGMI_R18}
\end{figure}

\begin{wraptable}{r}{0.5\textwidth}
\vspace{-0.5cm}
\centering
\caption{We evaluate multiple SOTA MI attacks on 64$\times$64 images, comparing their performance under NoDef, NLS, and our MIDRE defense.
$T$ = VGG16, $D_{priv}$ = CelebA, $D_{pub}$ = CelebA.
}
\begin{adjustbox}{width=0.5\textwidth,center}
\begin{tabular}{clccc}
\hline
Attack &  Defense & Acc $\uparrow$ & AttAcc $\downarrow$ &  $\Delta$ $\uparrow$  \\ 
\hline
\multirow{3}{*}{LOMMA + GMI} & NoDef  & 85.74 & 53.64 $\pm$ 4.64 &  -   \\
 &  NLS  & 80.02 & 39.16 $\pm$ 4.25 & 2.53 \\
 &  \textbf{MIDRE} & 79.85  & \textbf{26.62 $\pm$ 1.93} & \textbf{4.59}  \\ 
 \hline
\multirow{3}{*}{LOMMA + KedMI} & NoDef & 85.74 & 72.96 $\pm$ 1.92 & -  \\
 &  NLS  & 80.02 & 63.60 $\pm$ 1.37 & 1.64\\
 &  \textbf{MIDRE} & 79.85  & \textbf{41.82 $\pm$ 1.24 } & \textbf{5.29} \\ 
\hline
\multirow{3}{*}{PLGMI} & NoDef  & 85.74 & 71.00 $\pm$ 3.31 & -  \\
 &  NLS & 80.02 & 72.00 $\pm$ 2.50 & -0.17 \\
 &  \textbf{MIDRE} & 79.85  & \textbf{66.60 $\pm$ 2.94 } & \textbf{0.75} \\ 
\hline
\end{tabular}
\end{adjustbox}
\label{tab:MIA64_NLS_VGG16}
\vspace{-0.7cm}
\end{wraptable}
For low resolution attacks, we evaluate against four  MI attacks, including GMI \citep{zhang2020secret}, KedMI \citep{chen2021knowledge}, LOMMA \citep{nguyen2023re} with two variances (LOMMA+GMI and LOMMA+KedMI), and PLGMI \citep{yuan2023pseudo}
We follow the same setup and compare with NLS in Tab. \ref{tab:MIA64_NLS_VGG16}. In addition, we also use NoDef baseline in NLS paper to compare and estimate $\Delta$ in Tab. \ref{tab:MIA64_NLS_VGG16}.

Overall, our proposed method, MIDRE, achieves significant improvements in security for 64$\times$64 setups compared to SOTA MI defenses. MIDRE achieves this by demonstrably reducing top-1 attack accuracy while maintaining natural accuracy on par with other leading MI defenses.
More results of other attacks can be found in Supp. 



We show the comparison on qualitative results in Fig. \ref{fig:Attacking_Samples_Visual_IFGMI_R18}. Images are reconstructed using IF-GMI atack \citep{qiu2024closer}, $T$ = ResNet-18, $\D_{priv}$ = Facescrub, $\D_{pub}$ = FFHQ. 
Overall, our reconstructed images have less similarity to private images compared to those from the no-defense model,
demonstrating the effectiveness of our defense.
More reconstructed images are included in Supp and our project page.

The experiment results demonstrate that our defense model has a small impact on model utility while significantly enhancing the model's robustness against state-of-the-art MI attacks. Moreover, we are the first to report a substantial improvement in model utility among all existing defenses.

\subsection{Ablation study on data augmentation type.} 

To evaluate the effectiveness of our method compared to other data augmentation-based defenses, we compare 

\begin{minipage}{\textwidth}
\begin{minipage}[t]{0.475\textwidth}
\centering
\captionof{table}{We report the PPA attack on images with resolution 224$\times$224. 
$T$ = ResNet18, $D_{priv}$ = Facescrub, $D_{pub}$ = FFHQ to target models trained with different data augmentation. MIDRE consistently achieves the best balance between utility and privacy, significantly degrading MI attack performance while maintaining competitive natural accuracy.}
\begin{adjustbox}{width=0.84\columnwidth,center}
\begin{tabular}{lccccc}
\hline
Attack &  Defense & Acc  $\uparrow$  & AttAcc $\downarrow$  \\ 
\hline
\multirow{4}{*}{PPA} & NoDef & 94.22 & 88.67  \\
 & \textbf{MIDRE} & 97.28 & \textbf{48.16} \\
 & CutMix & 98.74 & 67.12 \\
 & Random Cropping & 92.24 & 74.22 \\
 & Gaussian Blur & 97.57 & 87.12 \\ \hline
\end{tabular}
\end{adjustbox}
\label{tab:MIA64_augmentation_ablation}
\end{minipage}
\hfill
\begin{minipage}[t]{0.50\textwidth}
\centering
\captionof{table}{The combination MIDRE with existing defense BiDO and NLS. 
The combine models significantly reduces attack accuracy compared to individual defenses. We denote ``OP" for $\Delta$ if the accuracy of the defense model outperforms that of the NoDef model.}
\vspace{-0.15cm}
\begin{adjustbox}{width=1.0\columnwidth,center}
\begin{tabular}{lllll}
\hline
\textbf{Setup} & \textbf{Defense} & \textbf{Acc ($\uparrow$)} & \textbf{AttAcc ($\downarrow$)} & \textbf{$\Delta (\uparrow)$} \\
\hline
\multirow{4}{*}{\textbf{Setup 1}} & NoDef & 95.43 & 86.51 & - \\
 & NLS & 91.50 & 13.94 & 18.47 \\
 & MIDRE & 95.47 & 15.97 & OP \\
 & MIDRE + NLS & 93.69 & 3.75 & 47.65 \\ \hline
\multirow{4}{*}{\textbf{Setup 2}} & NoDef & 86.90 &  81.80 $\pm$  1.44  & - \\
 & BiDO & 79.85 &  63.00 $\pm$  2.08  & 2.67 \\
 & MIDRE & 79.85 & 43.07 $\pm$  1.99  & 5.49 \\
 & MIDRE + BiDO & 82.15 &  39.00 $\pm$ 1.30 & 9.01 \\ \hline
\end{tabular}
\end{adjustbox}
\label{tab:combination}
\end{minipage}
\vspace{0.7cm}
\end{minipage}

MIDRE with models trained using CutMix, random cropping, and Gaussian blur. These augmentations are commonly employed to enhance model generalization.
Here, we use the setup:  attack method = PPA,  
$T$ = ResNet18, $D_{priv}$ = Facescrub, $D_{pub}$ = FFHQ, image resolution = 224$\times$224.

The results, summarized in Table \ref{tab:MIA64_augmentation_ablation}, indicate that while alternative augmentations provide some level of protection, \textit{MIDRE consistently achieves the best balance between utility and privacy, significantly degrading MI attack performance while maintaining competitive natural accuracy.}


\subsection{Combination with Existing Defenses}
\label{sec:combination}

Since MIDRE improves defense effectiveness from the training data perspective, our proposed method can be combined with other defense mechanism from the training objective perspective such as BiDO \citep{peng2022bilateral} and NLS \citep{struppekcareful}.
We use 2 setups: \textbf{Setup 1}: $T$ = ResNet152, $\D_{priv}$ = Facescrub, $\D_{pub}$ = FFHQ, Attack method = PPA, image size = 224 $\times$ 224. \textbf{Setup 2}: $T$ = VGG16, $\D_{priv}/\D_{pub}$ = CelebA, Attack method = LOMMA + KedMI, image size = 64 $\times$ 64. 
We use $a_e$ = [0.1,0.4] and $a_e$ = [0.1,0.8] for setup 1 and setup 2 to train MIDRE and during attack.

The results (see Tab. \ref{tab:combination}) demonstrate the effectiveness of combining MIDRE with either NLS or BiDO for enhancing defense against MI attacks, as our MIDRE takes a data-centric perspective for defense, complementary to existing defenses.  In both experimental setups, the combination models demonstrate a substantial reduction in attack accuracy compared to using MIDRE or the other defenses independently. 
In particular, in setup 1, the combination of MIDRE and Negative LS achieves a remarkable 3.75\% attack accuracy when using the state-of-the-art PPA attack while preserving model utility. For Setup 2, MIDRE + BiDO improves the natural accuracy of the model by 2.3\% while reducing the attack accuracy by 4.07\% and 24\% compared to MIDRE and BiDO, respectively. 
\textbf{This shows our effectiveness of combining MIDRE and existing defense for a better defense.} The combination ability of MIDRE supports that it examines a distinct aspect of the system by focusing on data input, setting it apart from other existing approaches to defend against model inversion attacks.

\section{Conclusion}

We propose a novel approach to MI Defense via Random Erasing (MIDRE). 
We conduct an analysis to demonstrate that RE possess two crucial properties to degrade MI attack while the impact on model utility is small. Furthermore, our features space analysis shows  that model trained with RE-private images following MIDRE leads to a discrepancy between the features of MI-reconstructed images and that of private images, resulting in reducing of attack accuracy.
Experiments validate that our approach achieves outstanding performance in balancing model privacy and utility.
The results consistently demonstrate the superiority of our method over existing defenses across various MI attacks, network architectures, and attack configurations.
The code and additional results can be found in the Supplementary section.






\section*{Acknowledgements}
This research is supported by Temasek Laboratories, Singapore University of Technology and Design.
This research is supported by the National Research Foundation, Singapore under its AI Singapore Programmes (AISG Award No.: AISG2-TC-2022-007); The Agency for Science, Technology and Research (A*STAR) under its MTC Programmatic Funds (Grant No. M23L7b0021). This research
is supported by the National Research Foundation, Singapore and Infocomm Media Development Authority under its Trust Tech Funding Initiative. Any opinions, findings and conclusions or recommendations expressed in this material are those of the author(s) and do not reflect the views of National Research Foundation, Singapore and Infocomm Media Development Authority.
This research is also part-funded by the European Union (Horizon Europe 2021-2027 Framework Programme Grant Agreement number 10107245. Views and opinions expressed are however those of the author(s) only and do not necessarily reflect those of the European Union. The European Union cannot be held responsible for them) and by the Engineering and Physical Sciences Research Council under grant number EP/X029174/1.

{
    \small
    \bibliographystyle{tmlr}
    \bibliography{references}
}

\clearpage
{\Large\textbf{Supplementary material}}

In this supplementary material, we provide additional experiments, analysis, ablation study, and details that are required to reproduce our results. These are not included in the main paper due to space limitations. 

Our code and additional results are available at: \url{https://ngoc-nguyen-0.github.io/MIDRE/}

\unhidefromtoc
\appendix
\renewcommand{\thesection}{\Alph{section}}
\tableofcontents
\renewcommand\thefigure{\thesection.\arabic{figure}}
\renewcommand\theHfigure{\thesection.\arabic{figure}}
\renewcommand\thetable{\thesection.\arabic{table}}  
\renewcommand\theHtable{\thesection.\arabic{table}}  

\setcounter{figure}{0} 
\setcounter{table}{0} 
\setcounter{section}{0} 

\section{Details on Experimental Setup}
\label{supp-sec:details}

\subsection{Evaluation Method}\label{supp-sec:eval}

To evaluate the attack, existing methods \cite{zhang2020secret,chen2021knowledge,nguyen2023re,struppekcareful,an2022mirror} train an evaluation model $E$ that has a distinct architecture and is trained on the private dataset $\D_{priv}$.  Similar to human inspection practices \citep{zhang2020secret}, the evaluation model acts as a human proxy for assessing the quality of information leaked through MI attacks. We report the details of the evaluation models in the Tab. \ref{tab:E}. All the evaluation models are provided by \cite{chen2021knowledge,struppek2022plug,an2022mirror}.

\begin{table*}[ht!]
\centering
\caption{Details of evaluation model $E$ in all the experimental setup. All the evaluation models are provided by \cite{chen2021knowledge,struppek2022plug,an2022mirror}.}
\begin{adjustbox}{width=1.0\columnwidth,center}
\begin{tabular}{lccclcc}
\hline
Attack & $T$ & $\D_{priv}$ & $\D_{pub}$ & Resolution & $E$ & $E$'s accuracy \\  \hline
GMI \citep{zhang2020secret} & \multirow{6}{*}{\begin{tabular}[c]{@{}l@{}}VGG16 \citep{simonyan2014very}\\ IR152 \citep{he2016deep}\\ FaceNet64 \citep{cheng2017know}\end{tabular}} & \multirow{6}{*}{CelebA} & \multirow{6}{*}{CelebA/FFHQ} & \multirow{6}{*}{64$\times$64} & \multirow{6}{*}{FaceNet112} &  \multirow{6}{*}{95.80}\\
KedMI \citep{chen2021knowledge} &  &  &  &  \\
LOMMA \citep{nguyen2023re} &  &  &  &  \\
PLGMI \citep{yuan2023pseudo} &  &  &  &  \\
RLBMI \citep{han2023reinforcement} &  &  &  &  \\
BREPMI \citep{kahla2022label} &  &  &  &  \\ \hline
\multirow{7}{*}{PPA \citep{struppek2022plug}} & ResNet18 \citep{he2016deep} & \multirow{6}{*}{Facescrub} & \multirow{6}{*}{FFHQ} & \multirow{7}{*}{224$\times$224} & \multirow{7}{*}{Inception-V3} & \multirow{7}{*}{96.20\%}\\
 & ResNet101 \citep{he2016deep} &  &  &  \\
 & ResNet152 \citep{he2016deep} &  &  &  \\
 & DenseNet121 \citep{huang2017densely} &  &  &  \\
 & DenseNet169 \citep{huang2017densely} &  &  &  \\
 & MaxVIT \citep{tu2022maxvit} &  &  &  \\ \cline{2-4} \cline{6-7} 
 & ResneSt101 & Stanford Dogs & AFHQ Dogs &  & Inception-V3 & 79.79\% \\ \hline
 \multirow{2}{*}{MIRROR \citep{an2022mirror}}
 & Inception-V1 \citep{Inceptionv1} & \multirow{2}{*}{VGGFace2} & \multirow{2}{*}{FFHQ} &  160$\times$160 & ResNet50 & 99.88\% \\ \cline{2-2} \cline{5-7} 
  & ResNet50 \citep{he2016deep}&   &   &  224$\times$224 & Inception-V1 & 99.65\% \\ \hline
   \multirow{2}{*}{IF-GMI \citep{qiu2024closer}}
 & ResNet18  & \multirow{2}{*}{Facescrub} & \multirow{2}{*}{FFHQ} &  \multirow{2}{*}{224$\times$224} & \multirow{2}{*}{Inception-V3} & \multirow{2}{*}{96.20\%}\\
  & ResNet152  &   &   &   \\ \hline 
 \end{tabular}
\end{adjustbox}

\label{tab:E}

\end{table*}

We evaluate defense methods using the following metrics: 
\begin{itemize}

\item \textbf{Natural Accuracy (Acc $\uparrow$)}. This metric measures the accuracy of the defended model on a private test set, reflecting its performance on unseen data. Higher natural accuracy indicates better performance of the primary task.

\item \textbf{Attack accuracy (AttAcc $\downarrow$) \cite{zhang2020secret}.} This metric measures the percentage of successful attacks, where success is defined as the ability to reconstruct private information from the model's outputs. Lower attack accuracy indicates a more robust defense. Following existing works \citep{zhang2020secret,chen2021knowledge, nguyen2023re,struppek2022plug}, we utilize a separate evaluation model $E$ to evaluate the inverted images. Higher attack accuracy on the evaluation model signifies a more effective attack, implying a weaker defense.

\item \textbf{K-Nearest Neighbor Distance (KNN Dist $\uparrow$) \cite{chen2021knowledge}.}
KNN distance measures the similarity between a reconstructed image of a specific identity and their private images. This is calculated using the $L_2$ norm in the feature space extracted from the penultimate layer of the evaluation model. In MI defense, a higher KNN Dist value indicates a greater degree of robustness against model inversion (MI) attacks and a lower quality of attacking samples on that model. 

\item \textbf{$\delta_{eval}$ and  $\delta_{face}$\cite{struppek2022plug}.} We also use $\delta_{eval}$ and $\delta_{face}$ metrics from \citep{struppek2022plug} to quantify the quality of inverted images generated by PPA. These two metrics are the same concept as KNN Dist, but different in the model to produce a feature to calculate distance. $\delta_{face}$ use pretrained FaceNet \citep{schroff2015facenet} as model to extract penultimate features, while $\delta_{eval}$ uses evaluation model for PPA attack.

\item \textbf{Trade-off value ( $\Delta$ $\uparrow$) \cite{ho2024model}.} To quantify the trade-off between model utility (natural accuracy) and attack performance (attack accuracy), we follow previous work and let NoDef model and defended model be $f_n$ and $f_d$ respectively, we compute 
$ \Delta = \frac{AttAcc_{f_n}-AttAcc_{f_d}}{Acc_{f_n}-Acc_{f_d}}$. 
This metric calculates the ratio between the decrease in attack accuracy and the decrease in natural accuracy when applying an MI attack to a model without defenses (NoDef) and defense models. We remark that this metric is used when defense models have lower natural accuracy compared to the no-defense model.
A higher $\Delta$ value indicates a more favorable trade-off. 

\end{itemize}

\subsection{Dataset}\label{supp-sec:dataset}
We use three datasets including CelebA \citep{liu2015deep}, Facescrub \citep{ng2014data}, and Stanford Dogs \citep{dataset2011novel} as private training data and use two datasets including FFHQ \citep{karras2019style} and AFHQ Dogs\citep{choi2020stargan} as public dataset. 

The CelebA dataset \citep{liu2015deep} is an extensive compilation of facial photographs, encompassing more than 200,000 images that represent 10,177 distinct persons. For MI task, we follow \citep{zhang2020secret,chen2021knowledge,nguyen2023re} to divide CelebA into private dataset and public dataset. There is no overlap between private and public dataset. All the images are resized to 64$\times$64 pixels.

Facescrub \citep{ng2014data} consists of a comprehensive collection of 106836 photographs showcasing 530 renowned male and female celebrities. Each individual is represented by an average of around 200 images, all possessing diversity of resolution. 
Following PPA \citep{struppek2022plug}, we resize the image to 224$\times$224 for training target models.

The FFHQ dataset comprises 70,000 PNG images of superior quality, each possessing a resolution of 1024x1024 pixels. FFHQ is used as a public dataset to train GANs using during attacks \citep{zhang2020secret,chen2021knowledge,struppek2022plug}. 
 
Stanford dogs \citep{dataset2011novel} contains more than 20,000 images encompassing 120 different dogs. AFHQ Dogs \citep{choi2020stargan} contain around 5,000 dog images in high resolution. Follow \citep{struppek2022plug}, we use Stanford dogs dataset as private dataset while AFHQ Dogs as the public dataset.

VGGFace2 \citep{cao2018vggface2} is a large-scale face recognition dataset designed for robust face recognition tasks. It consists of images that are automatically downloaded from Google Image Search, capturing a wide range of variations in factors such as pose, age, illumination, ethnicity, and profession. The diversity of the dataset makes it suitable for training and evaluating face recognition models across different conditions and demographics. It contains more than 3.3 milions images for 9000 identities.

\subsection{Train the Defense model using Random Erasing}
\label{sec:train_RE}
We depict our method in Algorithm \ref{alg:trainRE}.

\begin{algorithm*}[t]
\caption{Train the Defense model using Random Erasing}\label{alg:one}
\begin{algorithmic}
\State \textbf{Input:} Private training data $\D_{priv} = \{(x_i, y_i) \}^N_{i=1}$, model $T_\theta$, 
a maximum masking area portion $a_h$.
\State \textbf{Output:} The MIDRE-trained model $T_\theta$.
\State  Initialize $t \gets 0$
\While{$t<t_{RE}$}
\State  Sample a mini-batch $\D_b$ with size $b$ from $\D_{priv}$
\State  $\D_{RE} = \{\}$ 
\While{$(x,y)$ in $\D_b$} 
\State  $\tilde{x} = x$
\State  Randomly select $a_e$ within the range $[0.1,a_h]$
\State  $\tilde{x} = RE(x,a_e)$  \algorithmiccomment{\textit{This is following the procedure discussed in Sec. 2.2}}
\State  $\D_{mask} \leftarrow (\tilde{x},y)$
\EndWhile
\State  Compute ${\mathcal{L}}(\theta) = \frac{1}{b}\sum^{\D_{RE}} \ell(T_{\theta}(\tilde{x_i}), y_i)$
\State  Backward Propagation $\theta\gets\theta - \alpha \nabla {\mathcal{L}}(\theta)$
\EndWhile
\end{algorithmic}
\label{alg:trainRE}
\end{algorithm*}
 
\subsection{Hyper-parameters for Model Inversion Attack}\label{supp-sec:Attack}
In the case of GMI\citep{zhang2020secret}, KedMI\citep{chen2021knowledge}, and PLG-MI\citep{yuan2023pseudo}, BREPMI\citep{kahla2022label}, our approach is primarily based on the referenced publication outlining the corresponding attack. However, in certain specific scenarios, we adhere to the BiDO study due to its distinct model inversion attack configuration in comparison to the original paper. 
The LOMMA\citep{nguyen2023re} approach involves adhering to the optimal configuration of the method, which encompasses three augmented model architectures: EfficientNetB0, EfficientNetB1, and EfficientNetB2.  We adopt exactly the same  experimental configuration, including the relevant hyper-parameters, as described in the referenced paper. We also follow PPA and MIRROR paper's configuration \citep{struppek2022plug, an2022mirror} for our MI attack setups. 

\subsection{Hyper-parameters for MIDRE}
Our method only requires a hyper-parameter $a_h$, which is 0.4 for all low-resolution setups. According to high-resolution setups, we use $a_h=0.4$ and $a_h=0.8$ as two setups for our defense.
\section{Additional Experimental Results}
\label{sec:add_results}

\subsection{Experiments on low resolution images}
\label{sec:low_res}

We evaluate our method against existing Model Inversion defenses. We follow the experiment setup in BiDO \citep{peng2022bilateral} and report the results on the standard setup using $T$ = VGG16 and $\D_{priv}$ = CelebA in Tab.~\ref{tab:MIA64_VGG16}. We evaluate against six MI attacks, including GMI \citep{zhang2020secret}, KedMI \citep{chen2021knowledge}, LOMMA \citep{nguyen2023re} with two variances (LOMMA+GMI and LOMMA+KedMI), PLGMI \citep{yuan2023pseudo}, and a black-box attack, BREPMI \citep{kahla2022label}.


Overall, our proposed method, MIDRE, achieves significant improvements in security for 64$\times$64 setups compared to SOTA MI defenses. MIDRE achieves this by demonstrably reducing top-1 attack accuracy while maintaining natural accuracy on par with other leading MI defenses.
Specifically, compared to BiDO, MIDRE offers a substantial 43.74\% decrease in top-1 attack accuracy with sacrificing only 7.05\% in natural accuracy (measured using the KedMI attack method). Notably, while BiDO achieves similar natural accuracy to MIDRE, it suffers from a significantly higher top-1 attack accuracy (8.84\% higher than MIDRE).

\begin{table}[h!]  
\centering
\caption{We report the MI attacks under multiple SOTA MI attacks on images with resolution 64$\times$64. 
We compare the performance of these attacks against existing defenses including NoDef, BiDO, MID and our method.
$T$ = VGG16, $D_{priv}$ = CelebA, $D_{pub}$ = CelebA.
}
\begin{tabular}{clcccc}
\hline
Attack &  Defense & Acc $\uparrow$ & AttAcc $\downarrow$ & $\Delta$ $\uparrow$ & KNN Dist $\uparrow$\\ 
\hline
\multirow{4}{1.5cm}{LOMMA + GMI} & NoDef & 86.90 & 74.53 $\pm$ 5.65 &  - & 1312.93 \\
 &  MID & 79.16 & 54.53 $\pm$ 4.35 &  2.58 & 1348.21  \\
 &  BiDO & 79.85 & 53.73 $\pm$ 4.99 &  2.95 & 1422.75 \\
 &  \textbf{MIDRE} & 79.85  & \textbf{31.93 $\pm$ 5.10} & \textbf{6.04} &\textbf{1590.12} \\  
\hline
\multirow{4}{1.5cm}{LOMMA + KedMI}  & NoDef & 86.90 & 81.80 $\pm$ 1.44 & - & 1211.45\\
 &  MID & 79.16 & 67.20 $\pm$ 1.59  & 1.89 & 1249.18 \\
 &  BiDO & 79.85 & 63.00 $\pm$ 2.08 &  2.67 & 1345.94 \\
 &  \textbf{MIDRE} & 79.85 & \textbf{43.07 $\pm$ 1.99}  &  \textbf{5.49} & \textbf{1503.89} \\  
\hline
\multirow{4}{1.5cm}{PLGMI} &  NoDef & 86.90  &   97.47 $\pm$ 1.68 & - & 1149.67 \\
 &  MID & 79.16 & 93.00 $\pm$ 1.90 &  0.58 & 1111.61 \\
 &  BiDO & 79.85 & 92.40 $\pm$ 1.74 &  0.72 & 1228.36 \\
 &  \textbf{MIDRE} & 79.85 & \textbf{66.60 $\pm$ 2.94} & \textbf{4.38} & \textbf{1475.76} \\    \hline
\multirow{4}{1.5cm}{GMI} & NoDef & 86.90 &  20.07 $\pm$ 5.46 & - & 1679.18 \\
 &  MID & 79.16 & 20.93 $\pm$ 3.12 &  -0.11 & 1698.50 \\
 &  BiDO & 79.85 & 6.13 $\pm$ 2.98 &  1.98 & 1927.11\\
 &  \textbf{MIDRE} & 79.85 & \textbf{3.20 $\pm$ 2.15} & \textbf{2.39} & \textbf{2020.49}  \\  
\hline
\multirow{4}{1.5cm}{KedMI}  & NoDef & 86.90 &  78.47 $\pm$ 4.60 & - & 1289.46\\
 &  MID & 79.16 & 53.33 $\pm$ 4.97 &   3.25 & 1364.02 \\ 
 &  BiDO & 79.85 & 43.53 $\pm$ 4.00 &  4.96 & 1494.35\\
 &  \textbf{MIDRE} & 79.85 & \textbf{34.73 $\pm$ 4.15} & \textbf{6.20} & \textbf{1620.66}\\  
\hline
\multirow{4}{1.5cm}{BREPMI} &  NoDef & 86.90 & 57.40 $\pm$ 4.92  & - & 1376.94  \\
 &  MID & 79.16 & 39.20 $\pm$ 4.19  & 2.35 & 1458.61 \\
 &  BiDO & 79.85 & 37.40 $\pm$ 3.66  & 2.84 & 1500.45  \\
 &  \textbf{MIDRE} & 79.85 & \textbf{21.73 $\pm$ 2.99} & \textbf{5.06} & \textbf{1611.78} \\    \hline
\end{tabular}

\label{tab:MIA64_VGG16}
\end{table}

\begin{table}[t]
\centering
\caption{Results of IF-GMI\citep{qiu2024closer} attack on Facescrub dataset. Here, we use $T$ = ResNet18/ ResNet152, $\D_{priv}$ = Facescrub, $\D_{pub}$ = FFHQ, image resolution = 224$\times$224 images, attack method = IF-GMI.
}
\begin{tabular}{lcccccc}
\hline
Architecture &  Defense & Acc  $\uparrow$ & AttAcc $\downarrow$ & $\delta_{eval}$ $\uparrow$ & $\delta_{face}$ $\uparrow$ &  FID $\uparrow$ \\ 
\hline
\multirow{2}{*}{ResNet18} & NoDef & 94.22 & 98.30 & 110.04 & 0.647  & 40.239 \\
 &   \textbf{MIDRE (0.1, 0.4)} & 97.28 & 72.58 & 122.03 & 0.698 & 39.7238 \\
 &   \textbf{MIDRE (0.1, 0.8)} & 93.33 & \textbf{24.85} & \textbf{171.48} & \textbf{0.966} & \textbf{41.325} \\  \hline
 
 \multirow{2}{*}{ResNet152} & NoDef & 95.43 & 97.24 & 115.76 & 0.633 & \textbf{45.703} \\
 &   \textbf{MIDRE (0.1, 0.4)} & 97.90 & 74.50 & 133.22 & 0.662 & 40.669
 \\ 
 &   \textbf{MIDRE (0.1, 0.8)} & 95.74 & \textbf{31.43} & \textbf{150.89} & \textbf{0.847} & 40.388 \\  \hline
\end{tabular}

\label{tab:IF-GMI}
\vspace{-0.3cm}
\end{table}

\label{supp-sec:additional_results}
\subsection{Additional results}\label{supp-subsec:extensive_results}

We further show the effectiveness of our proposed method on a wide range of target model architectures including IR152, FaceNet64, DenseNet-169, ResNeSt-101, and MaxVIT. The results are shown in Tab. \ref{tab:MIA_nodef64_IR152}, \ref{tab:MIA_nodef64_FaceNet64}, and Tab.\ref{tab:MIA_nodef64_TLDMI_IR152} and \ref{tab:MIA_nodef64_TLDMI_FaceNet64} (for comparison with TL-DMI) for 64$\times$64 images and in Tab.\ref{tab:PPA_0} and \ref{tab:PPA_1} for 224$\times$224 images.

The experiment results consistently demonstrate the effectiveness of our proposed method. For example, with $T$ = IR152, we sacrifice only 6.25\% in natural accuracy, but the attack accuracies drop significantly, from 22.07\% (PLGMI attack) to 40\% (LOMMA + GMI attack). Similarly, when $T$ = FaceNet64, natural accuracy decreases by 6.94\%, while the attack accuracies drop significantly, from 24.47\% (PLGMI attack) to 45\% (LOMMA attack).
We report the results of additional setup in Tab. \ref{tab:MIA64_FFHQ_FaceNet64}. In particular, we use attack method = PLGMI, $T$ = VGG16/IR152/FaceNet64, $\D_{priv}$ = CelebA, $\D_{pub}$ = FFHQ.
In addition to measuring attack accuracy, we incorporate KNN distance to demonstrate the efficacy of our strategy across different evaluation scenarios. The specifics of KNN distance can be found in Sec. \ref{supp-sec:eval}.

For high resolution images, interestingly, with $\D_{priv}$ = Facescrub, we see a slight increase in natural accuracy (1.95\%) along with a significant reduction in attack accuracy of around 40\%. These results consistently show that MIDRE significantly reduces the impact of MI attacks. 
We report detailed results of PPA attack on our method compared to SOTA defense including MID, DP, BiDO, TL-DMI, NLS and RoLSS, SSF, TTS. the results are presented in Tab. \ref{tab:PPA_0} and \ref{tab:PPA_1}. We also use $\delta_{eval}$ and $\delta_{face}$, with details in Sec. \ref{supp-sec:eval} to evaluate quality of PPA inverted images.

\begin{table}[h!]
\centering
\caption{Additional results on 64$\times$64 images. We use $T$ = IR152. The target models are trained on $\D_{priv}$ = CelebA and  $\D_{pub}$ = CelebA. The results conclusively show that our defense model is effective compared to NoDef models.
}
\begin{tabular}{p{1.5cm}cccc}
\hline
Attack &  Defense & Acc  $\uparrow$ & AttAcc $\downarrow$ & KNN Dist $\uparrow$ \\ 
\hline
\multirow{2}{*}{GMI} & NoDef & 91.16 & 32.40 $\pm$ 4.88 & 1587.28  \\
 &   \textbf{MIDRE} & 84.91 & \textbf{7.87 $\pm$ 3.30} & \textbf{1888.47}  \\  \hline
\multirow{2}{*}{KedMI} &  NoDef & 91.16 & 78.93 $\pm$ 5.15  & 1262.44   \\
 &   \textbf{MIDRE} & 84.91 & \textbf{40.07 $\pm$ 4.99} & \textbf{1548.16}   \\  \hline
\multirow{2}{1.5cm}{LOMMA + GMI} &  NoDef & 91.16 & 80.93 $\pm$ 4.56 & 1253.03   \\
 &   \textbf{MIDRE} & 84.91 & \textbf{40.93 $\pm$ 6.11} & \textbf{1559.88}    \\ \hline
\multirow{2}{1.5cm}{LOMMA + KedMI} &   NoDef & 91.16 &  90.87 $\pm$ 1.31 & 1116.90   \\
 &   \textbf{MIDRE} & 84.91 & \textbf{52.13 $\pm$ 1.81} & \textbf{1481.70}    \\ \hline
\multirow{2}{*}{PLGMI} &   NoDef & 91.16 &  99.47 $\pm$ 0.93 & 1021.42   \\
 &   \textbf{MIDRE} & 84.91 & \textbf{77.40 $\pm$ 4.79} & \textbf{1470.46}   \\  \hline
\end{tabular}

\label{tab:MIA_nodef64_IR152}
\end{table}

\begin{table}[h!]
\centering
\caption{Additional results on 64$\times$64 images. We use $T$ = FaceNet64. The target models are trained on $\D_{priv}$ = CelebA and  $\D_{pub}$ = CelebA. The results conclusively show that our defense model is effective compared to NoDef models.
}
\begin{tabular}{p{1.5cm}cccc}
\hline
Attack &  Defense & Acc  $\uparrow$ & AttAcc $\downarrow$  & KNN Dist $\uparrow$  \\ 
\hline
\multirow{2}{*}{GMI} & NoDef & 88.50 &  29.60 $\pm$ 5.43 & 1607.86   \\
 &   \textbf{MIDRE} & 81.56 & \textbf{6.73 $\pm$ 3.42} & \textbf{1908.19}    \\  \hline
\multirow{2}{*}{KedMI} &  NoDef & 88.50 &  81.67 $\pm$ 2.63 & 1270.71   \\
 &   \textbf{MIDRE} & 81.56 & \textbf{36.33 $\pm$ 6.06} & \textbf{1545.93}   \\  \hline
\multirow{2}{1.5cm}{LOMMA + GMI} &  NoDef & 88.50 & 83.33 $\pm$ 3.40 & 1259.61    \\
 &   \textbf{MIDRE} & 81.56 & \textbf{37.60 $\pm$ 3.74}  & \textbf{1570.85}   \\ \hline
\multirow{2}{1.5cm}{LOMMA + KedMI} &   NoDef & 88.50 & 90.87 $\pm$ 1.31 & 1116.90  \\
 &   \textbf{MIDRE} & 81.56 & \textbf{54.33 $\pm$ 1.44}  & \textbf{1456.84}   \\ \hline
\multirow{2}{*}{PLGMI} &   NoDef & 88.50 & 99.47 $\pm$ 0.69 & 1091.51    \\
 &   \textbf{MIDRE} & 81.56 & \textbf{75.00 $\pm$ 4.30} & \textbf{1509.78}   \\  \hline
\end{tabular}

\label{tab:MIA_nodef64_FaceNet64}
\end{table}

\begin{table}[h!]
\centering
\caption{Additional results compared with TL-DMI on 64$\times$64 images. We use $T$ = IR152. The target models are trained on $\D_{priv}$ = CelebA and  $\D_{pub}$ = CelebA. The results conclusively show that our defense model is effective.}
\label{tab:MIA_nodef64_TLDMI_IR152}
\begin{tabular}{lccccc}
\hline
Attack &  Defense & Acc  $\uparrow$ & AttAcc $\downarrow$ & $\Delta$ $\uparrow$ & KNN Dist$\uparrow$  \\ 
\hline
\multirow{3}{1.3 cm}{GMI} & NoDef & 91.16 & 32.40 $\pm$ 4.88 & - & 1587.28  \\
&   \textbf{TL-DMI} & 86.70 & \textbf{8.93 $\pm$ 3.73} & 5.26 & \textbf{1819.00}  \\ 
 &   \textbf{MIDRE} & 87.94 & 11.07 $\pm$ 3.60 & \textbf{6.62}  & 1813.11 \\ \hline
\multirow{3}{1.3 cm}{KedMI} &  NoDef & 91.16 & 78.93 $\pm$ 5.15 & - & 1262.44   \\
&   \textbf{TL-DMI} & 86.70 & 64.60 $\pm$ 4.93 & 3.21  & 1333.00  \\
 &   \textbf{MIDRE} & 87.94 & \textbf{46.67 $\pm$ 5.45}  & \textbf{10.02} & \textbf{1455.88}   \\  \hline
\multirow{3}{1.3 cm}{LOMMA + GMI} &  NoDef & 91.16 & 80.93 $\pm$ 4.56 & - & 1253.03  \\
&  TL-DMI & 86.70 & \textbf{41.87 $\pm$ 5.37} & 8.76 & \textbf{1551.00}    \\
 &   \textbf{MIDRE} & 87.94 & 49.40 $\pm$ 6.30 & \textbf{9.79} & 1497.50    \\ \hline
\multirow{3}{1.3 cm}{LOMMA + KedMI} &   NoDef & 91.16 &  90.87 $\pm$ 1.31  & - & 1116.90 \\
&   TL-DMI & 86.70 &  77.73 $\pm$ 1.57 & 2.95 & 1305.00   \\
 &   \textbf{MIDRE} & 87.94 & \textbf{62.93 $\pm$ 2.15} & \textbf{8.68} & \textbf{1551.00}     \\ \hline
\end{tabular}

\end{table}

\begin{table}[h!]
\centering
\caption{Additional results compared with TL-DMI on 64$\times$64 images. We use $T$ = FaceNet64. The target models are trained on $\D_{priv}$ = CelebA and  $\D_{pub}$ = CelebA. The results conclusively show that our defense model is effective.
}
\begin{tabular}{lccccc }
\hline
Attack &  Defense & Acc  $\uparrow$ & AttAcc $\downarrow$ & $\Delta$ $\uparrow$  & KNN Dist $\uparrow$  \\ 
\hline
\multirow{3}{1.3 cm}{GMI} & NoDef & 88.50 & 29.60 $\pm$ 5.43 & - & 1607.86 \\
&   \textbf{TL-DMI} & 83.41 & 15.73 $\pm$ 4.58 & 2.72 & 1752.00  \\ 
 &   \textbf{MIDRE} & 85.74 & \textbf{7.47 $\pm$ 2.59} & \textbf{8.02} & \textbf{1898.29}  \\  \hline
\multirow{3}{1.3 cm}{KedMI} &  NoDef & 88.50 & 81.67 $\pm$ 2.63 & - & 1270.71   \\
&   \textbf{TL-DMI} & 83.41 & 73.40 $\pm$ 4.10 & 1.62 & 1265.00   \\
 &   \textbf{MIDRE} & 85.74 & \textbf{42.93 $\pm$ 5.22} & \textbf{14.04} & \textbf{1512.52}   \\  \hline
\multirow{3}{1.3 cm}{LOMMA + GMI} &  NoDef & 88.50 & 83.33 $\pm$ 3.40 & - & 1259.61  \\
&  TL-DMI & 83.41 & 43.67 $\pm$ 5.60 & 7.79 & \textbf{1616.00}   \\
 &   \textbf{MIDRE} & 85.74 & \textbf{43.33 $\pm$ 6.02} & \textbf{14.49} & 1550.77    \\ \hline
\multirow{3}{1.3 cm}{LOMMA + KedMI} &   NoDef & 88.50 &  90.87 $\pm$ 1.31  & - & 1116.90 \\
&   TL-DMI & 83.41 &  79.60 $\pm$ 1.78 & 2.21 & \textbf{1345.00}   \\
 &   \textbf{MIDRE} & 85.74 & \textbf{58.07 +/- 1.78}  &  \textbf{11.88} & 1386.67  \\ \hline

\end{tabular}

\label{tab:MIA_nodef64_TLDMI_FaceNet64}
\end{table}

\begin{table}[ht!]
\centering
\caption{We report the PLGMI attacks on images with resolution 64$\times$64. We compare to NoDef and BiDO methods. 
$T$ = VGG16, IR152 and FaceNet64, $D_{pub}$ = FFHQ. We remark that BiDO only releases their implementation and pretrained model in the setup of $T$ = VGG16.
}
\begin{tabular}{lccccc}
\hline
Architecture &  Defense & Acc  $\uparrow$  & AttAcc $\downarrow$ & $\Delta$ $\uparrow$ & KNN Dist $\uparrow$  \\ 
\hline
\multirow{3}{*}{VGG16} & NoDef & 86.90 & 81.80 $\pm$ 2.74 & - & 1323.27  \\
 &   BiDO & 79.85 & 60.93 $\pm$ 3.99 & 2.96 & 1440.16   \\ 
 & \textbf{MIDRE} & 79.85 & \textbf{36.07 $\pm$ 4.76} & \textbf{6.49} & \textbf{1654.41} \\ \hline
\multirow{2}{*}{IR152} & NoDef & 91.16 & 96.60 $\pm$ 2.11 & - & 1187.37  \\
 & \textbf{MIDRE} & 84.91 & \textbf{54.02 $\pm$ 4.86} & \textbf{6.81} & \textbf{1579.28} \\ \hline
\multirow{2}{*}{FaceNet64} & NoDef & 88.50 & 95.00 $\pm$ 2.56 & - & 1250.90  \\
 & \textbf{MIDRE} & 81.56 & \textbf{51.60 $\pm$ 3.61} & \textbf{6.25} & \textbf{1501.85} \\ \hline
\end{tabular}

\label{tab:MIA64_FFHQ_FaceNet64}
\end{table}

\begin{table*}[ht!]  
\centering
\caption{We report the PPA MI attacks on images with resolution 224$\times$224. 
We compare the performance of these attacks against existing defenses including NoDef, MID, DP, BiDO NLS, TLDMI, and MI-RAD variances.
$D_{priv}$ = Facescrub $D_{pub}$ = FFHQ, Arhchitecture is Resnet18, ResNet152 and ResNet101. 
We denote ``NA" for $\delta_{face}$ and $\delta_{eval}$ if these numbers are not available in the official paper \cite{ho2024model,koh2024vulnerability,struppekcareful}. We denote ``OP" for $\Delta$ if the accuracy of the defense model outperforms that of the NoDef model.
}
\begin{tabular}{clccccc}
\hline
Architecture &  Defense & Acc $\uparrow$ & AttAcc $\downarrow$ & $\delta_{eval}$ $\uparrow$ & $\delta_{face}$ $\uparrow$ & $\Delta$ $\uparrow$  \\ \hline
\multirow{7}{*}{ResNet18} & NoDef & 94.22 & 88.67 & 123.85 & 0.74 & - \\ 
& MID	& 91.15	& 65.47	& 137.75 &0.87 & 7.56 \\
& DP & 89.80 & 75.26 & 130.41 & 0.82 & 3.03 \\
& BiDO & 91.33 & 76.56 & 127.86 & 0.75 & 4.54 \\
& TL-DMI & 91.12 & 22.36 & NA & NA & 21.39\\
& MIDRE(0.1, 0.4) & \textbf{97.28} & 48.16 & 131.72 & 0.80 & OP\\
& MIDRE(0.1,0.8) & 93.33 & \textbf{13.89} & \textbf{154.79} & \textbf{0.97} & \textbf{84.02} \\
\hline
\multirow{9}{*}{ResNet152} & NoDef & 95.43 & 86.51 & 113.03 & 0.73 & - \\ 
& MID & 91.56 & 66.18 & 137.18 & 0.86 & 5.25 \\
& BiDO & 91.80 & 58.14 & 147.28	& 0.87 & 7.82 \\
& NLS & 91.50 & \textbf{14.34} & NA & \textbf{1.23} & 18.36 \\
& RoLSS & 93.00 & 64.98 & NA & NA & 8.86\\
& SSF & 93.79 & 70.71 & NA & NA & 9.63\\
& TTS & 93.97 & 73.59 & NA & NA & 8.85 \\
& MIDRE(0.1,0.4) & \textbf{97.90} & 42.44 & 139.66 & 0.82 & OP \\
& MIDRE(0.1,0.8) & 95.47 & \textbf{15.97} & \textbf{155.61} & 0.95 & OP \\
\hline
\multirow{11}{*}{ResNet101} & NoDef & 94.86 & 83.00 & 128.60 & 0.76 & - \\ 
& MID & 92.70 & 82.08 & 122.96 & 0.76 & 0.43\\
& DP & 91.36 & 74.88 & 131.38 & 0.82 & 2.32 \\
& BiDO & 90.31 & 67.07 & 139.15 & 0.84 & 3.50 \\
& TL-DMI & 90.10 & 31.82 & NA & NA & 10.75 \\
& NLS(-0.05) & 94.79 & 33.14 & 130.94 & 0.90 & 712.29 \\
& RoLSS	& 92.40	& 58.68 & NA & NA & 9.89 \\
& SSF &	93.79 &	71.06 & NA & NA & 11.16 \\
& TTS & 94.16 & 77.26 & NA & NA & 8.20\\
& MIDRE(0.1,0.4) & \textbf{98.02} & 43.58 & 139.01 & 0.81 & OP \\
& MIDRE(0.1,0.8) & 95.15	& \textbf{15.47}	& \textbf{155.80} & \textbf{0.96} & OP\\
\hline
\end{tabular}

\label{tab:PPA_0}
\end{table*}

\begin{table*}[ht!] 
\centering
\caption{We report the PPA MI attacks on images with resolution 224$\times$224. 
We compare the performance of these attacks against existing defenses including NoDef, MID, DP, BiDO NLS, TLDMI, and MI-RAD variances.
$D_{priv}$ = Facescrub $D_{pub}$ = FFHQ, Arhchitecture is DenseNet169, DenseNet121, ResneSt101, and MaxVIT.
We denote ``NA" for $\delta_{face}$ and $\delta_{eval}$ if these numbers are not available in the official paper \cite{ho2024model,koh2024vulnerability,struppekcareful}. We denote ``OP" for $\Delta$ if the accuracy of the defense model outperforms that of the NoDef model.
}
\begin{tabular}{clccccc}
\hline
Architecture &  Defense & Acc $\uparrow$ & AttAcc $\downarrow$ & $\delta_{eval}$ $\uparrow$ & $\delta_{face}$ $\uparrow$ & $\Delta$ $\uparrow$  \\ \hline
\multirow{5}{*}{DenseNet169} & NoDef & 95.49 & 87.80 & 124.74 &0.77 & -\\ 
& RoLSS	& 72.14 & 6.77 & NA & NA & 3.47\\
& SSF & 92.95 & 60.99 & NA & NA & 10.56\\
& MIDRE(0.1,0.4) & \textbf{97.99} & 46.67 & 136.18 & 0.81 & NA\\
& MIDRE(0.1,0.8) & 95.04 & \textbf{15.78} & \textbf{154.96} & \textbf{0.95} & \textbf{160.04} \\
\hline
\multirow{6}{*}{DenseNet121} & NoDef & 95.54 & 95.13 & 116.14 & 0.68 & -\\
& NLS(-0.05) & 92.13 & 40.69 & 179.53 & \textbf{0.97} & 15.96 \\
& RoLSS	& 74.25 & 10.24 & NA & NA & 3.99 \\
& SSF & 93.09 & 65.21 & NA & NA & 12.21\\
& MIDRE(0.1,0.4) & \textbf{98.19} & 46.98 & 134.86 & 0.81 & OP \\
& MIDRE (0.1,0.8) &	95.76 & 15.66 & \textbf{154.62} & 0.96 & OP\\
\hline
\multirow{4}{*}{ResneSt101} & NoDef & 95.38 & 84.27 & 129.18 & 0.81 & - \\
& NLS(-0.05) & 88.82 & 13.23 & \textbf{172.73} & \textbf{1.10} & 10.01 \\
& MIDRE(0.1,0.4) & \textbf{98.11} & 45.43 & 137.78 & 0.80 & NA \\
& MIDRE(0.1,0.8) & 95.09 & 15.54 & 156.44 & 0.96 & \textbf{237.00}\\
\hline
\multirow{6}{*}{MaxVIT} & NoDef & 98.36 & 80.66 & 110.69 & 0.69 & - \\
& TL-DMI & 93.01 & 21.17 & NA & NA & 10.59 \\
& NLS(-0.05) & 98.23 & 55.09 & 127.68 & 0.81 & 63.93 \\
& RoLSS & 95.09 & 25.17 & NA & NA & 15.68\\
& MIDRE(0.1,0.4) & \textbf{98.46} & 42.50 & 133.61 & 0.81 & OP \\
& MIDRE(0.1,0.8) & 96.52 & 13.92 & \textbf{155.31} & \textbf{0.96} & 31.63 \\
\hline
\end{tabular}

\label{tab:PPA_1}
\end{table*}

\subsection{User Study}
\label{sec:user_study}

\begin{figure*}[ht!]
\centering
\includegraphics[width=1.0\textwidth]{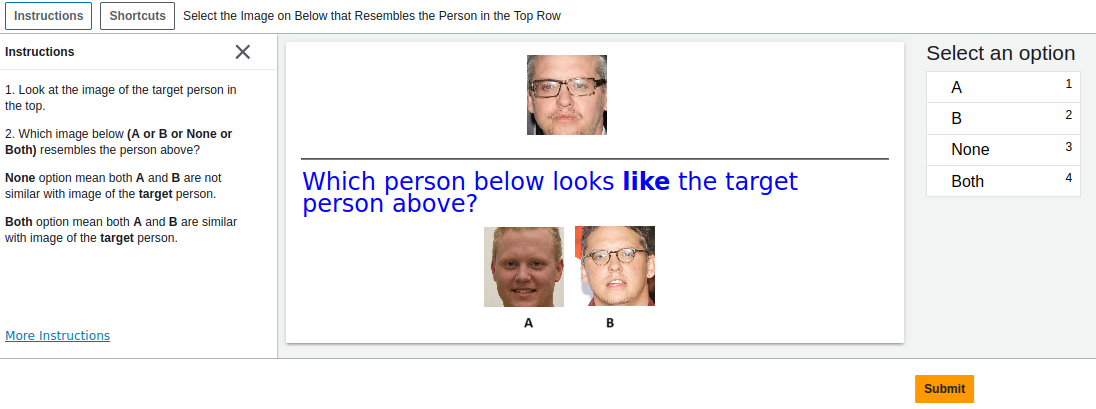}
\caption{Our Amazon Mechanical Turk (MTurk) interface for user study with model inversion attacking samples. Participants were asked to select one of 4 options: A, B, none, or both, for each image pair where A and B are the inverted images of our defense  and other defense model. Each pair was rated by three different users.}
\label{fig:user_study_interface}
\end{figure*}

In addition to attack accuracy measured by the evaluation model, we conduct a user study to further validate the attack's effectiveness. Overall, we conduct two setups for user study with low-resolution images and high-resolution images. Our interface for user study is illustrated in Fig. \ref{fig:user_study_interface}.

In the low-resolution setup, we compare our proposed method and BiDO  \citep{peng2022bilateral}. For fair comparison, we use the same setup as BiDO: $T$ = VGG16, $\D_{priv}$ = CelebA, $\D_{pub}$ = CelebA and use the pre-trained model of BiDO to generate their images. We use the attack method PLG-MI to generate the inverted images and randomly select one image for each identity for overall 150 first identities. We upload it to Amazon Mechanical Turk and designate three individuals to vote on two of our model's and BiDO's reconstructed images, for a total of 450 votes. 
Participants were asked to select one of 4 options: BiDO, MIDRE, none, or both, for each image pair. Each pair was rated by three different users.

In the high-resolution setup, we compare MIDRE and TL-DMI \citep{ho2024model}, which is a state-of-the-art MI defense. 
We use the setup: $T$ = ResNet101, $\D_{priv}$ = Facescrub, $\D_{pub}$ = FFHQ, attack method = PPA. For every defense, we create inverted images for each of the 530 classes, then select one image for each class. Finally, we upload them to Amazon Mechanical Turk and follow the same procedure as low-resolution images setup.

{\bf Comparing BiDO and our proposed MIDRE:}
According to the results, 221 users voted in favour of BiDO, 108 in favour of our approach, 119 in favour of neither, and 2 in favour of both. It suggests that the reconstructed image quality from our model is not as good as the reconstructed image quality from BiDO, therefore {\bf our proposed defense is more effective}. Our results are presented in Tab. \ref{tab:User_Study}. 

{\bf Comparing SOTA TL-DMI and our proposed MIDRE:}
According to the results in Tab. \ref{tab:User_Study_PPA}, 509 users chose images inverted from our model, while 537 users voted in favor of TL-DMI. This suggests that the inverted images from our models are of lower quality than those from TL-DMI. In addition, there are 522 people voted for none of the two images is similar with the original image, meanwhile only 22 users chose that both images are similar to the real image.

According to the final results of both settings, MIDRE is a better defense mechanism against MI than SOTA BiDO and TL-DMI, which is in line with the findings of other evaluation metrics.

\begin{table}[ht!]
\centering
\caption{We report results for an user study that was performed with Amazon Mechanical Turk. Reconstructed samples of PLG-MI/VGG16/CelebA/CelebA with first 150 classes.
The study asked users for  inputs regarding the similarity between a private training image and the reconstructed image from BiDO trained model and our trained model.  Less number of reconstructed images from our defensed model are selected by users, suggesting our defense is more effective.
}
\begin{tabular}{lp{5.5cm}}
\toprule
Defense
& Num of samples selected by users as more similar to private data
 \\ \midrule
BiDO & 221 \\ 
Ours & \textbf{108} \\ 
Both & 119 \\ 
None & 2 \\
\midrule
\end{tabular}

\label{tab:User_Study}
\end{table}

\begin{table}[ht!]
\centering
\caption{We report results for an user study that was performed with Amazon Mechanical Turk. Reconstructed samples of PPA/ResNet101/FaceScrub/FFHQ with all 530 classes.
The study asked users for  inputs regarding the similarity between a private training image and the reconstructed image from TL-DMI trained model and our trained model.  Less number of reconstructed images from our defensed model are selected by users, suggesting our defense is more effective.
}
\begin{tabular}{lp{5.5cm}}
\toprule
Defense
& Num of samples selected by users as more similar to private data
 \\ \midrule
TL-DMI &  537\\ 
Ours &  \textbf{509}\\ 
Both &  522\\ 
None &  22\\
\midrule
\end{tabular}

\label{tab:User_Study_PPA}
\end{table}

%



\subsection{Qualitative Results}

We provide inversion results from the recent IF-GMI attack in Fig.~\ref{fig:Attacking_Samples_Visual_IFGMI_R18} ($T = \text{ResNet-18}$) and Fig.~\ref{fig:Attacking_Samples_Visual_IFGMI_R152} ($T = \text{ResNet-152}$) and  $\D_{priv}$ = Facescrub, $\D_{pub}$ = FFHQ. These results further demonstrate the effectiveness of our proposed method.


\begin{figure*}
    \centering    \includegraphics[width=1.0\textwidth]{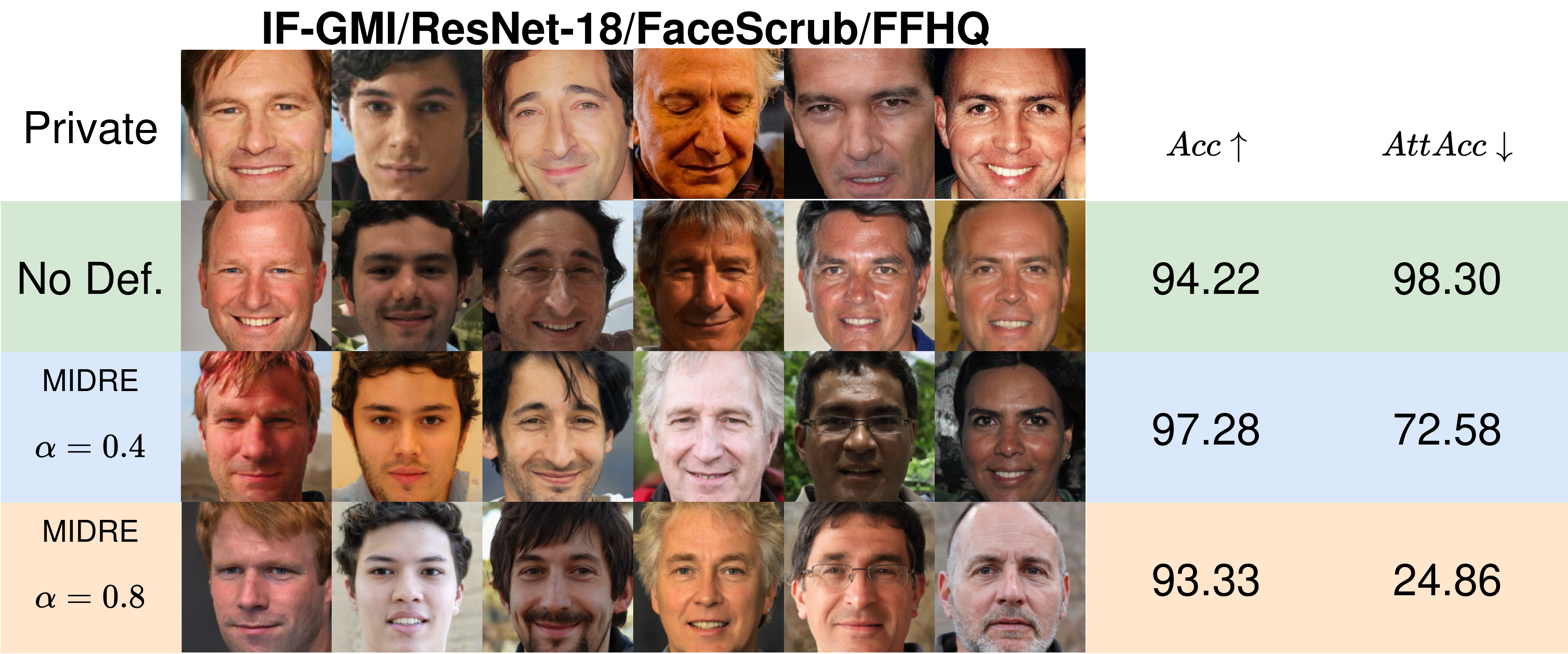} 
    \caption{
    Reconstructed image obtained from IF-GMI attack with $T$ = ResNet-18, $\D_{priv}$ = Facescrub, $\D_{pub}$ = FFHQ. The quality of the reconstructed image obtained  from the attack on  the model trained by MIDRE is comparatively worse when compared to that from NoDef method, suggesting the efficiency of our proposed defense MIDRE.}
\label{fig:Attacking_Samples_Visual_IFGMI_R18}
\end{figure*}

\begin{figure*}
    \centering    \includegraphics[width=1.0\textwidth]{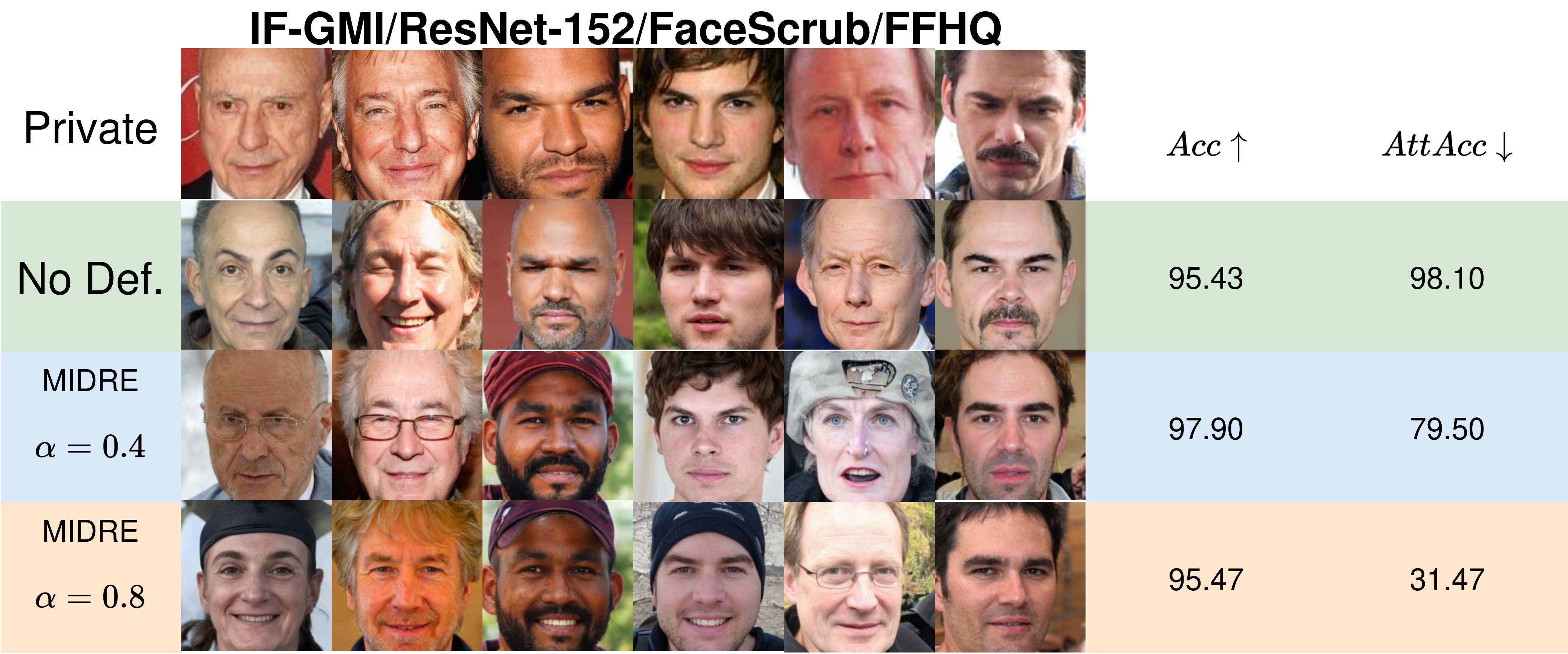} 
    \caption{Reconstructed image obtained from IF-GMI attack with $T$ = ResNet-152, $\D_{priv}$ = Facescrub, $\D_{pub}$ = FFHQ. The quality of the reconstructed image obtained  from the attack on  the model trained by MIDRE is comparatively worse when compared to that from NoDef method, suggesting the efficiency of our proposed defense MIDRE.}
\label{fig:Attacking_Samples_Visual_IFGMI_R152}
\end{figure*}

\section{Additional analysis of privacy effect of MIDRE}\label{Sec:supp_Analysis}

\subsection{Feature space analysis of Random Erasing's defense effectiveness}
In addition to the visualization of feature space analysis in Sec. 3.2 (main paper), we provide more visualization in other setup:  $T$ = ResNet-152 \citep{simonyan2014very}, $D_{priv}$ = Facecrub \citep{ng2014data}, $D_{pub}$ = FFHQ \citep{karras2019style}, attack method = PPA \citep{struppek2022plug}. 
 We observe  {\color{blue}\textbf{Property P1:}} \textbf{Model trained with RE-private images following our  MIDRE leads to a discrepancy between the features of MI-reconstructed images and that of  private images}, resulting in degrading of attack accuracy.

We use the following notation: $f_{train}$, \colorbox{priv}{$f_{priv}$}, \colorbox{privRe}{$f_{RE}$}, and \colorbox{inv}{$f_{recon}$} represent the features of training images, \colorbox{priv}{private images}, \colorbox{privRe}{RE-private images}, and \colorbox{inv}{MI-reconstructed images}, respectively. To extract these features, we first train the target model without any defense (NoDef) and another target model  with our MIDRE. Then, we pass images into these models to obtain the penultimate layer activations. Specifically, we input private images into the models to obtain $f_{priv}$. Next, we apply RE to private images, pass these RE-private images into the models to obtain $f_{RE}$. We also perform MI attacks to obtain reconstructed images from NoDef model (resp. MIDRE model), and then feed them into the NoDef model (resp. MIDRE model) to obtain $f_{recon}$.
Then, we visualize penultimate layer activations  $f_{priv}, f_{RE}, f_{recon}$ by both NoDef and our MIDRE model. We use $a_e$ = 0.4 to train MIDRE and to generate RE-private images.
Additionally, we visualize the convex hull of these features. For visualization, we employ PCA to reduce the dimension of the feature space.

\begin{figure*}[ht!]
 \centering
  \begin{subfigure}[b]{0.95\textwidth}
     \centering
     \includegraphics[width=0.24\textwidth]{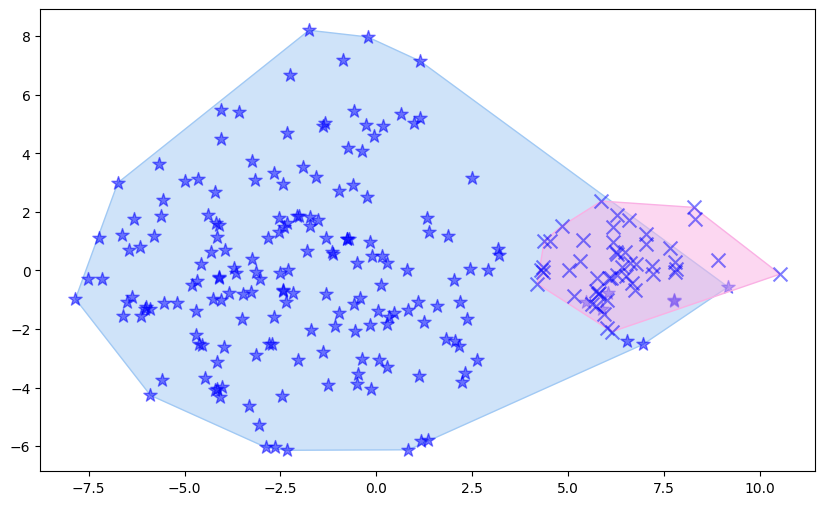}
     \includegraphics[width=0.24\textwidth]{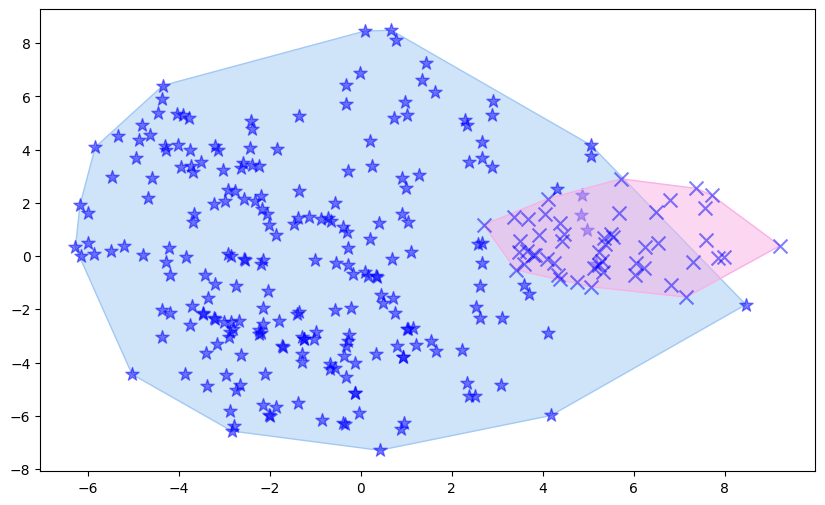}
     \includegraphics[width=0.24\textwidth]{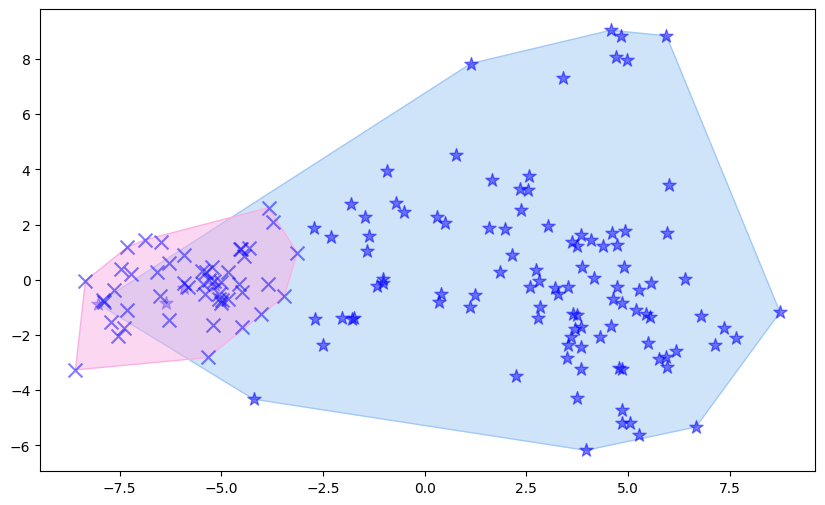}
     \includegraphics[width=0.24\textwidth]{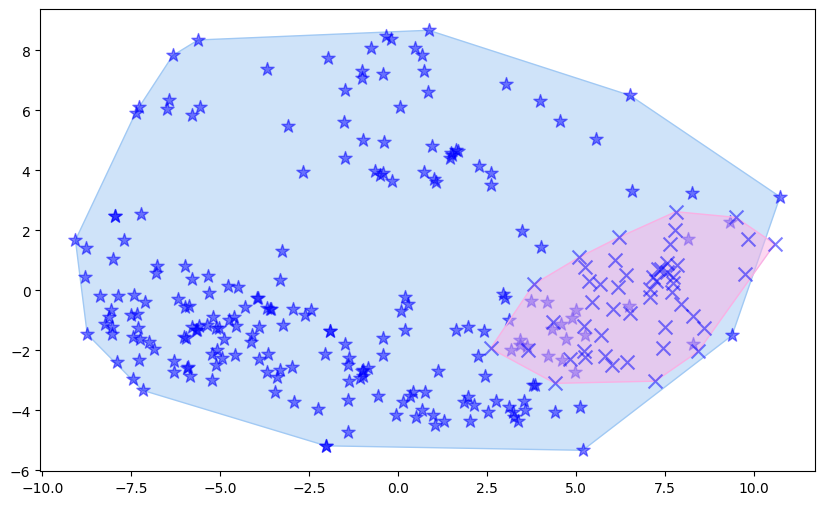}
    
\caption{NoDef, $T$ = ResNet152, $a_e$ = 0.4, AttAcc = 86.51\%, $f_{train}^{NoDef} =$ \colorbox{priv}{$f_{priv}^{NoDef}$}}
 \end{subfigure}
 
 \begin{subfigure}[b]{0.95\textwidth}
     \centering
     \includegraphics[width=0.24\textwidth]{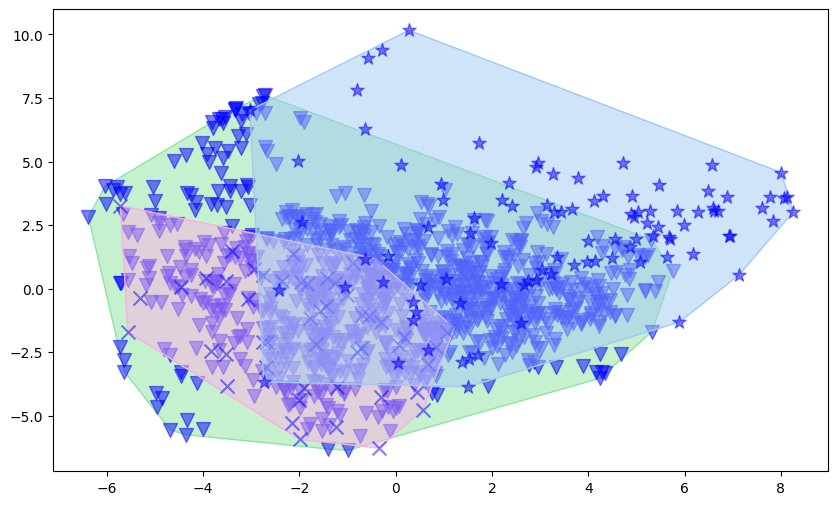}
     \includegraphics[width=0.24\textwidth]{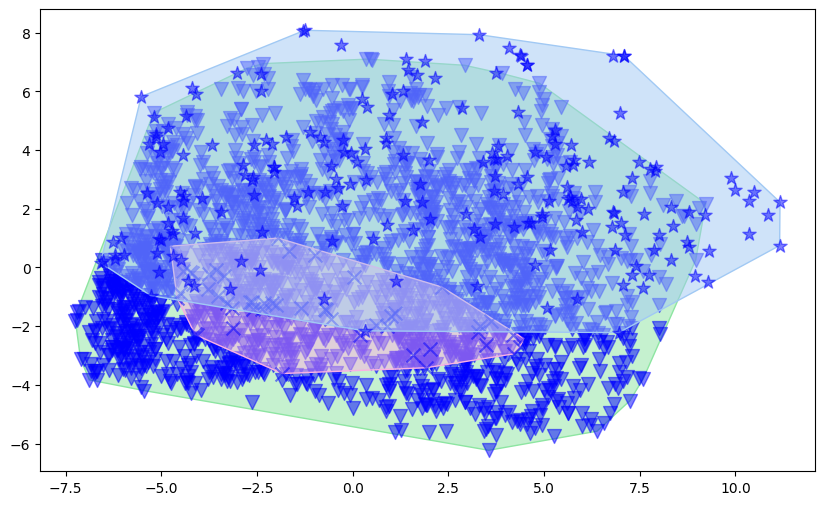}
     \includegraphics[width=0.24\textwidth]{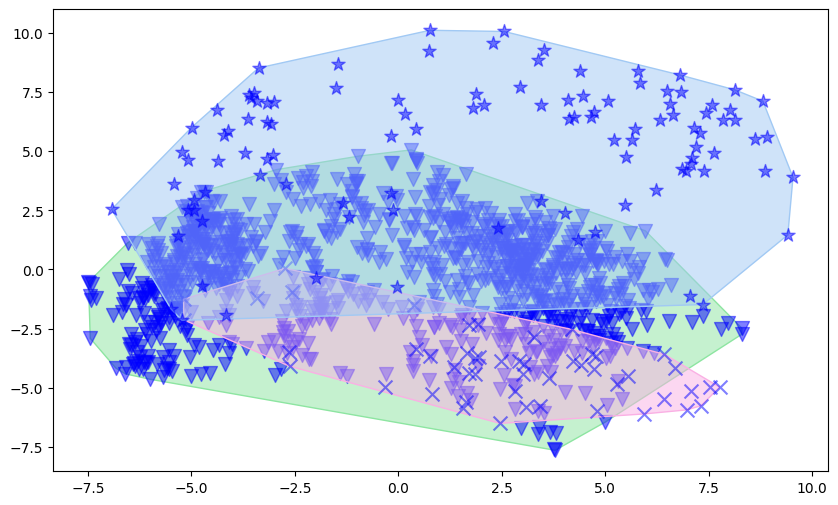}
     \includegraphics[width=0.24\textwidth]{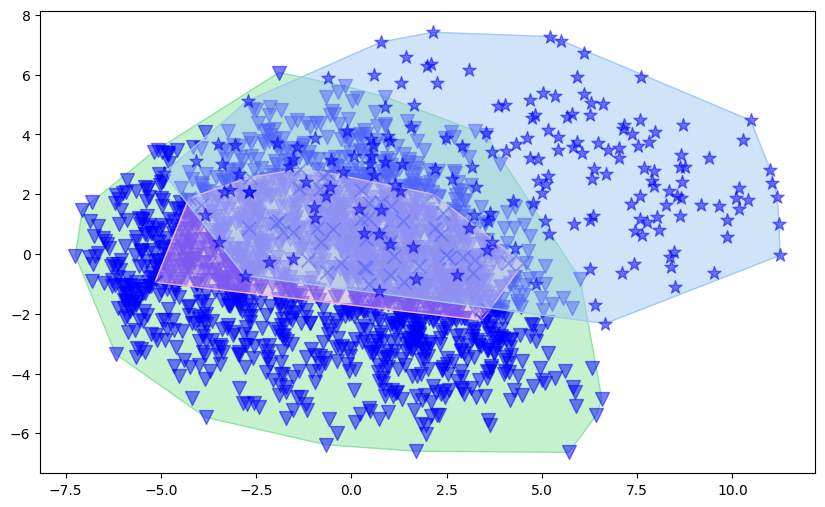}
\caption{MIDRE, $T$ = ResNet152, $a_e$ = 0.4, AttAcc = 15.97\%, $f_{train}^{MIDRE} =$ \colorbox{privRe}{$f_{RE}^{MIDRE}$}}
 \end{subfigure}
 
\caption{\textbf{Feature space analysis to show that, under MIDRE, $f_{recon}^{MIDRE}$ and $f_{priv}^{MIDRE}$ has a discrepancy, degrading MI attack.} We visualize penultimate layer activations of private images 
({\bf\color{blue}$\star$} $f_{priv}$), RE-private images 
({\color{blue}$\blacktriangledown$} $f_{RE}$), 
and MI-reconstructed images ({\bf\color{blue}$\times$} $f_{recon}$) generated by both (a) NoDef and (b) our MIDRE model. 
We also visualize the convex hull for  \colorbox{priv}{private images},  \colorbox{privRe}{RE-private images}, and \colorbox{inv}{MI-reconstructed images}. 
In (a), \colorbox{inv}{$f_{recon}^{NoDef}$} closely resemble \colorbox{priv}{$f_{priv}^{NoDef}$}, consistent with high attack accuracy. 
In (b), \colorbox{priv}{private images} and \colorbox{privRe}{RE-private images} share some similarity but they are not identical, with partial overlap between  \colorbox{priv}{$f_{priv}^{MIDRE}$} and \colorbox{privRe}{$f_{RE}^{MIDRE}$}.
Importantly, \colorbox{inv}{$f_{recon}^{MIDRE}$} closely resembles \colorbox{privRe}{$f_{RE}^{MIDRE}$} as RE-private is the training data for MIDRE. This results in \textbf{a reduced overlap between \colorbox{inv}{$f_{recon}^{MIDRE}$} and \colorbox{priv}{$f_{priv}^{MIDRE}$}, explaining that MI does not accurately capture the private image features under MIDRE.} 
}
\label{fig:f1_supp}
\end{figure*}

\begin{figure}
\vspace{-1cm}
    \centering
    \includegraphics[width=0.5\linewidth]{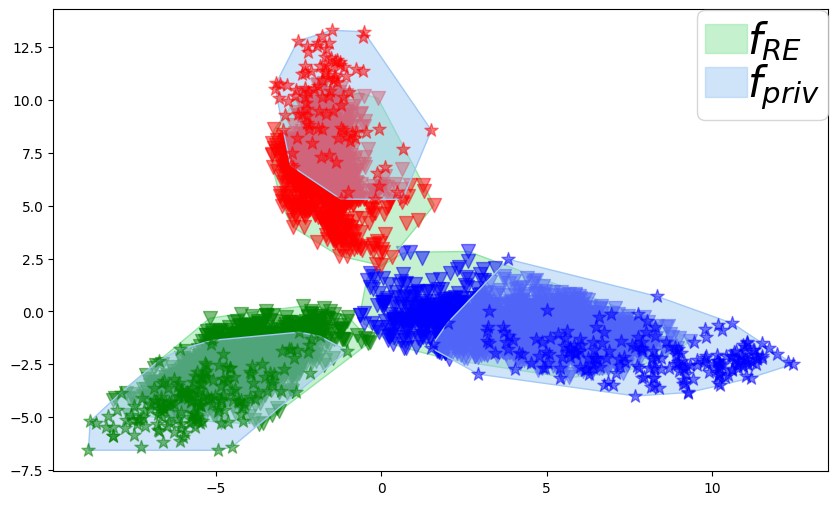}
    \caption{\textbf{MIDRE target model achieves high accuracy despite partial overlap of \colorbox{privRe}{$f_{RE}^{MIDRE}$} and \colorbox{priv}{$f_{priv}^{MIDRE}$}} using the target model $T$ = ResNet152. We visualize the penultimate layer activations of \colorbox{privRe}{RE-private images} and  \colorbox{priv}{private images}  for three identities. While $f_{RE}^{MIDRE}$ and $f_{priv}^{MIDRE}$ do not completely overlap, the model can still classify private images with high accuracy. 
This is because the private images remain distinct from other classes and distant from other classification regions, even when their representations are partially shared with  RE-private images (the training data).
We remark that RE randomly erases different regions from the images in different iterations, preventing the model to learn shortcut features and forcing the model to learn intr
insic features and become more generalizable beyond training data. 
}
\label{fig:midre_acc_supp}
\vspace{-1cm}
\end{figure}

The visualization in Fig. \ref{fig:f1_supp} shows the same trend as in Sec. 3.2. Specially, we observe the mismatch in feature space of MIDRE. 
Under MIDRE target model, 
\colorbox{privRe}{$f_{RE}^{MIDRE}$} and \colorbox{priv}{$f_{priv}^{MIDRE}$} have partial overlaps, but they  are not identical. Meanwhile, \colorbox{inv}{$f_{recon}^{MIDRE}$} tend to match with \colorbox{privRe}{$f_{RE}^{MIDRE}$} (RE-private images are training data for MIDRE, and follows the  discussion above).  
Therefore, $f_{recon}^{MIDRE}$ do not replicate $f_{priv}^{MIDRE}$, significantly degrading the MI attack. 
Furthermore, Fig. \ref{fig:midre_acc_supp} shows that  the mismatch between $f_{RE}^{MIDRE}$ and $f_{priv}^{MIDRE}$ does not cause the reduction of model utility. This is because the private images remain distinct from other classes and distant from other classification regions, even when their representations are partially overlapped with  RE-private images (the training data).

\subsection{Importance of partial erasure and random location for privacy-utility trade-off}

In this section, we analyse two properties of Random Erasing that are: {\color{blue}\textbf{Property P2:}} \textbf{Partial Erasure}, and {\color{blue}\textbf{Property P3:}} \textbf{Random Location}. 
In the addition of the setup in Sec. 3.2, we report a new experiment using the following setup: 
We use $T$ = MaxVIT, $D_{priv}$ = Facecrub \citep{ng2014data}, $D_{pub}$ = FFHQ \citep{karras2019style}, attack method = PPA \citep{struppek2022plug}. 

To evaluate the effectiveness of \textbf{Partial Erasure} and \textbf{Random Location}, we conduct experiments on three schemes: \textbf{Entire Erasing (EE), Fixed Erasing (FE)}, and \textbf{Random Erasing (RE)}. These schemes are compared against a No Defense baseline, which is trained for 100 epochs without any defense.
In Entire Erasing (EE) scheme, we progressively reduce the number of training epochs to simulate different levels of pixel concealment. Specifically, we train the model for 50, 60, 70, 80, 90, and 100 epochs, corresponding to 50\%, 40\%, 30\%, 20\%, 10\%, and 0\% pixel concealment, respectively.
For  Fixed Erasing (FE), a fixed location within each image is erased throughout the entire training process. However, the erased location varied across different images. For Random Erasing (RE), the location of erased areas is randomly selected for each image and training iteration. We train the RE model for 100 epochs with different values of the erasure ratio, $a_e$ = 0.5, 0.4, 0.3, 0.2, 0.1 corresponding to 50\%, 40\%, 30 \%, 20\%, and 10\% pixel concealment, respectively.

 We report the results in Tab. \ref{tab:reduc_epoch_supp}. 
 The results exhibit the same trend as outlined in Section 3.2 of the main paper. Specifically, \textbf{{\color{blue}\textbf{Property P2}}} demonstrates the privacy effect in defending against MI attacks, where partial erasure (fixed or random) proves more effective than entire erasure (reducing epochs) despite identical pixel concealment percentages.
\textbf{{\color{blue}\textbf{Property P3}}} validates the recovery of model utility, evidenced by the enhanced accuracy of RE models while archiving lower attack accuracy than FE models across varying erased portion ratios $a_e$.

\begin{table*}[ht]
\centering
\caption{We compare three different techniques for pixel concealment, to reduce the amount of private information presented to the model during training. Here, we use $T$ = MaxVIT, $\D_{priv}$ = Facescrub, $\D_{pub}$ = FFHQ, attack method = PPA. 
The results show that simply reducing epochs as in ``Entire Erasure'' is insufficient for degrading attack performance. Meanwhile, RE improves model utility while degrading attack accuracy effectively.}
\begin{tabular}{cccccccc}
\hline
\multirow{3}{*}{\textbf{Concealment}} & \multicolumn{5}{c}{\textbf{Partial Erasure}} &  \multicolumn{2}{c}{\multirow{2}{*}{\textbf{Entire Erasing}}} \\  \cmidrule{2-6} 

 & & \multicolumn{2}{c}{\textbf{Random Erasing}} & \multicolumn{2}{c}{\textbf{Fixed Erasing}} &  \\  \cmidrule{3-4} \cmidrule(l{2pt}r{2pt}){5-6}  \cmidrule(l{2pt}r{2pt}){7-8}

\multicolumn{1}{c}{} & \multicolumn{1}{c}{\textbf{$a_e$}} & \textbf{Acc ($\uparrow$)} & \textbf{AttAcc ($\downarrow$)} & \textbf{Acc ($\uparrow$)} & \textbf{AttAcc ($\downarrow$)} & \textbf{Acc ($\uparrow$)} & \textbf{AttAcc ($\downarrow$)} \\ \hline
0\% & 0 & 98.36 & 80.66 & 98.36 & 80.66 & 98.36 & 80.66 \\
10\% & 0.1 & \textbf{98.73} & \textbf{70.92} & 98.59 & 74.81 & 97.93	& 82.93 \\
20\% & 0.2 & \textbf{98.61} & \textbf{56.93} & 98.11 & 57.74 & 98.00 &	82.17 \\
30\% & 0.3 & \textbf{98.35} & \textbf{42.10} & 97.90 & 43.40 & 98.04	& 83.49  \\
40\% & 0.4 & \textbf{98.06} & \textbf{28.21} & 97.31 & 31.06 & 97.95	& 83.18  \\
50\% & 0.5 & \textbf{98.73} & 16.34 & 94.67 & \textbf{16.11} & 98.03	& 84.29 \\\hline
\end{tabular}

\label{tab:reduc_epoch_supp}
\end{table*}

\section{Ablation Study}

\subsection{Ablation study on alternative masking strategies}

\begin{figure}[ht!]
    \centering
    \begin{subfigure}[b]{0.45\textwidth}
        \centering
        \includegraphics[width=\textwidth]{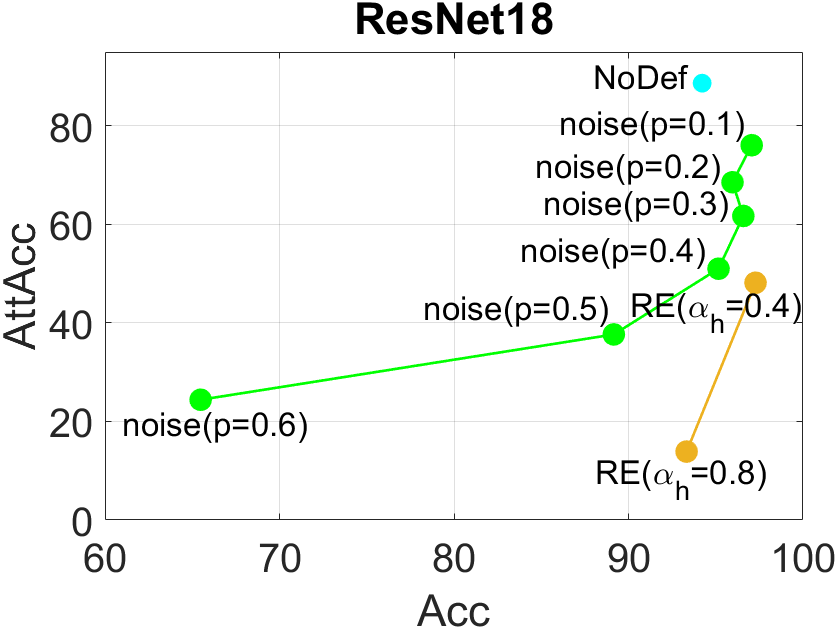}
        \caption{Masking Random Pixels}
        \label{fig:sub1}
    \end{subfigure}
    \hfill
    \begin{subfigure}[b]{0.45\textwidth}
        \centering
        \includegraphics[width=\textwidth]{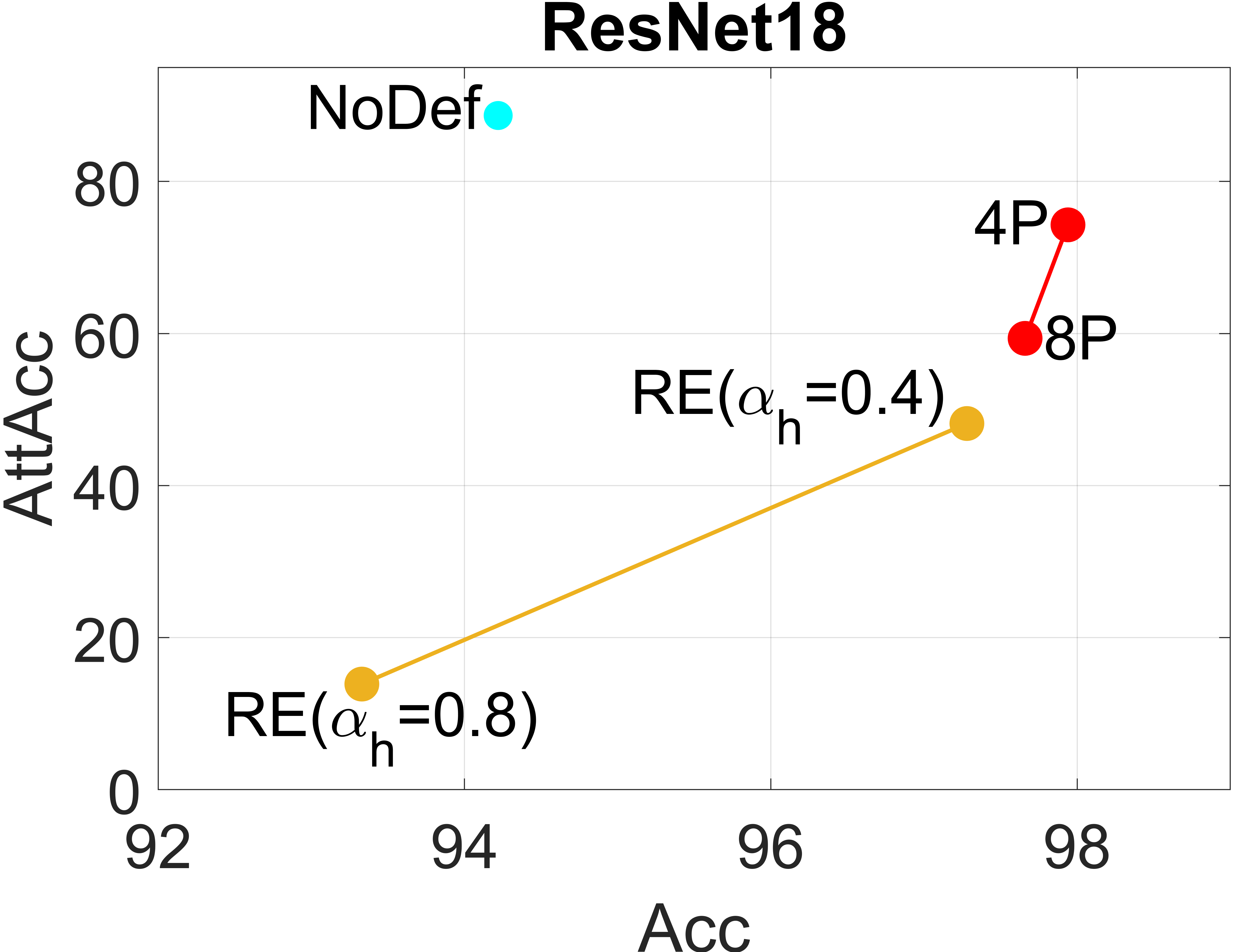}
        \caption{Masking Multiple Patches}
        \label{fig:sub2}
    \end{subfigure}
\caption{
We compare our RE masking strategy with two alternative masking approaches: (a) \textit{Masking Random Pixels}, and (b) \textit{Masking Multiple Patches}. In (a), we randomly mask a proportion of pixels from 10\% ($p=0.1$) to 60\% ($p=0.6$). In (b), we randomly mask either 4 small patches, denoted as 4P ($\alpha_e\in [0.025, 0.1]$), or 8 small patches, denoted as 8P ($\alpha_e\in [0.0125, 0.1]$). 
We evaluate these strategies using the PPA attack method with $T = \text{ResNet18}$, $\mathcal{D}_{\text{priv}} = \text{FaceScrub}$, and $\mathcal{D}_{\text{pub}} = \text{FFHQ}$. 
The results demonstrate that our RE masking strategy achieves a better privacy-utility trade-off compared to both Masking Random Pixels and Masking Multiple Patches.
}
\label{fig:masking_strategies}
\end{figure}

In this section, we conduct experiments using alternative masking strategies. In addition to the traditional random erasing method, we explore two additional approaches: (1) masking random pixels, and (2) masking multiple patches.

\begin{itemize}
    \item \textbf{Masking random pixels:} Instead of masking a square region as in our proposed Random Erasing (RE) method, we apply masking at the pixel level. For example, we randomly mask 10\% of the image pixels by replacing them with random values. In our experiments, we train the target model with varying levels of random pixel masking, ranging from 10\% ($p = 0.1$) to 60\% ($p = 0.6$).

    \item \textbf{Masking multiple patches:} Instead of masking a single large square region, we apply multiple smaller masks to the image. In our experiments, we randomly mask each training image with either 4 small patches (4P) or 8 small patches (8P). To ensure a fair comparison with MIDRE, we adjust the patch sizes accordingly. For 4P, we set $\alpha_e\in [0.025, 0.1]$, so that the total area of the four small patches is approximately equivalent to that of MIDRE with  $\alpha_e \in [0.1, 0.4])$ (RE($\alpha_h= 0.4$)). Similarly, for 8P, we use $\alpha_e\in [0.0125, 0.1]$, making the total masked area comparable to MIDRE with $\alpha_e \in [0.1, 0.8]$ (RE($\alpha_h= 0.8$)).
    
\end{itemize}

We summarize the results of the two alternative masking strategies in Fig.~\ref{fig:masking_strategies}. 
\begin{itemize}
    \item \textbf{Masking Random Pixels:} We clearly observe that this method performs better than the baseline (NoDef) in terms of reducing attack accuracy. However, it is less effective than our proposed MIDRE in both lowering attack accuracy and preserving natural accuracy.
    \item \textbf{Masking Multiple Patches:} Although distributing the masking across multiple smaller regions provides some privacy benefits, masking a single large region—as done in our approach—still achieves a better utility-privacy trade-off.
\end{itemize}

\subsection{Ablation study on Masking Values.}
In this section, we examine the effect of masking value to MIDRE performance. We select
attack method = PLGMI \citep{yuan2023pseudo}, $T$ = FaceNet64, $\D_{priv}$ = CelebA, $\D_{pub}$ = FFHQ.
We set $a_e$ = (0.2,0.2). 
Similar to \citep{zhong2020random}, we investigate four types of masking values: 0, 1, a random value, and the mean value. In case of random value, we randomly select it within a range (0,1). The mean value uses the ImageNet dataset's mean pixel values ([0.485, 0.456, 0.406]).

Tab. \ref{tab:Ablation_Substitue} demonstrates that the mean value offers the best balance between robustness against MI attacks and maintaining natural image accuracy.
Consequently, we adopt the Imagenet mean pixel values for masking in MIDRE.

\begin{table}[ht!]
\label{tab:bido}
\centering
\caption{ 
The effect of different masking value. We use attack method = PLGMI \citep{yuan2023pseudo}, $T$ = FaceNet64, $\D_{priv}$ = CelebA, $\D_{pub}$ = FFHQ. Overall, mean value achieves the best balance between robustness against MI attacks and maintaining natural image accuracy.
}
\begin{tabular}{lcccc}
\hline
Masking value & Acc  $\uparrow$ & AttAcc $\downarrow$ & $\Delta$  $\uparrow$ & Ranking \\ \hline
NoDef & 88.50&  95.00 $\pm$ 2.56  & - & -\\
\hline
0 & 83.72 & 69.20 $\pm$ 2.64 & 5.40 & 3 \\
1 & 83.68 & 70.00 $\pm$ 3.18 & 5.18 & 4 \\
random & 80.76 & 51.87 $\pm$ 4.43 & 5.57 & 2 \\
mean & 85.14 & 68.87 $\pm$ 3.97 & \textbf{7.78} & 1 \\
\hline
\end{tabular}

\label{tab:Ablation_Substitue}
\end{table}

\subsection{Ablation study on Area Ratio.}
In MIDRE, the area ratio $a_e$ controls the portion of an image masked to prevent MI attacks. This experiment investigates the impact of different $a_e$ values on MIDRE's performance. In particular, $a_e$ is randomly selected within the range (0.1, $a_h$), guaranting that at least 10\% of the image is always masked. We select three values for $a_h$: 0.3, 0.4, and 0.5.
Similar to the previous ablation study, we employ attack method = PLGMI \citep{yuan2023pseudo}, $T$ = FaceNet64, $\D_{priv}$ = CelebA, $\D_{pub}$ = FFHQ.
The masking process uses the ImageNet mean pixel values.

The results in Tab. \ref{tab:Ablation_Ratio} indicate that increasing $a_h$ strengthens MIDRE's defense against MI attacks, but this comes at the cost of reduced natural accuracy. To achieve a balance between robustness and natural accuracy, we opt $a_h$ = 0.4 in MIDRE.

\begin{table}[ht!]
\centering
\caption{The effect of area ratio. We use attack method = PLGMI \citep{yuan2023pseudo}, $T$ = FaceNet64, $\D_{priv}$ = CelebA, $\D_{pub}$ = FFHQ. To achieve a balance between robustness and natural accuracy, we opt $a_h$ = 0.4 in MIDRE.}
\label{tb:query_budget}
\begin{tabular}{lcccc}
\hline
$a_h$ & Acc  $\uparrow$ & AttAcc $\downarrow$ & $\Delta$  $\uparrow$ & Ranking \\ 
\hline
NoDef & 88.50&  95.00 $\pm$ 2.56 & - & -\\
\hline
0.3 & 83.55 & 65.07 $\pm$ 4.02 & 6.05 & 2 \\
0.4 & 81.65 & 51.60 $\pm$ 3.61 & \textbf{6.34} & 1 \\
0.5 & 78.50 &45.40 $\pm$ 3.85 & 4.96 & 3 \\
\hline
\end{tabular}

\label{tab:Ablation_Ratio}
\end{table}

\subsection{Ablation study on Aspect Ratio.}
We perform an ablation study on the aspect ratio of random erasing for model inversion defense. The results presented in Tab. \ref{tab:MIA64_aspect_ablation} demonstrate that the influence of aspect ratio on attack accuracy is not as significant as that of area ratio.

\begin{table*}[ht!]  
\centering
\caption{We report the LOMMA+KedMI attacks on images with resolution 64$\times$64. 
$T$ = VGG16, $D_{priv}$ = CelebA, $D_{pub}$ = CelebA with different aspect ratios of RE in MIDRE. We also put NoDef result as a baseline.
}
\begin{tabular}{lccccc}
\hline
Attack &  Defense & Acc  $\uparrow$  & AttAcc $\downarrow$ & $\Delta$ $\uparrow$ & KNN Dist $\uparrow$  \\ 
\hline
\multirow{4}{*}{LOMMA+KedMI} & NoDef & 86.90 & 81.80 $\pm$ 1.44 & - & 1211.45  \\
 & MIDRE & 79.85 & 43.07 $\pm$ 1.99 & 5.49 & 1503.89 \\
 & MIDRE(aspect ratio = 0.5) & 81.32 & 49.13 $\pm$ 1.53 & 5.85 & 1424.40 \\
 & MIDRE(aspect ratio = 2.0) & 81.65 & 51.87 $\pm$ 1.62 & 5.70 & 1440.00 \\ \hline
 
\end{tabular}

\label{tab:MIA64_aspect_ablation}
\end{table*}

\subsection{Adaptive attack}
\label{sec:adaptive}

We perform adaptive attacks in which the attacker knows the portions of the masking area $a_e$ and uses it during inversion attacks.
We use 2 setups: \textbf{Setup 1}: $T$ = ResNet152, $\D_{priv}$ = Facescrub, $\D_{pub}$ = FFHQ, Attack method = PPA, image size = 224 $\times$ 224. \textbf{Setup 2}: $T$ = VGG16, $\D_{priv}/\D_{pub}$ = CelebA, Attack method = LOMMA + KedMI, image size = 64 $\times$ 64. 
We use $a_e$ = [0.1,0.8] and $a_e$ = [0.1,0.4] for setup 1 and setup 2 to train MIDRE and during  attack. 


\textbf{Adaptive attacks fail to enhance attack performance in both two experimental setups} (See  Tab. \ref{tab:adaptive_attack}). This may be due to the dynamic masking positions employed in each attack iteration, hindering the convergence of the inverted images. Overall, even when attackers are fully informed about RE and use this knowledge to design an adaptive MI attack, they still fail to achieve accurate inversion results.

We compare the loss curves of the adaptive and normal attacks in Fig. \ref{fig:adaptive_attack_loss}. The results show that the dynamic masking positions in each iteration cause greater fluctuations in the adaptive attack loss compared to the normal attack. 
In addition,
PPA already incorporates learning rate adjustments during inversion, which do not reduce the loss fluctuations.

\begin{minipage}{0.5\textwidth}
    \centering
    \captionof{table}{We conduct the adaptive attacks where the attacker knows the masking area portions $a_e$ and uses them during inversion attacks. \textbf{Adaptive attacks (Adapt.Att) fail to enhance attack performance in both setups.}}
    \setlength{\tabcolsep}{2.5pt}
    \begin{tabular}{lll}
        \toprule
        \textbf{Setup} & \textbf{Attack} & \textbf{AttAcc} \\ \midrule
        \multirow{2}{*}{\textbf{Setup 1}} & MIDRE & 15.97 \\
         & MIDRE (Adapt.Att) & 10.50 \textbf{(-5.47\%)} \\ \midrule
        \multirow{2}{*}{\textbf{Setup 2}} & MIDRE & 43.07 \\
         & MIDRE (Adapt.Att) & 38.53 \textbf{(-4.54\%)} \\ \bottomrule
    \end{tabular}
    \label{tab:adaptive_attack}
\end{minipage}
\hfill
\begin{minipage}{0.45\textwidth}
    \centering
    \includegraphics[width=0.8\textwidth]{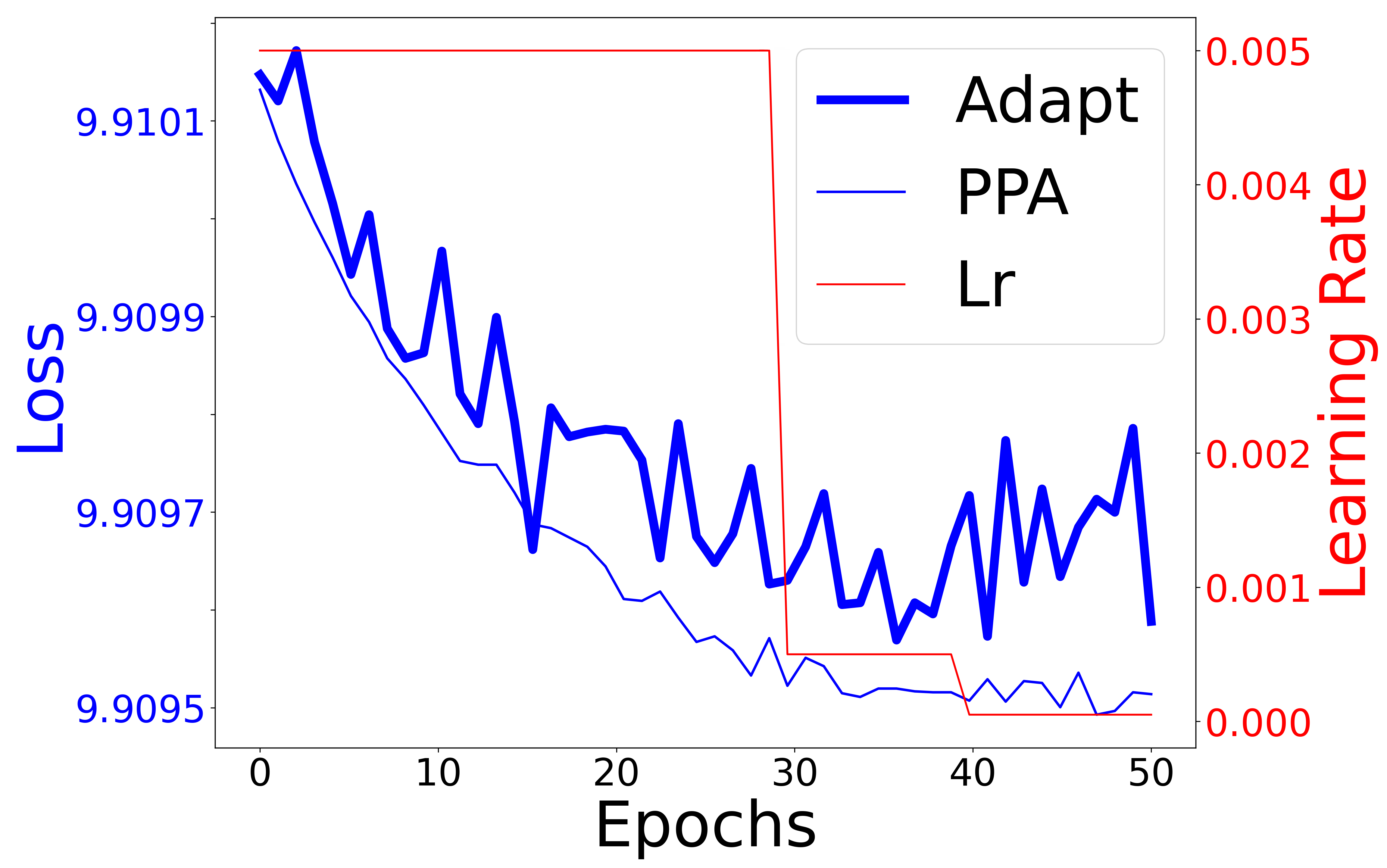}
    \captionof{figure}{PPA and PPA(Adapt) loss curves, with learning rate (Lr) adjustment.}
    \label{fig:adaptive_attack_loss}
\end{minipage}

\subsection{The effectiveness of substitute pixels generated by inpainting for MIDRE.}
We incorporated an inpainting method \citep{telea2004image} to replace masked values, following the experimental setup described earlier. Our results show that MIDRE (inpainting) modestly improves model accuracy while reducing the attack success rate by 4.34\%, which is indicated in Tab. \ref{tab:MIA64_inpaint_ablation}. However, this approach incurs a higher computational cost compared to RE.

\begin{table*}[ht!]  
\centering
\caption{We report the LOMMA+KedMI attack on images with resolution 64$\times$64. 
$T$ = VGG16, $D_{priv}$ = CelebA, $D_{pub}$ = CelebA to target models trained with RE with substitue pixel generate by inpaiting.
}
\begin{tabular}{lcccc}
\hline
Attack &  Defense & Acc  $\uparrow$  & AttAcc $\downarrow$ & KNN Dist  $\uparrow$ \\ 
\hline
\multirow{3}{*}{LOMMA+KedMI} & NoDef & 86.90 & 81.80 $\pm$ 1.44 & 1211.45  \\
 & MIDRE & 79.85 & \textbf{43.07 $\pm$ 1.99 } & 1503.89 \\
 & \textbf{MIDRE (inpainting)} & 80.42 & \textbf{38.73 $\pm$ 1.27 } & \textbf{1508.28} \\ \hline
 
\end{tabular}

\label{tab:MIA64_inpaint_ablation}
\end{table*}

\section{Discussion}
\label{sec:discussion}

We propose a new defense against MI attacks using Random Erasing (RE) during training. RE reduces private information exposure while significantly lowering MI attack success, with small impact on model accuracy. Our method outperforms existing defenses across 34 experiment setups using 7 SOTA MI attacks, 11 model architectures, 6 datasets, and user study.

\subsection{Broader Impacts}
Model inversion attacks, a rising privacy threat, have garnered significant attention recently. By studying defenses against these attacks, we can develop best practices for deploying AI models and build robust safeguards for applications, especially those that rely on sensitive training data. Research on model inversion aims to raise awareness of potential privacy vulnerabilities and strengthen the defense.

\subsection{Limitation}
Firstly, we currently focus on enhancing the robustness of classification models against MI attacks. 
This is really important because these models are being used more and more in real-life situations where privacy and security are a major concern.
In the future, we plan to expand our research scope to encompass MI attacks and defenses for a broader range of machine learning tasks.

Secondly, while our current experiments are comprehensive compared to prior works \citep{zhang2020secret,chen2021knowledge,nguyen2023re,kahla2022label,struppek2022plug, ho2024model, koh2024vulnerability} which mainly focus on image data, real-world applications often involve diverse private/sensitive training data. Addressing these real-world data complexities through a comprehensive approach will be essential for building robust and trustworthy machine learning systems across various domains.

\section{Experiments Compute Resources}
\label{sec:Compute_Resource}
In order to carry out our experiments, we utilise a workstation equipped with the Ubuntu operating system, an AMD Ryzen CPU, and 4 NVIDIA RTX A5000 GPUs. Furthermore, we utilise a secondary workstation equipped with the Ubuntu operating system, an AMD Ryzen CPU, and two NVIDIA RTX A6000 GPUs.

\section{Related Work}
\label{sec:related_wor}
\subsection{Model Inversion Attacks}
The GMI \citep{zhang2020secret} is a pioneering approach in model inversion to leverages publicly available data and employs a generative model GAN to invert private datasets. This methodology effectively mitigates the generation of unrealistic data instances. KedMI \citep{chen2021knowledge} can be considered an enhanced iteration of the GMI model, as it incorporates the transmission of knowledge to the discriminator through the utilization of soft labels. PLGMI \citep{yuan2023pseudo} is the current leading model inversion method in the field. It leverages pseudo labels derived from public data and the target model.  LOMMA \citep{nguyen2023re} employs an augmented model in order to reduce the model inversion overfitting. The augmented model is trained to distill knowledge from a target model by utilizing public data. During attack, the attackers generate images in order to minimize the identity loss in both the target model and the augmented model. However, it should be noted that the aforementioned four approaches are specifically designed for target models that have been trained on low-resolution data, specifically 64x64 for the CelebA private dataset. Recently, PPA \citep{struppek2022plug}, MIRROR \citep{an2022mirror}, and  DMMIA \citep{qi2023model}, IF-GMI\citep{qiu2024closer} perform the attack on high resolution images.
In addition, Kahla et. al. \citep{kahla2022label} introduced the BREPMI attack as a form of label-only model inversion attack, where the assault is based on the predicted labels of the target model. Another work is RLBMI \citep{han2023reinforcement}, which utilizes a reinforcement learning approach to target a model in a black box scenario.

\subsection{Model Inversion Defenses}

To defend against MI attacks, differential privacy (DP) \citep{dwork2006differential,dwork2008differential}
has been  studied in earlier works.
Studies in \citep{dwork2006differential,dwork2008differential} have shown
that current DP mechanisms do not mitigate MI attacks while
 maintaining  desirable model utility at the same time.
More recently, regularizations have been proposed for MI defenses \citep{wang2021improving,peng2022bilateral,struppekcareful}.
\citep{wang2021improving} propose regularization loss to the training objective to   
 limit the dependency between the model inputs and outputs.
In BiDO \citep{peng2022bilateral}, 
they propose regularization to limit the  the dependency between
the model inputs and  latent representations.
However, these
regularizations  conflict with the training loss and harm model utility considerably.
To restore the model utility partially, \citep{peng2022bilateral} propose to add another  regularization loss to maximize  the dependency between latent representations and the outputs.
However, searching for  hyperparameters for two regularizations in BiDO requires  computationally-expensive.  
Recently, \citep{ye2022one} introduced a new approach that utilises differential privacy to protect against model inversion. \citep{gong2023gan} proposed a novel Generative Adversarial Network (GAN)-based approach to counter model inversion attacks.
In this paper, we do not conduct experiments to compare to these methods as the code is not available.
\citep{struppekcareful} study the effect of label smoothing regularization on model privacy leakage. Their findings demonstrate that positive label smoothing factors can amplify privacy leakage, whereas negative label smoothing factors mitigate privacy concerns at the cost of a substantial decrease in model utility, resulting in a more favorable utility-privacy trade-off. Recently, 
\citep{ho2024model} introduce a novel approach to defending against model inversion attacks by focusing on the model training process. Their proposed Transfer Learning-based Defense against Model Inversion (TL-DMI) aims to restrict the number of layers that encode sensitive information from the private training dataset into the model. As restricting the number of model parameters that encode private information can potentially impact the model's performance.
\citep{koh2024vulnerability} study the  impact of DNN architecture designs, particularly skip connections, on model inversion attacks.
They found that removing skip connections in the last layers can enhance model inversion robustness. However, this approach necessitates searching for optimal skip connection removal and scaling factor combinations, which can be computationally intensive.
Both TL-DMI and MI-RAD experiences difficulty in achieving competitive balance between utility and privacy.
We show comparison of several defense approaches with our MIDRE in Fig. 1 (main paper).

\end{document}